\documentclass[twocolumn,english,aps,prl,showpacs,superscriptaddress,groupedaddress]{revtex4-2}
\usepackage{amsmath}
\usepackage{amssymb}

\usepackage{graphicx}

\makeatletter

\usepackage{hyperref}

\usepackage[sort&compress]{natbib}
\usepackage{notoccite}

\makeatother

\begin{document}
\title{Statistical Mechanics of Deep Linear Neural Networks:\\
The Back-Propagating Kernel Renormalization }
\author{Qianyi Li$^{1,2}$and Haim Sompolinsky$^{2,3,4}$}
\affiliation{$^{1}$The Biophysics Program, Harvard University, Cambridge, MA 02138,
USA~~\\
 $^{2}$Center for Brain Science, Harvard University, Cambridge, MA
02138, USA~~\\
 $^{4}$Racah Institute of Physics, Hebrew University, Jerusalem 91904,
Israel~~\\
 $^{5}$Edmond and Lily Safra Center for Brain Sciences, Hebrew University,
Jerusalem 91904, Israel}
\begin{abstract}
The groundbreaking success of deep learning in many real-world tasks
has triggered an intense effort to understand theoretically the power
and limitations of deep learning in the training and generalization
of complex tasks, so far with limited progress. In this work we study
the statistical mechanics of learning in Deep Linear Neural Networks
(DLNNs) in which the input-output function of an individual unit is
linear. Despite the linearity of the units, learning in DLNNs is highly
nonlinear, hence studying its properties reveals some of the essential
features of nonlinear Deep Neural Networks (DNNs). Importantly, we
solve exactly the network properties following supervised learning
using an equilibrium Gibbs distribution in the weight space. To do
this, we introduce the Back-Propagating Kernel Renormalization (BPKR),
which allows for the incremental integration of the network weights
layer-by-layer starting from the network output layer and progressing
backward until the first layer's weights are integrated out. This
procedure allows us to evaluate important network properties, such
as its generalization error, the role of network width and depth,
the impact of the size of the training set, and the effects of weight
regularization and learning stochasticity. BPKR does not assume specific
statistics of the input or the task's output. Furthermore, by performing
\emph{partial} integration of the layers, the BPKR allows us to compute
the emergent properties of the neural representations across the different
hidden layers. We have proposed a heuristic extension of the BPKR
to nonlinear DNNs with rectified linear units (ReLU). Surprisingly,
our numerical simulations reveal that despite the nonlinearity, the
predictions of our theory are largely shared by ReLU networks of modest
depth, in a wide regime of parameters. Our work is the first exact
statistical mechanical study of learning in a family of Deep Neural
Networks, and the first successful theory of learning through the
successive integration of Degrees of Freedom in the learned weight
space.
\end{abstract}
\maketitle

\section{Introduction}

Gradient-based learning in multilayered neural networks has achieved
surprising success in many real-world problems including machine vision,
speech recognition, natural language processing, and multi-agent games
\citep{foerster2016learning,goldberg2017neural,lecun1999object,deng2013new}.
Deep learning (DL) has been applied successfully to basic and applied
problems in physical, social and biomedical sciences and has inspired
new neural circuit models of information processing and cognitive
functions in animals and humans \citep{banino2018vector,guo2016deep}.
These exciting developments have generated a widespread interest in
advancing the theoretical understanding of the success and limitations
of DL, and more generally, in computation with Deep Neural Networks
(DNNs). Nevertheless, many fundamental questions remain unresolved
including the puzzling ability of gradient-based optimization to avoid
being trapped in poor local minima, and the surprising ability of
complex networks to generalize well despite the fact that they are
usually heavily over-parameterized – namely, the number of learned
weights far exceeds the minimal number required for perfectly fitting
the training data \citep{poggio2020theoretical,zhang2021understanding}.
These problems have fascinating ramifications for statistical mechanics,
such as energy landscapes in high dimensions, glassy dynamics, and
the role of degeneracy, symmetry and invariances \citep{baity2018comparing,ballard2017energy,becker2020geometry,rifai2011contractive}.
Indeed, statistical mechanics has been one of the most fruitful theoretical
approaches to learning in neural networks \citep{carleo2019machine,engel2001statistical,mezard2009information,advani2017high}.
However, its classical phenomenology of capacity, learning curves,
and phase transitions, was formulated largely in the context of single-layer
or shallow architectures.

In this work we develop a new statistical mechanical theory, appropriate
for learning in deep architectures. We focus on the statistical mechanics
of weight space in deep linear neural networks (DLNNs) in which single
neurons have a linear input-output transfer function. DLNNs do not
possess superior computational power over a single-layer linear perceptron
\citep{yuan2012recent}. However, because the input-output function
of the network depends on products of weights, learning is a highly
nonlinear process and exhibits some of the salient features of the
nonlinear networks. Indeed, in very interesting recent work \citep{saxe2019mathematical,saxe2014exact},
the authors investigated the nonlinear gradient descent dynamics of
DLNNs. These studies focused on the properties of the dynamic trajectories
of gradient-based learning. To tackle the problem analytically, they
had to rely on restrictive assumptions about initial weights and on
simplifying assumptions about the data statistics. In contrast, we
focus on the equilibrium properties of the distribution in weight
space induced by learning, allowing us to address some of the fundamental
problems in DL, such as the features determining the DNN's ability
to generalize despite over-parameterization, the role of depth and
width, as well as the size of the training set, and the effect of
regularization and learning stochasticity (akin to temperature).

To analyze the property of the weight distribution, we consider the
posterior probability distribution in the weight space after learning
with a Gaussian prior under a Bayesian framework \citep{tishby1989consistent,mackay1992practical,neal2012bayesian}.
As introduced in Section $\text{\ref{sec:The-Back-propagating-KR}}$,
the posterier distribution of the weights can also be formulated as
a Gibbs distribution with a cost function consisting of the training
error and an $L_{2}$ weight regularization term. The Bayesian formulation
and the Gibbs distribution of the weights have become a standard framework
for analyzing statistical properties of neural network models and
have been applied in various studies on the statistical mechanics
of learning \citep{bahri2020statistical,amit1987statistical,advani2013statistical,engel2001statistical,seung1992statistical,watkin1993statistical}.
In most of our analysis we constrain ourselves to the zero-temperature
limit, in which case the network attains zero training error when
operating below capacity, and the Gaussian prior introduces bias to
the weight distribution to favor weights with smaller $L_{2}$ norms
within the weight space that yields zero training error. 

\begin{figure*}[tp]
\includegraphics[clip,width=1\textwidth]{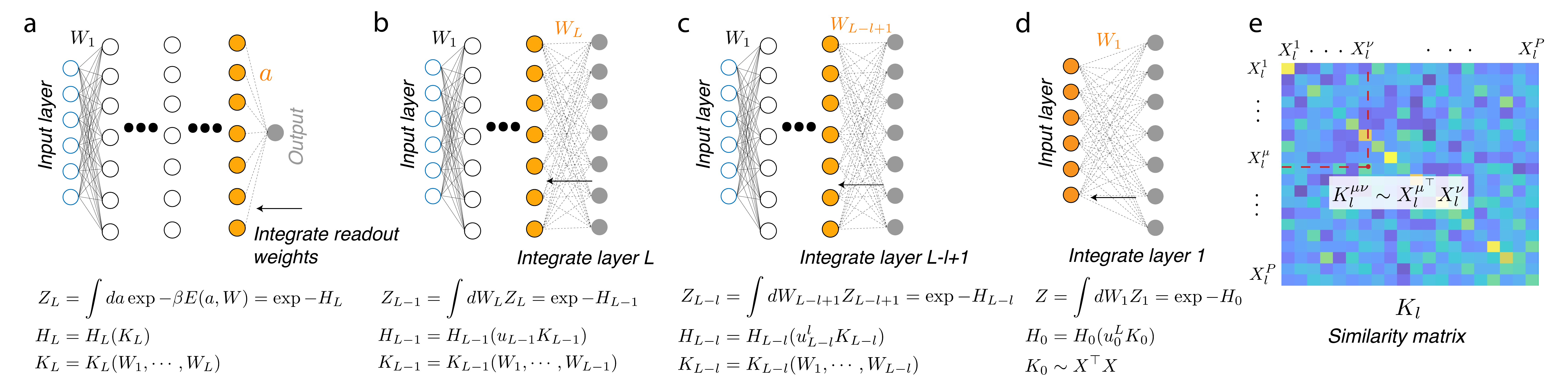}

\caption{\label{fig:schematics}Schematics of the Back-Propagating Kernel Renormalization. (a) Integrating out the readout weights a of the network yields a partial partition function $Z_{L}$ in the weight space with an effective Hamiltonian $H_{L}$, which is a function of all the hidden layers' weights through its dependence on the $L$-th layer kernel matrix (see (e)) (with an additional $L_{2}$ regularization on the remaining weights neglected here). (b) Integrating out layer $L$ yields a partial partition function $Z_{L-1}$ in the remaining weight space with an effective Hamiltonian $H_{L-1}$ which has the same structure as $H_{L}$ except that the $L-1$-th layer kernel is multiplied by an order parameter $u_{L-1}$, a scalar renormalization variable which ‘summarizes’ the effect of the $L$-th layer weights on the effective Hamiltonian. (c) Similarly, integrating out all weights downstream of layer $L-l$ yields a partial partition function $Z_{L-l}$ with an effective Hamiltonian with an order parameter $u_{L-l}$, ‘summarizing’ the effect of all $l$ upstream integrated layers. (d) Integrating out all the weights in the network yields the total partition function $Z$ with the total free energy of the system $H_{0}$ with an order parameter $u_{0}$, which depends only on the training inputs $X^{\mu},\mu=1,...,P$. (e) The $l$-th layer kernel matrix is the similarity matrix of this layer's responses to the $P$ training inputs up to a normalization factor and is a function of all upstream weights.}
\end{figure*}

We evaluate statistical properties in weight space induced by DL by
successive \emph{backward} integration of the weights layer-by-layer
starting from the output layer. As shown in $\text{Fig.\ref{fig:schematics}}$,
each stage of the successive integration of a layer of weights yields
an effective Hamiltonian of the remaining upstream weights. As Section
$\text{\ref{sec:The-Back-propagating-KR}}$ shows, this effective
Hamiltonian is expressed in terms of a renormalized kernel matrix
$K_{l}$ which is the \emph{$P\times P$} matrix of the overlaps of
all pairs of vectors of activations of the $l$-th layer induced by
the $P$ inputs of the training set. This matrix is a function of
all upstream weights, and in the successive integration process is
renormalized by a scalar renormalization variable which ‘summarizes’
the effect of the integrated downstream weights on the effective Hamiltonian
of the remaining weights. Therefore, we refer to the successive backward
integration process as Back-Propagating Kernel Renormalization (BPKR).
Using mean field techniques, this scalar renormalization variable
can be evaluated by a self-consistent equation, exact in the thermodynamic
limit. Thus, our theory is the first exact statistical mechanical
study of the weight space properties of DNNs and the first discovery
of kernel renormalization of the learned degrees of freedom (DoFs).
Our BPKR is schematically explained in $\text{Fig.\ref{fig:schematics}}$,
and described in detail in Section $\text{\ref{sec:The-Back-propagating-KR}}$.

Our work is closely related to the interesting recent research on
infinitely wide DNNs \citep{lee2017deep,cho2009kernel}. It is well
known that that the ensemble of input-output functions implemented
by infinitely wide networks are equivalent to a Gaussian Process (GP)
in function space with covariance matrix defined by a Gaussian kernel,
which is the kernel matrix averaged over weights sampled from the
Gaussian distribution. This GP limit holds when the network width,
the number of neurons in each layer, $N$, approaches infinity while
the size of the training data, $P$, is held constant, severely limiting
its applicability to most realistic conditions. In contrast, our theory
holds in the thermodynamic limit assumed in most statistical mechanical
studies of neural computation \citep{advani2013statistical,amit1987statistical,chung2018classification,ganguli2010statistical,ganguli2012compressed},
namely letting both $N$ and $P$ approach infinity while keeping
the load $\alpha=P/N$ fixed. As we show here, the behavior of the
system at finite $\alpha$ is often qualitatively different from the
infinite width limit, and our theory correctly predicts network behavior
in a much wider range of parameters that are more relevant in real
learning tasks, providing much richer insight into the complex properties
of DL in large networks, and into the role of the training data in
shaping the network performance and emergent representations. Our
theory applies to the entire range of $0\leq\alpha\leq\infty$ . We
introduce the new notions of \emph{wide} and \emph{narrow} networks
depending on whether $\alpha$ is smaller or larger than $1$. Narrow
networks in particular, deviate qualitatively from the $\alpha\rightarrow0$
limit of the GP theory. This is because when $\alpha>1$, a zero training
error solution cannot be achieved just by optimizing the readout weights
but necessarily requires appropriate changes in the hidden-layer weights.

In Section $\text{\ref{sec:Generalization}}$ we apply our theory
to derive the network generalization performance and its dependence
on the architectural parameters of width and depth as well as the
size and statistics of the training dataset. We calculate the system's
phase diagram and show the important role of the $L_{2}$ weight regularization
parameter, $\sigma$. In Section $\text{\ref{sec:Extensions}}$ we
present two important extensions. First, we extend our network architecture
to include multiple outputs (denoted as $m>1$) and show that in this
case the BPKR is characterized by an $m\times m$ kernel renormalization
\emph{matrix}. Interestingly, their $m$ eigenvalues are independent
and obey self-consistent equations similar to the single-output case.
While most of our study focuses on the zero-temperature limit of the
weight space Gibbs distribution, taking into account only the portion
of weight space that yields zero training error, we show in Section
$\text{\ref{sec:Extensions}}$ that our BPKR is readily applicable
to the finite temperature case and discuss the way kernel renormalization
affects the effect of temperature for networks of different depths. 

The power of the BPKR is that it allows the computation of not only
the system's performance as a whole but also the representation of
the data at each layer, readily captured by the statistics of the
mean layerwise kernel matrices. We show in Section $\text{\ref{sec:Changes-in-Representations}}$
how both the input and the task statistics affect these representations:
for instance, revealing underlying block structures of the task. In
Section $\text{\ref{sec:Deep-ReLu-Networks}}$ we present a heuristic
extension of the BPKR to nonlinear deep networks and test numerically
its predictions for ReLU networks. Surprisingly, we find that this
approximation nicely predicts the behavior of ReLU networks with modest
depth and not too small width $N$. Our results are discussed in the
last Section.

\section{\label{sec:The-Back-propagating-KR}The Back-propagating Kernel Renormalization
for DLNNs}

\subsection{\label{subsec:Statistical-mechanics-of} Statistical mechanics of learning in deep networks}

We consider a multilayer network with $L$ hidden layers whose input-output
mapping is given by

\begin{equation}
f(x,\Theta)=\frac{1}{\sqrt{N_{L}}}\sum_{i=1}^{N_{L}}a_{i}\phi_{i}(x,W)\label{eq:predictor}
\end{equation}
where $x\in R^{N_{0}}$, is an input vector of dimension $N_{0}$,
$\phi_{i}$ is the response of a neuron in the top hidden layer (of
size $N_{L}$) to that input. In general, $\phi$ is a nonlinear function
of $x$ and the network hidden weights, denoted by $W$. The output
of the network is a scalar which sums linearly the top layer activations
weighted by the readout weights $a_{i}$. We denote all network weights
by $\Theta=(a,W)$.

We assume supervised learning with the following cost function,

\begin{equation}
E(\Theta)=\frac{1}{2}\sum_{\mu=1}^{P}\left(f(x^{\mu},\Theta)-y^{\mu}\right)^{2}+\frac{T}{2\sigma^{2}}\Theta^{\top}\Theta\label{eq:Loss-1}
\end{equation}
The first term is the mean squared deviation of the network outputs
on a set of $P$ training input vectors $x^{\mu},\mu=1,...,P$ from
their target labels $y^{\mu}.$ The second term, with amplitude $T\sigma^{-2}$,
is a regularization term which favors weights with small $L_{2}$
norm. The temperature parameter $T$ in this term means that the $L_{2}$
regularization acts as an entropic term. In particular, in the regime
where we are mostly interested, $T\rightarrow0$, the first term will
enforce minimization of the training error while the $L_{2}$ term
shapes the statistical measure of the weight vectors that minimize
the training error, biasing it in favor of weights with small norms.
Without this term, all weights that minimize the error would have
the same probability. On the other hand, the $L_{2}$ term is irrelevant
at low $T$ if the minimum of the error is unique. We will call the
parameter $\sigma$, the weight \emph{noise} parameter as it controls
the amount of fluctuation in the weights at zero $T$.

We investigate the properties of the equilibrium distribution of the
weights, defined by the Gibbs distribution, $P(\Theta)=Z^{-1}\exp(-E/T)$,
where $Z$ is the partition function, $Z=\int d\Theta\exp(-E/T)$.
The Gibbs distribution is equivalent to the posterior distribution
of the weights with a Gaussian prior. 

The fundamental statistical mechanical properties of the system can
be derived from the partition function and its extensions as shown
below. However, calculating $Z$ exactly is intractable, but integrating
out the readout weights is straightforward, and by doing this we write
$Z=\int dWZ_{L}(W)$ , $Z_{L}(W)=\int da\exp(-E/T)=\exp[-H_{L}(W)]$
where $H_{L}(W)$ is the effective Hamiltonian of the hidden-layer
weights $W$ after integrating out the readout weights $a$ (Fig.$\text{\ref{fig:schematics}}$(a),
see details in Appendix $\text{\ref{subsec:appendixa}}$),

\begin{multline}
H_{L}(W)=\frac{1}{2\sigma^{2}}\text{Tr}\ensuremath{W^{\top}W}+\frac{1}{2}Y^{\top}(K_{L}(W)+TI)^{-1}Y\\
+\frac{1}{2}\log\det(K_{L}(W)+TI)\label{eq:H_L0T}
\end{multline}
where $Y$ is the $P\times1$ column vector of the training target
labels. The matrix $K_{L}$ is a $P\times P$ kernel matrix of the
top layer. We assume for simplicity that all the hidden layers have
equal width $N$. For each layer, we define its kernel matrix by

\begin{equation}
K_{l}=\frac{\sigma^{2}}{N}X_{l}^{\top}X_{l}\label{eq:layerkernels}
\end{equation}
where $X_{l}$ is the $N\times P$ matrix of activation of the $l$-th
layer in response to the training inputs, $X_{i,l}^{\mu}=\phi_{i}^{l}(x^{\mu},W')$,
$W'=\{W_{k}\}_{k<l+1}$ denotes all the weights upstream of $l$.
{[}Importantly, unlike other uses of kernels (e.g., in SVMs and DNNs,
\citep{lee2017deep,cho2009kernel}) here we define kernels as simply
the un-averaged dot products of the representations at the corresponding
layers, hence the $l$-th kernel matrix is a function of all the weights
upstream of $X_{l}$, and in particular $K_{L}$ depends on all the
hidden-layer weights $W$.{]}

\textbf{Thermodynamic limit: }The results we will derive throughout
are exact in the thermodynamic limit, which is defined as $N,N_{0},P\rightarrow\infty$
while $\alpha=P/N$ and $\alpha_{0}=P/N_{0}$ remain finite. Aside
from these limits, we do not make any assumptions about the training
inputs $x^{\mu}$ or the target outputs $Y$.

\textbf{Zero temperature: }Although our theory is developed for all
temperatures (Section \ref{subsec:Finite-temperature} and Appendix
\ref{subsec:appendixa}), our primary focus is on the limit of zero
temperature, exploring the statistical properties of the solution
weight space, namely the space of all $\Theta$ that yields zero training
error. The zero $T$ theory is particularly simple for $\alpha<1$
in which case the kernel matrices Eq.\ref{eq:layerkernels} are full
rank. Substituting the zero-temperature limit, $H_{L}(W)$ reduces
to
\begin{multline}
H_{L}(W)=\frac{1}{2\sigma^{2}}\text{Tr}\ensuremath{W^{\top}W}+\frac{1}{2}Y^{\top}K_{L}(W)^{-1}Y\\
+\frac{1}{2}\log\det(K_{L}(W))\label{eq:H_L}
\end{multline}
We will call networks with $N>P$ \emph{wide networks}. \emph{Narrow
networks} ($\alpha>1$) will be discussed at the end of this Section.

\textbf{Linear neurons:} Integrating over the weight matrices $\{W_{k}\}_{k\leq L}$
is an intractable problem in general. Here we focus on the simple
case where all the input-output functions $\phi$ are linear, so that
$x_{i,l}=\frac{1}{\sqrt{N}}w_{l}^{i\top}x_{l-1}$.

\subsection{The Back-Propagating Kernel Renormalization}

Even in the linear case the Hamiltonian Eq.\ref{eq:H_L} is not quadratic
in the weights, thus integrating out the weights is highly non-trivial.
Instead we compute the full partition function $Z$ by successive
integrations, in each of them only a single-layer weight matrix is
integrated and yields a partial partition function of the remaining
DoFs in the `weight space' (see schematics in Fig.$\text{\ref{fig:schematics}}$).
Starting from the top-layer $W_{L},$ we can move backward until all
weights are integrated out. Integrating the top hidden-layer weight
matrix $W_{L}$ yields a partial partition function, $Z_{L-1}=\int dW_{L}Z_{L}(W)=\exp[-H_{L-1}]$

\begin{multline}
H_{L-1}(W',u_{L-1})=\frac{1}{2\sigma^{2}}\text{Tr}\ensuremath{W'^{\top}W'}+\frac{1}{2u_{L-1}}Y^{\top}K_{L-1}^{-1}Y\\
+\frac{1}{2}\log\det(K_{L-1}u_{L-1})-\frac{N}{2}\log u_{L-1}+\frac{1}{2\sigma^{2}}Nu_{L-1}\label{eq:H_L-1}
\end{multline}
where $W'$ $=\{W_{k}\}_{k<L}$ denotes all the weights upstream of
$W_{L}$ (schematically shown in Fig.$\text{\ref{fig:schematics}}$(b)).
The first three terms are similar in form to $H_{L}$ , Eq.\ref{eq:H_L},
with $K_{L-1}(W')$ denoting the $P\times P$ kernel matrix of the
$L-1$ layer (Eq.\ref{eq:layerkernels} with $l=L-1$), which is now
a function of $W'$. The kernel terms in $H_{L-1}$ are renormalized
by a scalar $u_{L-1}$, representing the effect of the integrated
$W_{L}$ on the effective Hamiltonian of $W'$. While $u_{L-1}$ originally
appears as an auxiliary integration variable (Appendix $\text{\ref{subsec:appendixa}}$),
in the thermodynamic limit it is an order parameter determined self-consistently
by minimizing $H_{L-1}$,

\begin{equation}
1-\sigma^{-2}u_{L-1}=\alpha(1-u_{L-1}^{-1}r_{L-1})\label{eq:u_L-1}
\end{equation}
where we have denoted for general $l$,

\begin{equation}
r_{l}=\frac{1}{P}Y^{\top}K_{l}^{-1}Y\label{eq:b_l}
\end{equation}
We call this quantity the $l$-th layer\textbf{ mean squared readout},
since it equals the squared mean of the vector of output weights that
readout the target labels directly from layer $l$ (after training
the full network) (see Appendix $\text{\ref{subsec:appendixa}}$).
As will be shown, the mean square readouts are key parameters that
capture the effect of the task on the properties of the trained network.

To proceed we note that apart from scaling by the order parameter
$u_{L-1}$, Eq.\ref{eq:H_L-1} has exactly the same form as Eq.\ref{eq:H_L},
hence the steps of integration of the remaining weights can be repeated
layer-by-layer. After the $l$-th iteration, we obtain (Fig.$\text{\ref{fig:schematics}}$(c))

\begin{multline}
H_{L-l}(W')=\frac{1}{2\sigma^{2}}\text{Tr}\ensuremath{W'^{\top}W'}+\frac{1}{2u_{L-l}^{l}}Y^{\top}K_{L-l}^{-1}Y\\
+\frac{1}{2}\log\det(K_{L-l}u_{L-l}^{l})-\frac{lN}{2}\log u_{L-l}+\frac{lN}{2\sigma^{2}}u_{L-l}\label{eq:HL-l}
\end{multline}
which is a function of $W'=\{W_{k}\}_{k<L-l+1}$ (note that the superscript
in $u_{L-l}^{k}$ denotes a power, $u_{L-l}^{k}\equiv(u_{L-l})^{k}$).
Thus, at each stage a scalar kernel renormalization appears, $u_{L-l}$,
summarizing the effect of the downstream layers that have been integrated
out. This order parameter obeys the mean field equation

\begin{equation}
1-\sigma^{-2}u_{L-l}=\alpha(1-u_{L-l}^{-l}r_{L-l})\label{eq:u_L-l}
\end{equation}
the solution of which depends on the remaining weights $W'$ and the
task, through the layerwise mean squared readout, Eq.\ref{eq:b_l}.

Finally, integrating out all the weights yields an equation for the
network scalar renormalization factor $u_{0}$,

\begin{equation}
1-\sigma^{-2}u_{0}=\alpha(1-u_{0}^{-L}r_{0})\label{eq:u0}
\end{equation}
with the \textbf{input layer's} \textbf{mean squared readout}

\begin{equation}
r_{0}=\frac{1}{P}Y^{\top}K_{0}^{-1}Y\label{eq:b0}
\end{equation}
Here $K_{0}=\frac{\sigma^{2}}{N_{0}}X^{\top}X$ is the input kernel,
where $X$ is the $N\times P$ input data matrix.

Standard techniques using the partition function as a generating functional
allow for the derivation of important statistics of the system, in
particular its generalization performance. While the statistics of
the performance of the system are evaluated by completing the integration
over all weights, i.e., $l=L$ as in Eqs.\ref{eq:u0},\ref{eq:b0}
above, the results of partial weight integration are important in
evaluating the properties of the representations in individual layers
(Section $\text{\ref{sec:Changes-in-Representations}}$).

\textbf{Comparison with the Gaussian Process (GP) theory: }We will
compare our results to that of the GP theory (\citep{lee2017deep})
for infinitely wide networks. In the GP theory, the kernels $K_{l}$,
Eq.\ref{eq:layerkernels}, are \emph{self-averaged} and furthermore
the weight distribution is Gaussian so that the weight dependent kernels
can be replaced by their average over Gaussian weights (with variances
$\sigma^{2}/N$, or $\sigma^{2}/N_{0}$ for the first layer). For
a linear network this amounts to having the kernel of layer $l$ being
simply $\sigma^{2}$ times the kernel of layer $l-1$, hence, $K_{l}=\sigma^{2l}K_{0}.$

\textbf{Interpretation of order parameters: }Importantly, unlike the
predictions of the GP theory, we will show below that the statistics
of the kernel matrices induced by learning are complex and the relation
between kernel statistics at one stage of integration and the next
cannot be fully captured by a simple renormalization of the entire
matrix by a scalar factor. Instead, different statistics change differently
upon integrating the degrees of freedom. 

Nevertheless, several quantities depending on the kernel do undergo
a simple renormalization. In particular, the layerwise mean squared
readout, $r_{l}$, undergoes a simple scaling under weight averaging,
i.e., 
\begin{equation}
r_{l-1}=u_{l-1}\left\langle r_{l}\right\rangle _{l}\label{eq:u vs r-1}
\end{equation}
for all $1\leq l\leq L$. The subscript in $\left\langle \cdot\right\rangle _{l}$
denotes averaging over the upstream weights from $l$ to $L$ so that
$l-1$ is the top unintegrated layer. %

Likewise, upon successive integrations of all weights, we have $r_{0}=u_{0}^{l}\left\langle r_{l}\right\rangle $.

 This relation is important as it provides an operational definition
of the order parameters $u_{l}$ which can be used for their direct
evaluation in numerical simulations. Also, we will show below that
the mean and variance of the network predictor transform simply by
renormalizing the associated kernels with $u_{0}$ (see Eqs.$\text{\ref{eq:varfiteration}\ref{eq:meanfiteration}}$).

\textbf{Dependence on the size of data:} As can be seen from Eq.\ref{eq:u0}
our results hold when the input kernel is full rank, which implies
$\alpha_{0}=P/N_{0}<1$. This condition is understandable, since for
$\alpha_{0}>1$ there is no $W$ that achieves zero training error
(in the linear networks). We denote $\alpha_{0}=1$ \emph{as the interpolation
threshold }of our network (below which the training data can be exactly
matched). This threshold holds for generic input (i.e., such that
the rank of $K_{0}$ is $\min(P,N_{0})$) and for a target function
that is not perfectly realized by a linear input-output mapping (otherwise
zero error can be achieved for all $\alpha_{0}$). In most of our
work we will focus on the properties of zero error solution space,
i.e., we will assume $\alpha_{0}<1$ .

\subsection{\label{subsec:BPKR-for-narrow}BPKR for narrow architectures: }

As stated above, the BPKR at finite temperatures is well defined for
all $\alpha$. However, the zero-temperature limit is subtle when
$\alpha>1$ since the $P\times P$ kernel matrices, Eq.\ref{eq:layerkernels},
of the hidden layers, are of rank $N<P$, while we have assumed above
that the kernel matrices are invertible. Indeed, the above results
Eqs.$\text{\ref{eq:H_L-1}}$-$\text{\ref{eq:u_L-l}}$,$\text{\ref{eq:u vs r-1}}$
hold only for $\alpha<1$.

This difference between wide and narrow architectures reflects the
difference in the impact of learning on $W$ in the two regimes. While
in the wide regime, even for generic untrained $W$, the training
data can be perfectly fit by an appropriate choice of readout weights
$a$, in the narrow regime, a perfect learning of the task cannot
be achieved without an appropriate modification of $W$. Specifically,
at every stage of the integration, after averaging out the weights
upstream to the $l$-th layer, the remaining weights must ensure that
$Y$ is in the $N$-dimensional subspace of $\mathbb{R}^{P}$ spanned
by the $N$ ($P$-dimensional) vectors $X_{i,l},i=1,...,N$, induced
by the $P$ training inputs. 

As shown in Appendix $\text{\ref{subsec:appendixa}}$, these constraints
lead to replacement for Eq.\ref{eq:u_L-l} by

\begin{equation}
u_{L-l}^{l+1}=\alpha\sigma^{2}r_{L-l}\label{eq:u_L-l-2-2}
\end{equation}
where the layer mean squared readout at zero temperature, $r_{L-l}$,
is given by

\begin{equation}
r_{L-l}=\frac{1}{P}Y^{\top}K_{L-l}^{+}Y\label{eq:b_l-1-2}
\end{equation}
for $1\leq l<L$. Here $K_{L-l}$ is the kernel of the $L-l$-th layer
for a set of weights $W_{L-l}$ that yields zero training error, and
$K_{L-l}^{+}$ denotes the pseudo-inverse of $K_{L-l}$. In addition,
the scaling relationship, Eq.$\text{\ref{eq:u vs r-1}}$ which provides
an operational definition of the kernel renormalization OPs, still
holds for $1\leq l<L$. However, for the average of $r_{l}$ over
\emph{all} hidden weights, the relation with $r_{0}$ is given by
\begin{equation}
\langle r_{l}\rangle=u_{0}^{-l}r_{0}-1+\frac{1}{\alpha}\label{eq:opinterpretation}
\end{equation}
where $r_{0}$ is given by Eq.\ref{eq:b0} (see Appendix $\text{\ref{subsec:appendixa}}$
and SM IA for details). Thus, $\langle r_{l}\rangle$ has a cusp as
a function of $\alpha$ at $\alpha=1$ (see SM Fig.1).

Importantly, Eqs.$\text{\ref{eq:u0}}$,$\text{\ref{eq:b0}}$ between
$u_{0}$ and $r_{0}$ hold for $0\leq\alpha<\infty$, as $K_{0}$
is full rank as long as $\alpha_{0}<1$. Hence, many important system
properties such as the generalization error and the predictor statistics
which depend on $\alpha$ through $u_{0}$ , are \emph{smooth} functions
of $\alpha$ for all $\alpha$ (Section $\text{\ref{sec:Generalization}}$).

\subsection{\label{subsec:predictor-statistics}Predictor statistics: }

The generalization performance is closely related to the learning-induced
statistics of the predictor, Eq.\ref{eq:predictor}, for a new input
vector, $x$. First, we note that when the hidden-layer weights $W$
are fixed, the predictor $f(x)$ obeys Gaussian statistics (from the
fluctuations in the readout weights $a$) with

\begin{equation}
\langle f(x)\rangle_{a}=k_{L}^{\top}(x_{L})K_{L}^{-1}Y\label{eq:w fixed meanf}
\end{equation}
\begin{equation}
\langle\left(\delta f(x)\right)^{2}\rangle_{a}=K(x_{L},x_{L})-k_{L}^{\top}(x_{L})K_{L}^{-1}k_{L}(x_{L})\label{eq:w fixed varf}
\end{equation}
where $x_{L}$ is the vector of top-layer activations in response
to the new input $x$; $k_{L}(x_{L})$ is $P\times1$ vector given
by $k_{L}^{\mu}(x_{L})=K(x_{L},x_{L}^{\mu})$ where for any two vectors
$x,y$,

\begin{equation}
K(x,y)=N^{-1}\sigma^{2}x^{\top}y
\end{equation}
The subscript $a$ in Eqs.\ref{eq:w fixed meanf},\ref{eq:w fixed varf}
denotes averaging w.r.t. $a$ only. Thus the moments of the predictor
depend on $W$ through the $P\times P$ kernel matrix $K_{L}$ and
through $x_{L}$ and $x_{L}^{\mu}$. Evaluating its first two moments
w.r.t the full averaging (over $\Theta$) we find (Appendix $\text{\ref{subsec:Generalization}}$)

\begin{equation}
\langle f(x)\rangle=k_{0}^{\top}(x)K_{0}^{-1}Y\label{eq:meanfiteration}
\end{equation}
where $k_{0}$ is $P\times1$ vector given by $k_{0}^{\mu}=K_{0}(x,x^{\mu})$
where for any two input vectors $x,y$, $K_{0}(x,y)=N_{0}^{-1}\sigma^{2}x^{\top}y$.
Thus, at zero temperature, the mean predictor is independent of network
archiecture or noise level $\sigma$ and retains its value predicted
by the GP limit. This makes sense as at zero temperature multiplying
the kernels in the numerator and denominator by a scalar cancels out.
The variance of the predictor takes into account the $W$-average
of Eq.\ref{eq:w fixed varf}, which is the mean contribution from
the fluctuations in $a$ as well as the variance of the conditioned
mean Eq.\ref{eq:w fixed meanf}. These contributions produce the following
simple result,

\begin{equation}
\langle\left(\delta f(x)\right)^{2}\rangle=u_{0}^{L}\left(K_{0}(x,x)-k_{0}^{\top}(x)K_{0}^{-1}k_{0}(x)\right)\label{eq:varfiteration}
\end{equation}
(Appendix $\text{\ref{subsec:Generalization}}$). Thus, the predictor
variance equals the variance of the $L=0$ network (Eq.\ref{eq:w fixed varf}
for $L=0$) scaled by the kernel renormalization factor $u_{0}^{L}$,
which makes sense since the variance scales linearly with the kernel.
This variance renormalization due to the presence of hidden layers
has an important impact on the generalization error, which depends
on both moments of $f$, and can be written as $\varepsilon_{g}(x)=\langle(f(x)-y(x))^{2}\rangle=(\langle f(x)\rangle-y(x))^{2}+\langle\left(\delta f(x)\right)^{2}\rangle$
where $y(x)$ is the target label of the new input $x$.

The network properties after integrating the weights of all hidden
layers depends on the kernel renormalization factor $u_{0}$ but not
the intermediate renormalization factors $u_{l}$ ($1\leq l<L$).
We summarize below several important expressions for the renormalization
factor $u_{0}$, and the equations for the predictor statistics that
depend on $u_{0}$ that holds for $0\leq\alpha<\infty$. These results
are the main conclusions from this section, and they will be useful
for analyzing the generalization performance and how it depends on
various network parameters in Section $\text{\ref{sec:Generalization}}$.%

\fbox{\begin{minipage}[t]{1\columnwidth} 
Kernel renormalization factor $u_{0}$:

$u_{0}$ relates the average of the top-layer mean squared readout
$r_{L}$ (over all $L$ weight matrices) to the input-layer mean squared
readout $r_{0}$ 

\begin{equation}
r_{L}=\frac{1}{P}Y^{\top}K_{L}^{+}Y
\end{equation}

and

\begin{equation}
r_{0}=\frac{1}{P}Y^{\top}K_{0}^{-1}Y
\end{equation}

through

\begin{equation}
r_{0}=u_{0}^{L}\left[\langle r_{L}\rangle-\max(1-\frac{1}{\alpha},0)\right]\label{eq:opdefu0-sum}
\end{equation}

The matrix $K_{L}^{+}$ is the (psuedo) inverse of the top-layer representation,
$K_{L}=\frac{\sigma^{2}}{N}X_{L}^{\top}X_{L}$, and $K_{0}$ is the
input-layer kernel . 

The predictor statistics for an input $x$ is:
\begin{align}
\langle f(x)\rangle & =k_{0}^{\top}(x)K_{0}^{-1}Y\label{eq:mean-sum}\\
\langle\left(\delta f(x)\right)^{2}\rangle & =u_{0}^{L}\left(K_{0}(x,x)-k_{0}^{\top}(x)K_{0}^{-1}k_{0}(x)\right)\label{eq:var-sum}
\end{align}

where $k_{0}(x)=K_{0}(x,X_{0})$ and $X_{0}$ stands for the $P$
training vectors. 

The self-consistent equation for $u_{0}$ is: 

\begin{equation}
1-\sigma^{-2}u_{0}=\alpha(1-u_{0}^{-L}r_{0})\label{eq:u0-sum}
\end{equation}
\end{minipage}}

\subsection{Qualitative differences between wide and narrow architectures}

\begin{figure}
\includegraphics[width=1\columnwidth]{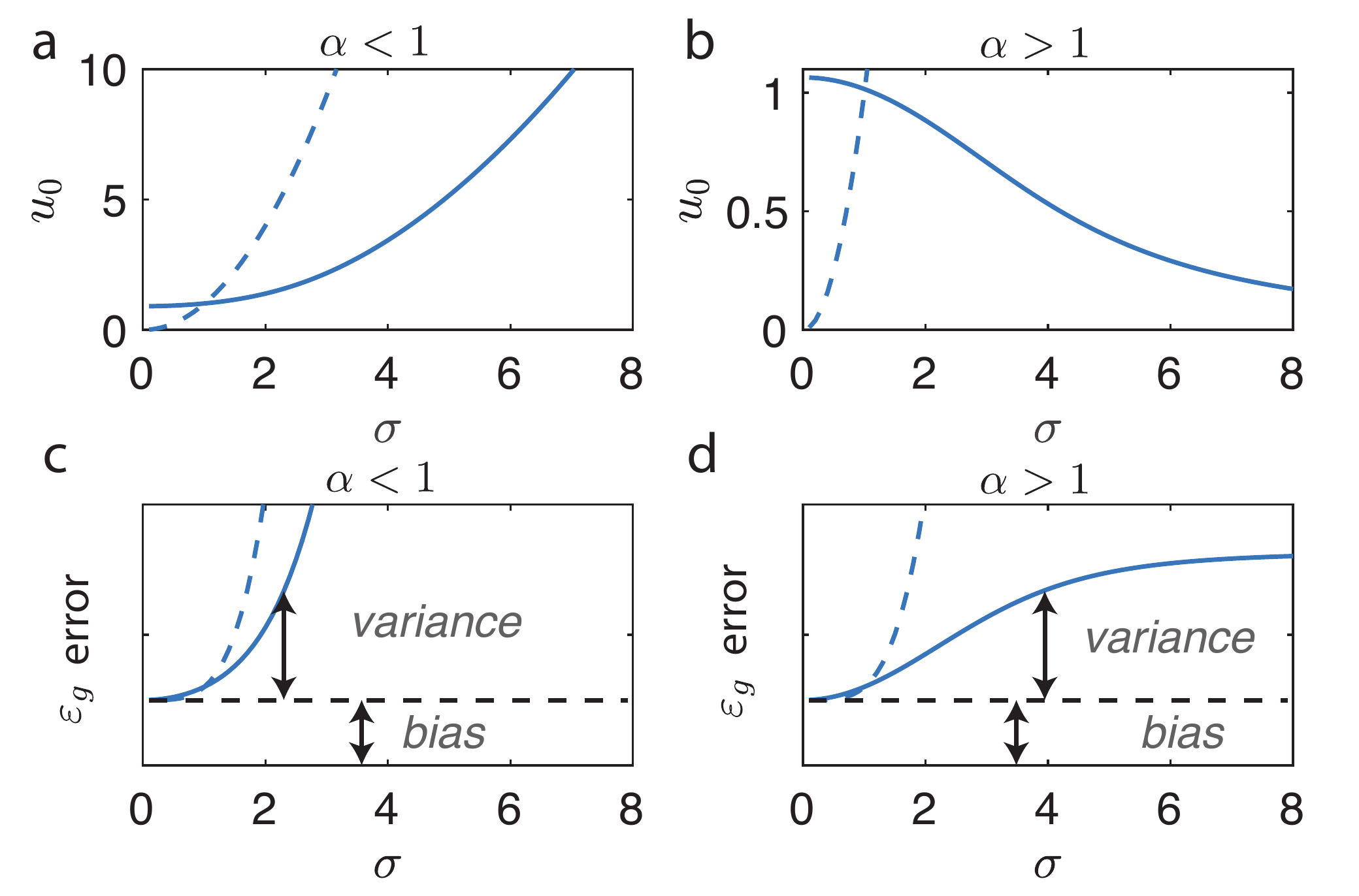}

\caption{\label{fig:u0sig}(a,b) Dependence of the order parameter $u_{0}$
on the noise parameter $\sigma$ in wide (a ; $\alpha=0.8$) and narrow
networks (b; $\alpha=1.1$). Blue lines: theory. Blue dashed lines:
the prediction of GP theory ($u_{0}=\sigma^{2})$. In both (a) and
(b) $u_{0}$ is finite for low $\sigma$. Additionally, in (a) ( $\alpha<1$)
$u_{0}$ diverges as $\sigma^{2}(1-\alpha)$ for $\sigma\rightarrow\infty$
, slower than in the GP theory . In (b), for $\alpha>1$, $u_{0}$
vanishes as $\sigma\rightarrow\infty$, drastically different from
the GP. (c,d) Dependence of the generalization error on $\sigma$
for wide ($\alpha=0.8$) and wide $(\alpha=1.1$) regimes. The change
with $\sigma$ is due to the change in variance, which scales as $\sigma^{2}u_{0}$
(Eq.\ref{eq:var-sum}). The bias contribution is independent of $\sigma$,
Eq.\ref{eq:mean-sum} and black dashed lines. (c) The generalization
error diverges slower than in GP theory for $\alpha<1$ as $\sigma\rightarrow\infty$.
(d) The generalization error increases and approaches a finite limit
as $\sigma\rightarrow\infty$ for $\alpha>1$, in stark contrast to
the divergence predicted by the GP theory. }
\end{figure}

To highlight the qualitative differences between wide and narrow architectures,
we use the equation for $u_{0}$ given by Eq.$\text{\ref{eq:u0-sum}}$
and show in Fig.$\text{\ref{fig:u0sig}}$(a,b) $u_{0}$ vs. $\sigma$
for different $\alpha$ (see SM $\mathrm{IIA}$). Note because $K_{0}$
scales with $\sigma^{2}$, when we vary $\sigma$ we hold $\sigma^{2}r_{0}$
constant. The limit of infinite width corresponds to $\alpha\rightarrow0$.
In this limit Eq.\ref{eq:u0-sum} yields $u_{0}=\sigma^{2}$ which
is the prediction of the GP theory for a linear network. First, in
contrast to the GP theory, for finite $\alpha$, $u_{0}$ attains
a nonzero value for $\sigma\rightarrow0$. Furthermore, the dependence
of $u_{0}$ on $\sigma$ is qualitatively different in the wide and
narrow regimes. For $\alpha<1,$ $u_{0}$ increases monotonically
with $\sigma$, diverging for large $\sigma$, $u_{0}\rightarrow\sigma^{2}(1-\alpha)$.
Importantly, the behavior is reversed for $\alpha>1$ . Here $u_{0}$
decreases monotonically with $\sigma$ and vanishes for large $\sigma$
as $u_{0}^{L}\rightarrow\sigma^{-2}(\alpha\sigma^{2}r_{0}/(\alpha-1)$).
This difference in behavior particularly for large $\sigma$, reflects
the differences in the effect of learning on the weight space, as
discussed at the beginning of Section $\text{\ref{subsec:BPKR-for-narrow}}$.
Learning imposes more constraints on $W$ in the narrow regime and
only a small fraction of $W$ have nonzero Gibbs probability, deviating
strongly from the predictions of the GP limit.

\section{\label{sec:Generalization}Generalization}

In a linear network, the mapping between input and output is given
by an $N_{0}$-dimensional effective weight vector $\sim a^{\top}W_{L}W_{L-1}\cdots W_{1}$.
As mentioned above, we here assume the system is below the interpolation
threshold, i.e., $\alpha_{0}=P/N_{0}<1$, hence our network learns
perfectly the training input-output relations as $T\rightarrow0$,
even without hidden layers (i.e., $L=0$). Thus, our deep network
(with $L\geq1$ and $\alpha=P/N$ of $\mathcal{O}(1)$) is always
in the heavily over-parameterized regime, where the number of modifiable
parameters is much larger than the number of parameters needed to
satisfy the training data. Naively, this would imply that the system
is extremely poor in generalization. However, as we will show, this
is not necessarily so due to the presence of 'inductive bias' in the
form of $L_{2}$-regularization. In this section, we will discuss
how the generalization error depends on various network parameters
including the noise parameter $\sigma$, the network width $N$, and
the network depth $L$, which may provide helpful insights for selecting
network parameters during training.

\subsection{Dependence of generalization on noise}

From the predictor statistics Eqs.$\text{\ref{eq:var-sum},\ref{eq:mean-sum}}$
we conclude that the contribution of the squared bias $(\langle f(x)\rangle-y(x))^{2}$
to $\varepsilon_{g}$ is constant, independent of the network parameters,
$N$, $L$, and $\sigma$. As for the contribution from the variance,
this tracks the behavior of $u_{0}^{L}\sigma^{2}$ (the factor $\sigma^{2}$
stems from the noise dependence of the kernels). Thus, from our previous
analysis of $u_{0}$ in Section $\text{\ref{subsec:BPKR-for-narrow}}$
we can predict the generalization error's dependence on the noise
as shown schematically in Fig.\ref{fig:u0sig}(c) for wide and in
Fig.$\text{\ref{fig:u0sig}}$(d) for narrow networks. In both regimes,
the variance grows monotonically with noise, but for large noise,
in the narrow regime, the generalization error does not diverge but
saturates to a finite value $\alpha r_{0}\sigma^{2}/(\alpha-1)$ ($r_{0}$
scales as $\sigma^{-2}$), while in the wide regime the generalization
error diverges as $\sigma^{4}(1-\alpha)$. (See SM $\mathrm{IIA}$)

\subsection{Dependence of generalization on width. }

\begin{figure*}
\includegraphics[width=1\textwidth]{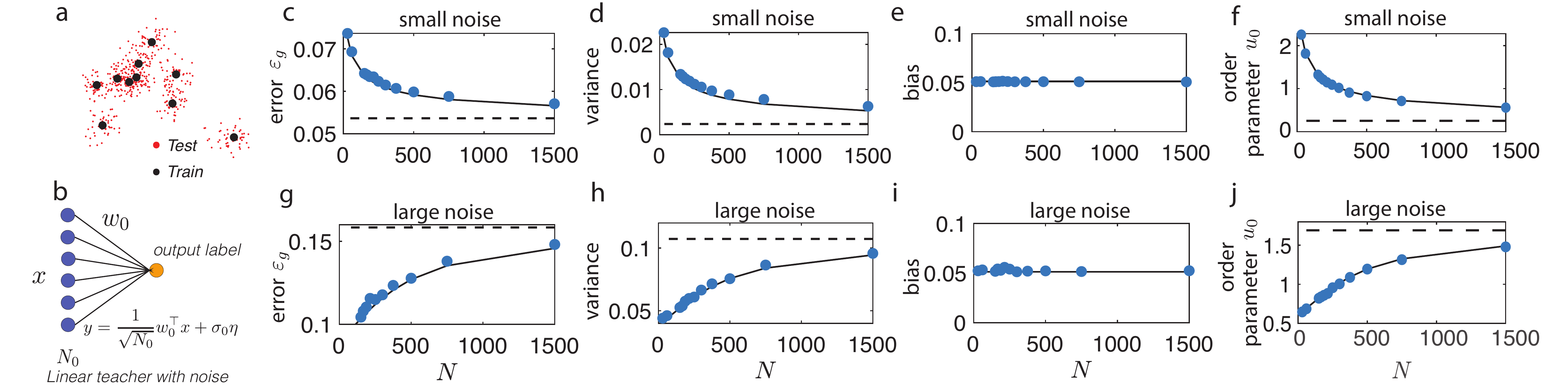}

\caption{\label{fig:egvsN}Dependence of network behavior on hidden-layer width
$N$. (a) Schematics of inputs in the template model: $N_{0}$ dimensional
vectors clustered around $P$ templates (shown as black dots) which
are used as training vectors. The testing data (shown as red dots)
are sampled from the $P$ clusters by adding Gaussian noise to the
templates (see Appendix $\text{\ref{a)-Template-model}}$). (b) Target
labels are outputs of a single-layer linear teacher network with weight
vector $w_{0}$ and additive output noise. (c)-(j) Behavior of a linear
network with a single hidden layer $(L=1)$ vs. hidden-layer size
(width) $N$; generalization error (c,g), variance (d,h) and bias
(e,i) of the predictor averaged over the test data, and normalized
by the amplitude of the labels, and the order parameter $u_{0}$ (f,j),
we performed the same averaging and normalization for all the following
results. Black lines: theory. Blue dots: simulation. Black dashed
lines: GP limit ($N=\infty)$. Top row: parameter regime with small
noise where the generalization error increases with $\alpha$ (i.e.,,
decreases with $N$). Bottom row: parameter regime with large noise
where the generalization error decreases with $\alpha$ (i.e., increases
with $N$). (See detailed parameters for the simulation in Appendix
$\text{\ref{a)-Template-model}}$)}
\end{figure*}

We now consider in detail the dependence of $\varepsilon_{g}$ on
$\alpha$ for different levels of noise. A detailed analysis (SM ${\rm IIB}$)
shows that when other parameters are fixed, the generalization error
varies monotonically with width, increasing if

\begin{equation}
\sigma^{2(L+1)}>\sigma^{2}r_{0}
\end{equation}
Otherwise, it decreases with width. The latter case is an example
where, despite increasing model complexity through increasing $N$,
generalization performance improves, as shown in Fig.$\text{\ref{fig:egvsN}}$.
In the example shown, we use normally distributed training input vectors
and training labels $Y$ generated by a noisy linear teacher. The
generalization error is measured here on the network outputs generated
by inputs which are corrupted versions of the training vectors (detailed
in Appendix $\text{\ref{a)-Template-model}}$). Thus, this example
corresponds to the case where the training data plays the role of
$P$ templates and the test inputs are sampled from Gaussian noise
around these templates (\citep{babadi2014sparseness}). The model
introduces a relation between target outputs and data statistics,
which is a more realistic situation than a ‘vanilla’ normally distributed
test data, where there is no inherent relation between inputs and
outputs (aside from weak correlation induced by the random teacher
weights). We emphasize that although we chose a specific example to
present our numerics, our theory is not limited to any specific type
of input-output distribution.

\subsection{Dependence of generalization on depth }

First, we discuss the limit of large $L$, analyzing the fixed point
of Eq.\ref{eq:u0}, i.e., the solution for $u_{0}(L\rightarrow\infty).$

We recall that in the GP limit $u_{0}(L)=u_{0}=\sigma^{2L}$, hence
if $\sigma<1$, $u_{0}\rightarrow0$, and the entire deep network
collapses to a single-layer network, or if $\sigma>1$, $u_{0}$ diverges.
Thus, in order to obtain a non-trivial behavior the noise needs to
be fine tuned to $\sigma=1$ . Similar fine-tuning is required, in
the GP theory, in nonlinear network in the limit of large $L$ . As
we will show below, the behavior of our networks is strinkingly different.
In fact, we will show that in the low-noise regime, the system is
self-tuned in that $u_{0}$ approaches a finite fixed point value.

Specifically, Eq. \ref{eq:u0} predicts that in the low-noise regime,
defined by

\begin{equation}
\sigma^{2}(1-\alpha)<1\label{eq:lownoise}
\end{equation}
$u_{0}\rightarrow u_{\infty}=1$, independent of $\sigma$. Furthermore,
the approach to this limit is inversely proportional to $L$, $u_{0}\approx1-\frac{v_{0}}{L}$,
where the pre-factor $v_{0}$ obeys

\begin{equation}
\exp-v_{0}=\frac{\alpha\sigma^{2}r_{0}}{(1-\sigma^{2}(1-\alpha))}\label{eq:ev0}
\end{equation}
Thus, there are two sub-regimes. If $\alpha\sigma^{2}r_{0}<1-\sigma^{2}(1-\alpha)$,
then $v_{0}>0$, implying that $u_{0}(L)<1$ and \emph{increases}
with\textbf{ $L$} toward its fixed-point value $1$. If this inequality
does not hold, $u_{0}$\textbf{ }\emph{decreases} with\textbf{ $L$}
towards its fixed point. Note that narrow networks in which $\alpha>1$,
are always in the low-noise regime given by Eq.\ref{eq:lownoise},
for all values of $\sigma$.

\begin{figure*}
\includegraphics[width=0.9\textwidth]{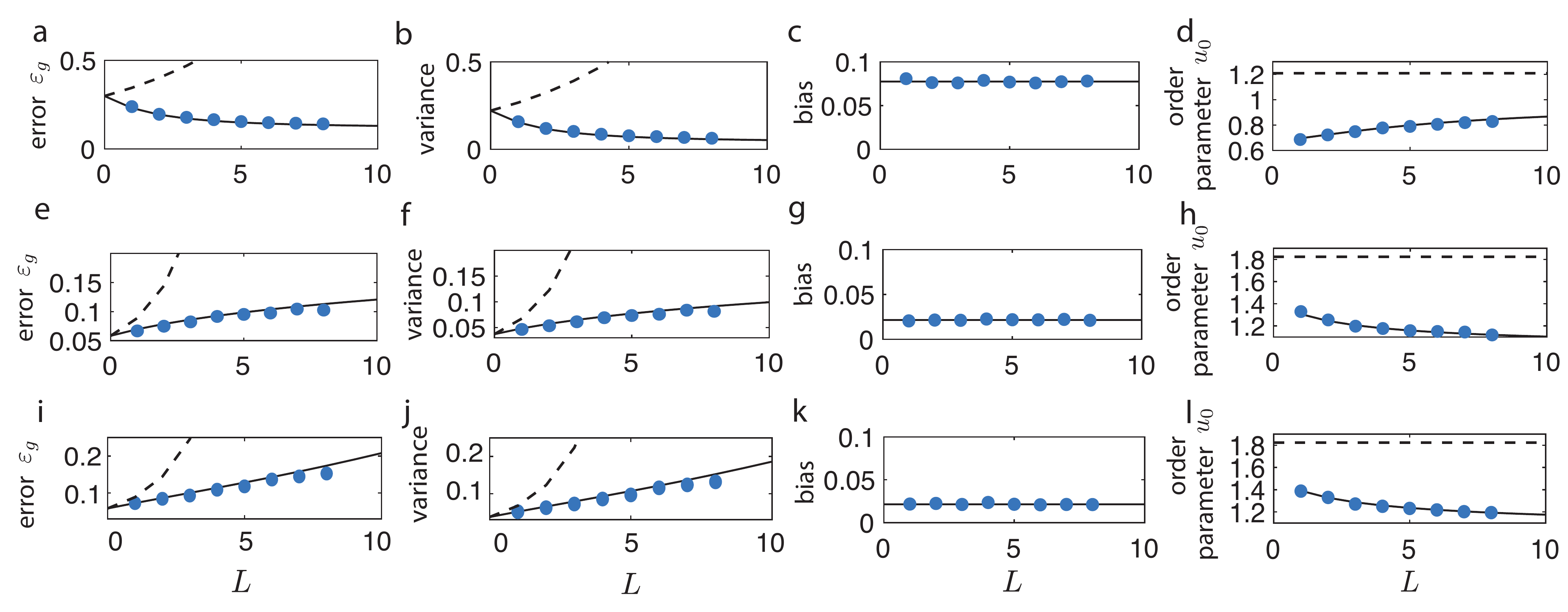}

\caption{\label{fig:egvsL}Dependence of the generalization error on network
depth $L$. Generalization error (a,e,i), variance (b,f,j), bias of
the predictor (c,g,k), and the order parameter $u_{0}$ (d,h,l) as
a function of $L$. Black lines: theory. Blue dots: simulation. Black
dashed lines: the GP limit ($N\rightarrow\infty$). (Details of the
example model and parameters are described in Appendix $\text{\ref{a)-Template-model}}$)
(a)-(d) The sub-regime where the generalization error decreases with
$L$. (e)-(h) The sub-regime where the generalization error increases
with $L$ approaching a finite limit. (i)-(l) The high-noise regime
where the generalization error increases with $L$ and diverges as
$L\rightarrow\infty$. }
\end{figure*}

\begin{figure}
\includegraphics[width=1\columnwidth]{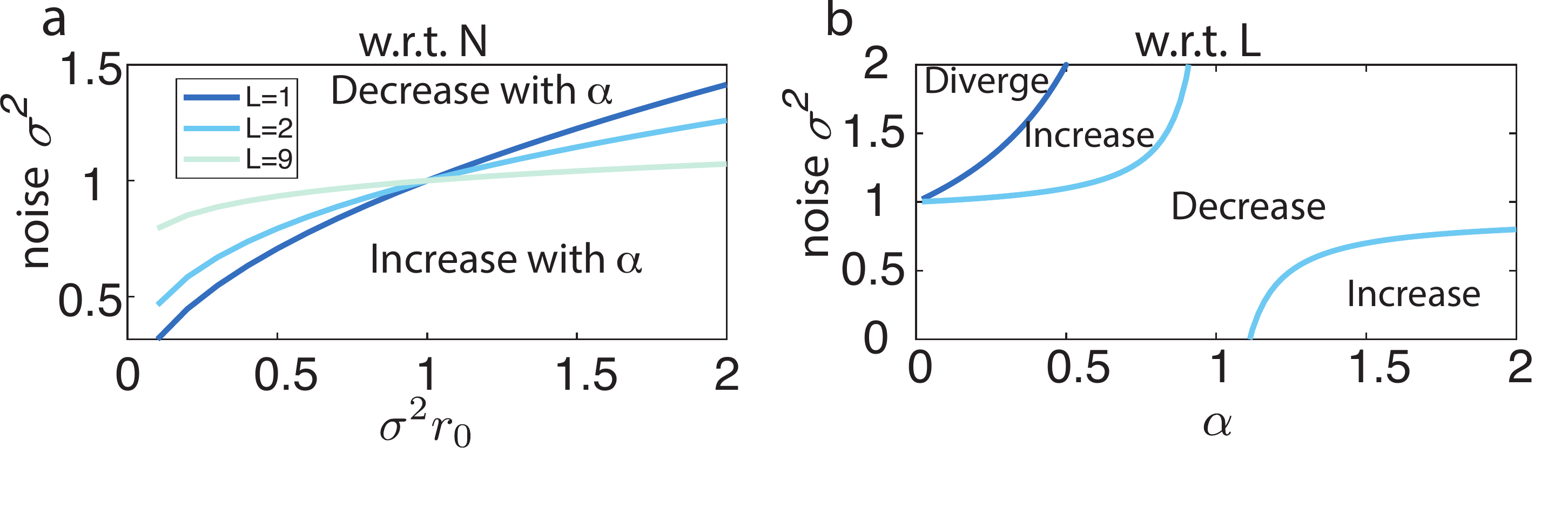}

\caption{\label{fig:summaryplot}A summary plot showing different parameter
regimes for the dependence of the generalization error on width ($N$)
and depth ($L$). (a) The plane of noise $\sigma^{2}$ and $\sigma^{2}r_{0}$
(the input readout parameter, $r_{0}$ is normalized by $\sigma^{2}$
so that $\sigma^{2}r_{0}$ is independent of $\sigma$). Above the
boundary lines the generalization error decreases with $\alpha$,
below it the generalization error increases with $\alpha$. The three
different curves are for $L=1$, $L=2$ and $L=9$. The separating
line becomes flatter as $L$ increases. (b) For a fixed $\sigma^{2}r_{0}$,
in the plane of $\alpha$ and $\sigma^{2}$, there are three types
of behaviors of $\varepsilon_{g}$ vs. $L$, it can decrease, increase
or diverge. Note that in the narrow regime ($\alpha>1$) $\varepsilon_{g}$
never diverges, consistent with Eq.$\text{\ref{eq:lownoise}}$. }
\end{figure}

In contrast, in the high-noise regime $\sigma^{2}(1-\alpha)>1$, the
fixed-point value is $u_{0}\rightarrow u_{\infty}=\sigma^{2}(1-\alpha)>1$.
These results have important implications for the predictor variance
and the generalization error. Inspecting Eq.\ref{eq:varfiteration}
we conclude that in the low-noise regime, the predictor variance and
the generalization error reach a finite value, since $u_{0}^{L}\rightarrow\exp-v_{0}$,
which depends on $\alpha$, $\sigma,$and $r_{0}$. On the other hand,
in the high-noise limit $u_{0}^{L}$ diverges exponentially with $L$,
yielding a divergent generalization error. Further analysis shows
that when all other parameters are held fixed, the generalization
error is monotonic with $L$ (SM ${\rm IIC}$). In the low-noise regime,
it saturates to a finite value, decreasing when $\nu_{0}>0$ (because
$u_{0}^{L}$ is inversely related to $u_{0}$) and increasing otherwise.
These three behaviors are illustrated in Fig.$\text{\ref{fig:egvsL}}$
for the same ‘template’ noisy linear teacher model as illustrated
in Fig.$\text{\ref{fig:egvsN}}$.
The phase diagram of the generalization error depicting its different
behaviors is shown in Fig.$\text{\ref{fig:summaryplot}}$. In Fig.$\text{\ref{fig:summaryplot}}$(a)
we show the trend w.r.t. the width (equivalently $\alpha$) in the
plane of noise ($\sigma^{2})$ and input mean squared readout ($\sigma^{2}r_{0})$
(scaled so that it is independent of noise), where the boundaries
are given by $\sigma^{2(L+1)}=\sigma^{2}r_{0}$. In Fig.$\text{\ref{fig:summaryplot}}$(b)
we show the behaviors w.r.t. the depth in the plane of $\sigma^{2}$
and $\alpha$ (for $\sigma^{2}r_{0}=0.8$). Note that in the GP limit
($\alpha\rightarrow0$) the behavior of the generalization error w.r.t.
$L$ is either decreasing and goes to 0 or divergent

\subsection{Varying the size of the training set}

Until now we have considered the dependence of $\varepsilon_{g}$
on network parameters for a fixed training set, in particular fixed
training set size $P$. Here we consider the effect of varying $P$.
In addition to varying $\alpha$, $\alpha_{0}$, changes in the training
set affect $r_{0}$, as well as the kernels appearing in Eqs.\ref{eq:meanfiteration},\ref{eq:varfiteration}.
The exact effect of changing $P$ depends, of course, on the details
of the input and output data. Here we address what happens near and
above the interpolation threshold, $\alpha_{0}=1$. For $\alpha_{0}>1$
, the minimal training error is nonzero. However, there is a huge
degeneracy of weights that minimize this error, defined by all values
of $\Theta$ that yield the same input-output linear mapping, given
by input output effective weights (see SM ${\rm IID}$) that obey

\begin{equation}\label{eq:pseudo-inverse}
\frac{1}{\sqrt{N_{0}}}\frac{1}{\sqrt{N^{L}}}a^{\top}W_{L}W_{L-1}\cdots W_{1}=(XX^{\top})^{-1}XY
\end{equation}

This implies that the predictor is uniquely given by

\begin{equation}
f(x)=x^{\top}(XX^{\top})^{-1}XY
\end{equation}
for all $x$ (whether belonging to the training set or not). Hence
the training and the generalization error in the deep network is identical
to that of a single-layer network. The generalization error is given
by the bias component of the error, \emph{as the predictor variance
is zero. }

The singularity of the input kernel at $\alpha_{0}=1$ gives rise
to a simple example of a ‘double descent’ in the generalization error.
The divergence at the interpolation threshold is known to be suppressed
by the addition of $L_{2}$ regularizers \citep{advani2017high}.
The reason for the persistence of this divergence in our theory is
the fact that our $L_{2}$ regularization term is scaled by the temperature
$T$, see Eq.$\text{\ref{eq:Loss-1}}$, which means that at zero $T$
it does not lift the degeneracy of the solutions (and the concentration
of the solution space in large norm weights at $\alpha_{0}=1$). Indeed,
this degeneracy is halted, in our theory, only by finite temperature,
as shown in Section $\text{\ref{subsec:Finite-temperature}}$. 

\begin{figure*}
\includegraphics[width=0.9\textwidth]{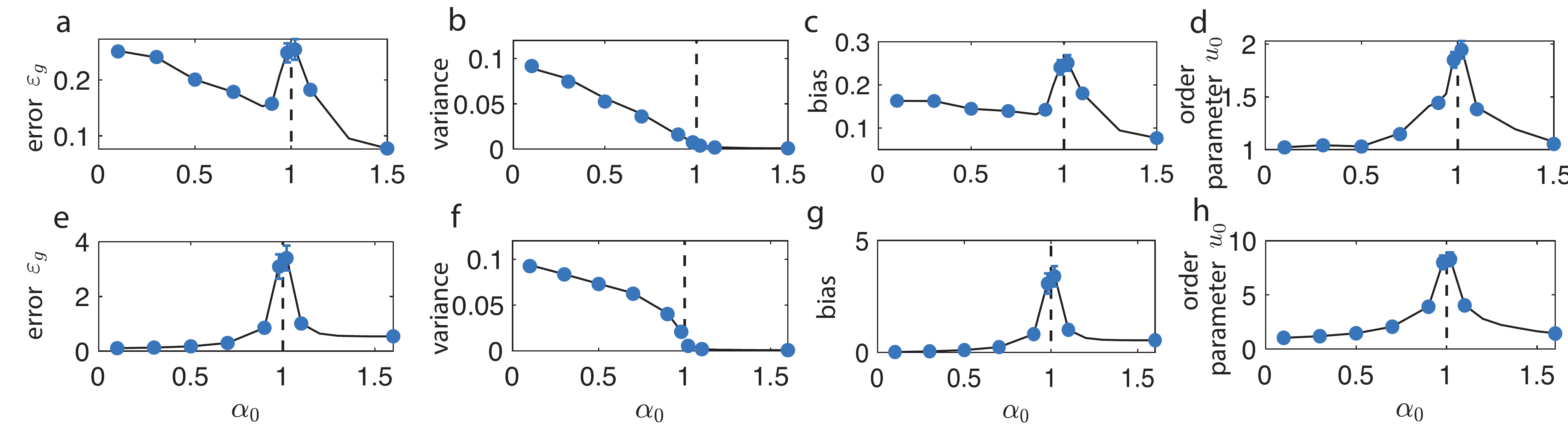}

\caption{\label{fig:egvsP}Dependence of the generalization error on the training
set size $P$ for a single hidden-layer ($L=1$) network. Generalization
error (a,e), variance (b,f), bias of the predictor (c,g) , and the
order parameter $u_{0}$ (d,h) as a function of $\alpha_{0}$. Note
that $u_{0}$, $\langle f(x)\rangle^{2}$ and $\varepsilon_{g}$ all
diverge as $\alpha_{0}\rightarrow1$ from both sides. The variance
of the predictor vanishes as $\alpha_{0}\rightarrow1^{-}$ and remains
zero for $\alpha_{0}>1$. Black lines: theory. Black dashed lines:
the interpolation threshold at $\alpha_{0}=1$. Blue dots with error
bars: simulations, error bars show the s.t.d. of the mean across different
realizations of the noise in the training data labels. (a)-(d) The
network is trained on the ‘template’ model and the noisy linear teacher
task with parameters described in Appendix $\text{\ref{a)-Template-model}}$.
(e)-(h) The network is trained on the ‘template’ model with random
labeling of each cluster, with parameters described in Appendix $\text{\ref{a)-Template-model}}$.
{[}In this figure we use the finite T result (Section $\text{\ref{subsec:Finite-temperature}}$)
for the theory curves, since as $\alpha_{0}\rightarrow1$ increasingly
lower $T$ needs to be used if we want to approximate the zero $T$
limit (see details in Appendix $\text{\ref{3.-Finite-T}}$ and SM
IIE). {]}}
\end{figure*}

The singularity may affect drastically the behavior near it on both
sides of the interpolation threshold. For instance, if the input is
sampled from a standard Gaussian i.i.d. distribution then $r_{0}$
diverges when $\alpha_{0}\rightarrow1$ as $\propto|1-\alpha_{0}|^{-1}$,
leading to vanishing of the variance of the predictor (averaged over
the testing example) as $(1-\alpha_{0})^{1/L}$. The sample average
of the squared mean, $\langle f(x)\rangle^{2}$, diverges as $\propto|1-\alpha_{0}|^{-1}$
(see Appendix $\text{\ref{subsec:Generalization}}$ for the assumptions
made here and additional analysis in SM ${\rm IID}$). Hence the generalization
error on both sides of $\alpha_{0}=1$ is dominated by the bias and
diverges as $|1-\alpha_{0}|^{-1}$. This non-monotonicity of $\varepsilon_{g}$
is reminiscent of \emph{double descent }\citep{mei2019generalization,belkin2019reconciling}.
However, genuine double descent, namely, non-monotonicity of $\varepsilon_{g}$
when increasing the \emph{number of} \emph{network parameters }for
a \emph{fixed training set}, does not occur in our system, since the
error is always monotonic with $\alpha$ (as shown in Fig.$\text{\ref{fig:egvsN}}$).
In Fig.$\text{\ref{fig:egvsP}}$, we show these results for the ‘template’
model where the inputs are clustered, with two rules for the labels:
a noisy linear teacher as in Fig.$\text{\ref{fig:egvsN}}$, and random
labels where the label for each cluster is binary and drawn randomly
(both detailed in Appendix $\text{\ref{a)-Template-model}}$).

For the noisy linear teacher task, the minimum generalization error
is achieved on the RHS of the interpolation threshold. Due to the
linearity of the task, we do not need a large number of parameters
(i.e., small $\alpha_{0}$) to generalize well. However, for the random
labeling task, the minimum generalization error is achieved on the
LHS of the interpolation threshold. Because of the nonlinearity of
the task itself, having $N_{0}>P$ is required for good performance.

\section{\label{sec:Extensions}Extensions}

\subsection{\label{subsec:Multiple-Outputs}Multiple outputs}

We now consider the case where there are $m>1$ linear outputs, with
$N\times m$ readout weight matrix $A$. The rest of the architecture
is the same as above with $L$ hidden layers of width $N$. Extending
our theory we obtain, instead of \emph{scalar} renormalization factors
$u_{l}$($m=1)$, $m\times m$ \emph{renormalization matrices}\textbf{
}$\mathcal{U}_{l}$ per layer (see details in Appendix $\text{\ref{sec:Multiple-Outputs}}$).
Here we focus on the zero-temperature limit. We first consider $\alpha<1$.
We define the \emph{hidden-layer} $m\times m$ readout covariance
matrix $\mathcal{R}_{l}=\frac{1}{P}Y^{\top}K_{l}^{-1}Y$. This matrix
is diagonalized via

\begin{equation}
\mathcal{R}_{l}=\frac{1}{P}Y^{\top}K_{l}^{-1}Y=V_{l}\mathrm{diag}(r_{1l},\cdots,r_{kl},\cdots,r_{ml})V_{l}^{\top}\label{eq:B0-2}
\end{equation}
Here $\mathrm{diag}(r_{1l},\cdots,r_{kl},\cdots,r_{ml})$ denotes
a diagonal matrix with components $\{r_{kl}\}_{k=1,..,m}$, and $V_{l}$
is the unitary matrix diagonalizing $\mathcal{R}_{l}$ and depends
on the specific realization of $\{W_{k}\}_{k<l+1}$. We find that
the matrix $\mathcal{U}_{l}$ is diagonalized by the same unitary
matrix $V_{l}$ , i.e., $\mathcal{U}_{l}=V_{l}\mathrm{diag}(u_{1l},\cdots,u_{kl},\cdots,u_{ml})V_{l}^{\top}$$.$
Each layer-wise renormalization factor $u_{kl}$ obeys the same equation
as single-output scalar renormalization 
\begin{equation}
1-\alpha=\sigma^{-2}u_{kl}-\alpha u_{kl}^{-(L-l)}r_{kl},1\leq k\leq m\label{eq:layerUl}
\end{equation}
Similarly, the matrix $\mathcal{U}_{0}$ is diagonalized by the unitary
matrix $V_{0}$ which diagonalizes the \emph{input-layer} $m\times m$
readout covariance matrix, 

\begin{equation}
\mathcal{R}_{0}=\frac{1}{P}Y^{\top}K_{0}^{-1}Y=V_{0}\mathrm{diag}(r_{10},\cdots,r_{k0},\cdots,r_{m0})V_{0}^{\top}\label{eq:R0}
\end{equation}

The renormalization matrix $\mathcal{U}_{0}$ obeys $\mathcal{U}_{0}=V_{0}\mathrm{diag}(u_{10},u_{20},\cdots,u_{ml})V_{0}^{\top}$
with $u_{k0}$ obeying similar equation as Eq.\ref{eq:u0}, i.e.,
\begin{equation}
1-\alpha=\sigma^{-2}u_{k0}-\alpha u_{k0}^{-L}r_{k0},1\leq k\leq m\label{eq:multipleU0}
\end{equation}
Similar to the single-output case, $\mathcal{R}_{l}=Y^{\top}K_{l}^{-1}Y$
undergoes a matrix product renormalization under weight averaging
(Appendix $\text{\ref{sec:Multiple-Outputs}}$ and SM IIIA), i.e.,
\begin{equation}
\mathcal{R}_{l-1}=\mathcal{U}_{l-1}\langle\mathcal{R}_{l}\rangle_{l}\label{eq:u vs r-1-1}
\end{equation}
The average of the $m$-dimensional vector $f(x)$ has the same form
as in the single-output case, $\langle f(x)\rangle=k_{0}^{\top}(x)K_{0}^{-1}Y$.
The covariance matrix of the predictor is 
\begin{multline}
\langle\delta f(x)\delta f(x)^{\top}\rangle=\mathcal{U}_{0}^{L}(K_{0}(x,x)-k_{0}^{\top}(x)K_{0}^{-1}k_{0}(x))\\
=V_{0}\mathrm{diag}(u_{10}^{L},\cdots,u_{k0}^{L},\cdots,u_{m0}^{L})V_{0}^{\top}\\
(K_{0}(x,x)-k_{0}^{\top}(x)K_{0}^{-1}k_{0}(x))
\end{multline}
Thus, the fluctuations in the predictor of each mode (eigenvectors
in $V_{0}$) are independent and are given as in the scalar output
case. However, the overall behavior may be different than in the $m=1$
case, since individual outputs typically consist of contributions
from multiple modes. For instance, the generalization error may not
be monotonic with either $\alpha$ or with $L$.

\textbf{Narrow architecture with multiple outputs:} Similar to BPKR
for the single output case, for $\alpha>1$ the zero-temperature limit
is affected by the singularity of the kernel matrices $K_{l}$. Eqs.$\text{\ref{eq:B0-2}},\text{\ref{eq:layerUl}},\text{\ref{eq:u vs r-1-1}}$
hold only for $\alpha<1$, and need to be modified for $\alpha>1$
as we introduce here. We find (Appendix $\text{\ref{sec:Multiple-Outputs}}$
and SM IIIB) that the renormalization matrices $\mathcal{U}_{l}$
are still diagonalized by the unitary matrix $V_{l}$ which diagonalizes
the $m\times m$ readout covariance readout matrices, $\mathcal{R}_{l},$
which for $\alpha>1$ are given as,

\begin{equation}
\mathcal{R}_{l}=\frac{1}{P}Y^{\top}K_{l}^{+}Y=V_{l}\mathrm{diag}(r_{1l},\cdots,r_{kl},\cdots,r_{ml})V_{l}^{\top}\label{eq:b_l-1-2-1}
\end{equation}
where $r_{kl}$'s are the eigenvalues of $\mathcal{R}_{l}$ and $K_{l}^{+}$
is the pseudo-inverse of $K_{l}$. The renormalization eigenvalues
$u_{kl}$'s of the matrix $\mathcal{U}_{l}$ are related to the eigenvalues
of $\mathcal{R}_{l}$ by 
\begin{equation}
u_{kl}^{L-l+1}=\alpha\sigma^{2}r_{kl}\label{eq:multiplenarrow}
\end{equation}
The relation given in Eq.$\text{\ref{eq:u vs r-1-1}}$, which provides
an operational definition of the matrix kernel renormalization OPs,
still holds for $1\leq l<L$. However, for the full average over weights,
it is replaced by 
\begin{align}
\langle\mathcal{R}_{l}\rangle & =\mathcal{U}_{0}^{-l}\mathcal{R}_{0}-(1-\frac{1}{\alpha})I\label{eq:narrowop}
\end{align}
(see Appendix $\text{\ref{sec:Multiple-Outputs}}$ and SM IIIC for
details). 

As in the single-output case, Eqs.$\ref{eq:R0}$,$\text{\ref{eq:multipleU0}}$
hold for $0\leq\alpha<\infty$, as $K_{0}$ is full rank as long as
$\alpha_{0}<1$. Hence, $\mathcal{U}_{0}$ and other quantities such
as the generalization error are a smooth function of $\alpha$ for
all $\alpha$ .

Similarly as for the single-output case in Section $\text{\ref{sec:The-Back-propagating-KR}}$,
the network properties after integrating the weights of all hidden
layers depends on the kernel renormalization matrix $\mathcal{U}_{0}$
but not the intermediate renormalization factors $\mathcal{U}_{l}$
($1\leq l<L$). We summarize below several important expressions for
the renormalization matrix $\mathcal{U}_{0}$, and the equations for
the predictor statistics that depend on $\mathcal{U}_{0}$. These
expressions hold for $0\leq\alpha<\infty$.

\fbox{\begin{minipage}[t]{1\columnwidth}
Kernel renormalization $mxm$ matrix $\mathcal{U}_{0}$ for a network
with $m$ outputs:

The renormalization matrix $\mathcal{U}_{0}$ relates the average
mean squared top-layer and the input-layer readout covariance matrices
via, 

\begin{equation}
\mathcal{R}_{0}=\mathcal{U}_{0}^{L}\left[\langle\mathcal{R}_{L}\rangle-\max(1-\frac{1}{\alpha},0)I\right]\label{eq:opdefu0-sum-mul}
\end{equation}

where $\mathcal{R}_{L}=\frac{1}{P}Y^{\top}K_{L}^{+}Y$ and $\mathcal{R}_{0}=\frac{1}{P}Y^{\top}K_{0}^{-1}Y$
. The predictor statistics:
\begin{multline}
\langle f(x)\rangle=k_{0}^{\top}(x)K_{0}^{-1}Y\\
\langle\delta f(x)\delta f(x)^{\top}\rangle=\mathcal{U}_{0}^{L}(K_{0}(x,x)-k_{0}^{\top}(x)K_{0}^{-1}k_{0}(x))\label{eq:var-sum-mul}
\end{multline}

Using the diagonal form of $\mathcal{R}_{0}$ , $\mathcal{R}_{0}=V_{0}\mathrm{diag}(r_{10},\cdots,r_{k0},\cdots,r_{m0})V_{0}^{\top},$the
self consistent equation of $\mathcal{U}_{0}$ is: 

\begin{align}
\mathcal{U}_{0} & =V_{0}\mathrm{diag}(u_{10},\cdots,u_{k0},\cdots,u_{m0})V_{0}^{\top}\nonumber \\
1-\alpha & =\sigma^{-2}u_{k0}-\alpha u_{k0}^{-L}r_{k0}\label{eq:u0sum-mul}
\end{align}
\end{minipage}}

\subsection{\label{subsec:Finite-temperature}Finite temperature}

Until now we focused on the limit of zero-temperature. We now consider
briefly the effect of finite temperature, i.e., when the training
error is not strictly minimized (see details in Appendix $\text{\ref{subsec:appendixa}}$).
Our BPKR framework holds for general temperature as well, where the
sole effect of temperature is to add to the renormalized $W$-dependent
kernel matrix, a regularizing diagonal term, $TI$, see Eq.\ref{eq:H_L0T}.
In particular, after $l$ successive integration of layer weights,
the effective Hamiltonian becomes, 
\begin{multline}
H_{L-l}(W')\\
=\frac{1}{2\sigma^{2}}\text{Tr}\ensuremath{W'^{\top}W'}+\frac{1}{2}Y^{\top}[u_{L-l}^{l}K_{L-l}+TI]^{-1}Y\\
+\frac{1}{2}\log\det(u_{L-l}^{l}K_{L-l}+TI)-\frac{lN}{2}\log u_{L-l}+\frac{lN}{2\sigma^{2}}u_{L-l}\label{eq:H_L-l-1}
\end{multline}
where the the kernel still undergoes kernel renormalization with the
scalar renormalization factor $u_{L-l}^{l}$. 

After integration of all the weights, the equations for the kernel
renormalization factor, $u_{0}$ become

\begin{multline}
1-\sigma^{-2}u_{0}=-\frac{1}{N}Y^{\top}(u_{0}^{L}K_{0}+TI)^{-2}u_{0}^{L}K_{0}Y\\
+\frac{1}{N}\mathrm{Tr}((u_{0}^{L}K_{0}+TI)^{-1}u_{0}^{L}K_{0})\label{eq:u_L-l-1}
\end{multline}

Furthermore, at finite $T$, the predictor mean and variances are
given as 
\begin{equation}
\langle f(x)\rangle=u_{0}^{L}k_{0}^{\top}(x)(u_{0}^{L}K_{0}+TI)^{-1}Y\label{eq:meanfiteration-1}
\end{equation}

\begin{multline}
\langle\left(\delta f(x)\right)^{2}\rangle\\
=u_{0}^{L}\left(K_{0}(x,x)-u_{0}^{L}k_{0}^{\top}(x)(u_{0}^{L}K_{0}+TI)^{-1}k_{0}(x)\right)\label{eq:varfiteration-1}
\end{multline}
Thus, at finite temperature both the mean and the variance of the
predictor differ from their GP counterparts, by renormalization of
all kernels by the factor $u_{0}^{L}$. 

\begin{figure*}
\includegraphics[width=0.8\textwidth]{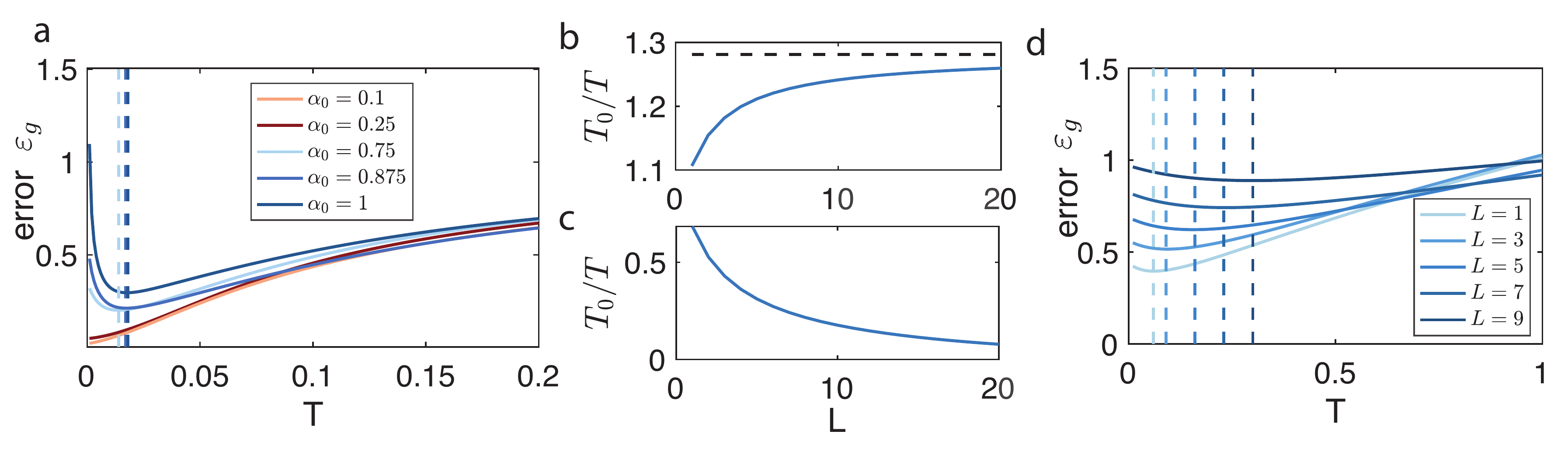}

\caption{\label{fig:egvsT}Finite temperature theory according to Eqs.$\text{\ref{eq:meanfiteration-1}, \ref{eq:varfiteration-1}}$
for a network trained on the ‘template’ model with random labels (see
parameters in Appendix $\text{\ref{a)-Template-model}}$) with finite
temperature. (a) The generalization error against the temperature
for three different $\alpha_{0}$'s. Dashed lines: the optimal temperature
for each $\alpha_{0}$. For this specific type of input data, for
small $\alpha_{0}$ ($\alpha_{0}=0.1,0.25$) the optimal temperature
is at $T=0$, for $\alpha_{0}$ close to 1 ($\alpha_{0}=0.75,0.875,1$)
the minimum generalization error is achieved at $T>0$. (b) For $\sigma^{2}(1-\alpha)<1$,
$u_{0}$ approaches $1$ and $\lambda=u_{0}^{-L}$ approaches the
finite value $\exp v_{0}$(black dashed line). (c) In another regime
where $\sigma^{2}(1-\alpha)>1$, $u_{0}$ approaches a limit larger
than 1 and $\lambda\rightarrow0$ as $L\rightarrow\infty$ . (d) The
generalization error vs. $T$ for five different $L$'s. Dashed lines:
the optimal temperature for each $L$. In the regime $\sigma^{2}(1-\alpha)>1$,
since $\lambda$ goes to $0$ and the temperature term $TI$ becomes
more neglegible compared to the renormalized kernel $u_{0}^{L}K_{0}$
as $L$ increases, larger $T$ is required to compensate for the small
$\lambda$ to obtain optimal generalization error.}
\end{figure*}

Although the predictor value for $x^{\mu}$ in the training set is
not $y^{\mu}$ (as is evident from Eq.\ref{eq:meanfiteration-1}),
the regularization provided by finite temperature may improve the
generalization error, in particular in the neighborhood of $\alpha_{0}=1$
where otherwise it would diverge. Indeed, for any specific task there
is an optimal temperature that minimizes the generalization, as shown
in the example of Fig.$\text{\ref{fig:egvsT}}$(a). Depending on the
specific training task and parameters, there exists cases where the
optimal temperature is 0, and also where the optimal temperature is
above 0, in which case the training error for optimal generalization
is nonzero. The existence of an optimal $T$ with minimum generalization
error may provide guidance for choosing an appropriate regularization
strength during training.

The effect of temperature is analogous to the effect of \emph{early
stopping} in the gradient descent dynamics (see \citep{advani2017high}
and SM VIII for more details). Finally, we discuss how the effect
of temperature changes with the depth of the network. From Eq.$\text{\ref{eq:varfiteration}}$
we observe that the effect of temperature on generalization performance
is controlled by the relative strength between the temperature term
$TI$ and the renormalized kernel $u_{0}^{L}K_{0}$. For finite temperature,
both $TI$ and $K_{0}$ are of $\mathcal{O}(1)$, and the relative
strength between the temperature term and the renormalized kernel
is thus controlled by $\lambda\equiv u_{0}^{-L}$, if $\lambda$ is
small, the effect of temperature is also small. The finite temperature
order parameter $u_{0}$ behaves in the limit of large $L$ similarly
to that at zero temperature. In the low-noise regime $\sigma^{2}(1-\alpha)<1$,
$u_{0}$ approaches unity for large $L$ as $u_{0}\approx1-\frac{v_{0}}{L}$,
hence $\lambda$ approaches the finite value $T\exp v_{0}$, as shown
in Fig.$\text{\ref{fig:egvsT}}$(b) (see Eq.$\text{\ref{eq:v0finiteT}}$
in Appendix $\text{\ref{subsec:appendixa}}$). On the other hand,
in the high-noise regime, $\sigma^{2}(1-\alpha)>1$, $u_{0}$ approaches
a limit larger than $1$ hence $\lambda$ goes to zero for deep networks
for all finite $T$ (Fig.$\text{\ref{fig:egvsT}}$(c)), as $L\rightarrow\infty$,
implying that the temperature term given by $TI$ can be neglected
compared to the renormalized kernel, and the behavior of the network
becomes similar to the zero-temperature behavior. In Fig.$\text{\ref{fig:egvsT}}$(d),
we look at the effect of temperature on the generalization performance
for networks with different depth $L$ in the large noise regime.
Since the temperature term $TI$ becomes more neglegible compared
to the renormalized kernel as $L$ increases, we see that the curve
becomes shallower as $L$ increases, suggesting the behavior at finite
$T$ becomes more similar to $T=0$; also for larger $L$, larger
$T$ is required to compensate for the small $\lambda$ to achieve
optimal performance.

\section{\label{sec:Changes-in-Representations}Changes in Representations
Across Layers}

Until now we discussed the behavior of the network output for different
parameter regimes. We now turn to ask how the representation of the
data changes across the different layers.

\subsection{\label{subsec:Mean-Layer-Kernels}Layerwise mean kernels }

The kernel matrices of the hidden layers are an important indicator
of the stimulus features represented by these layers, similar to the
role of similarity matrices \citep{shawe2004kernel,hofmann2008kernel}.
Hence, it is interesting to consider the statistics of the layerwise
kernels in our system. We find that the \emph{weight averaged} kernel
matrices are to leading order in $N$ identical to those resulting
from Gaussian weights (i.e., as in the GP limit). The non-Gaussianity
appears in mean kernels only in the $\mathcal{O}(1/N)$ corrections.
Specifically, for the single-output case

\begin{equation}
\langle K_{l}\rangle_{l}=\sigma^{2}(1-\frac{1}{N})K_{l-1}+\frac{\sigma^{2}}{Nu_{l-1}^{L-l+1}}YY^{\top}\label{eq:avekernel}
\end{equation}
The subscript $l$ emphasizes that the average is over $W_{l}$ and
the upstream weights, i.e., $\{W_{k}\}_{k>l-1}$. Proceeding to successively
integrate all upstream weights, the fully averaged kernel of the $l$-th
layer ($1\leq l\leq L$) is given by

\begin{equation}
\langle K_{l}\rangle=\sigma^{2l}(1-\frac{1}{N})^{l}K_{0}+\frac{m_{l}}{N}YY^{\top}\label{eq:Kl-1 averaged}
\end{equation}
where the amplitudes $m_{l}$ consist of the sum of the geometric
series with terms such as in Eq.\ref{eq:avekernel}, yielding $m_{l}=\sigma^{2l}u_{0}^{-L}\left(\frac{u_{0}^{l}\sigma^{-2l}-1}{u_{0}\sigma^{-2}-1}\right)$(see
Appendix $\text{\ref{subsec:Mean-layer-kernels}}$ and SM VA). Thus,
the correction term in Eq.\ref{eq:Kl-1 averaged} encodes the output
task via the output similarity matrix $YY^{\top}$(and $u_{0}$).
In the following analysis and examples, we consider the regime $\sigma\ll1$,
where the second term in Eq.$\text{\ref{eq:Kl-1 averaged}}$ becomes
evident compared to the first term. The shape of this correction is
the same in all layers; however, its relative strength compared to
the GP term increases with the depth of the layer (see Appendix $\text{\ref{subsec:Mean-layer-kernels}}$
and SM VB).

The situation is richer in the case of multiple outputs. Here the
layerwise mean kernels are ($1\leq l\leq L$) 
\begin{equation}
\langle K_{l}\rangle=\sigma^{2l}(1-\frac{m}{N})^{l}K_{0}+\frac{1}{N}YV_{0}M_{l}V_{0}^{\top}Y^{\top}\label{eq:Kl-1 averagedm}
\end{equation}
where $M_{l}$ is the diagonal matrix $M_{l}$ whose $k$-th eigenvalue
is $m_{kl}=\sigma^{2l}u_{k0}^{-L}\left(\frac{\sigma^{-2l}u_{k0}^{l}-1}{\sigma^{-2}u_{k0}-1}\right)$.
For all $l$, the maximal eigenvalue of $M_{l}$ corresponds to the
mode with the smallest eigenvalue of $R_{0}$ and this mode may dominate
the correction to the mean kernel matrix (Appendix $\text{\ref{subsec:Mean-layer-kernels}}$
and SM VB). Note that with multiple outputs the corrections represented
by the last term in Eq.\ref{eq:Kl-1 averagedm} are not simply proportional
to the output similarity matrix as in the single-output case. The
difference is pronounced if the spectrum of $U_{0}$ (or equivalently
that of $R_{0}$) departs substantially from uniformity.

\begin{figure}
\includegraphics[width=1\columnwidth]{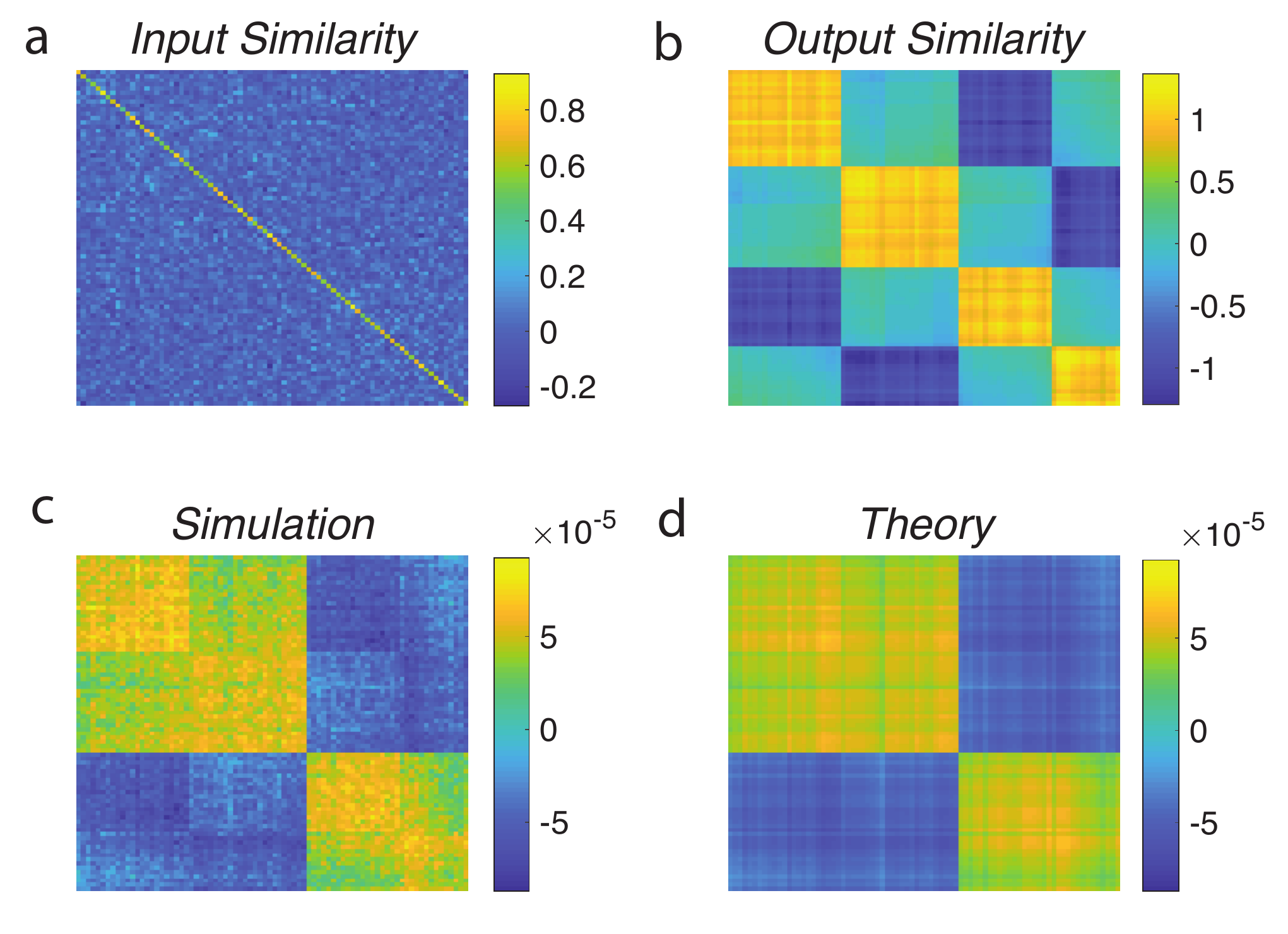}

\caption{\label{fig:examplesynthetic} Hidden representations with multiple
outputs. Results for a single hidden-layer network ($L=1$) trained
on the synthetic example described by Eq.$\text{\ref{eq:synthetic}}$.
In this figure and the next Fig.$\text{\ref{fig:deepkernels}}$, to
enhance the relative strength of the task-relevant structure, we choose
small $\sigma$, corresponding to strong regularization. (a) The input
similarity matrix. (b) The output similarity matrix, showing 4 diagonal
blocks corresponding to the 4 categories of the 2 labels (positive/negative
projection onto $P-1$-th/$P$-th dimension). (c) Simulation result
corresponding to the non-GP correction term in Eq.$\text{\ref{eq:Kl-1 averagedm}}$
shows only 2 blocks corresponding to the classification into 2 categories
along the $P-1$-th direction, the structure of blocks corresponding
to the $P$-th classification direction as shown in (b) is now suppressed.
(d) Theory for the non-GP correction term, Eq.$\text{\ref{eq:Kl-1 averagedm}}$.
(See Appendix $\text{\ref{b)-Synthetic-example}}$ for details.)}
\end{figure}

A synthetic example is shown in Fig.$\text{\ref{fig:examplesynthetic}}$,
in which the $P$ input vectors $x_{\mu}$ are linear combinations
of $P$ orthogonal vectors $z_{i}$ with $z_{i}^{T}z_{j}=\delta_{ij}$.
\begin{equation}
x_{\mu}=\sum_{i=1}^{P-1}w_{i}^{\mu\top}z_{i}+w_{P}^{\mu\top}z_{P}\label{eq:synthetic}
\end{equation}
The linear coefficients $w_{i}^{\mu}$ are sampled i.i.d from $\mathcal{N}(0,I)$
for $i\leq P-1$ but from $\mathcal{N}(0,\frac{1}{10}I)$ for $i=P$.
The output is two-dimensional, classifying the inputs according to
the sign of their projections on the $P-1$-th and the $P-$th basis
vectors, respectively (i.e., $Y=\mathrm{sgn}([z_{P-1},z_{P}]^{\top}X)+\sigma_{0}\eta$).
As a result, the output similarity matrix $\frac{1}{m}YY^{\top}$shows
4 blocks corresponding to the 4 categories (Fig.$\text{\ref{fig:examplesynthetic}}$(b)).
However, because the input is not fully aligned with the output, and
the output direction corresponding to $z_{P}$ has smaller variance
than that corresponding to $z_{P-1}$, the block corresponding to
the $P$-th classification direction is suppressed in the non-GP correction
to the kernel of the hidden layer, Eq.\ref{eq:Kl-1 averagedm} (Fig.$\text{\ref{fig:examplesynthetic}}$(c,d)).
Note that the observed similarity pattern is not a linear combination
of the input similarity $\frac{1}{N}X^{\top}X$ matrix (which is almost
structureless, Fig.$\text{\ref{fig:examplesynthetic}}$(a)) and the
output similarity matrix (Fig.$\text{\ref{fig:examplesynthetic}}$(b)).

\begin{figure}
\includegraphics[width=1\columnwidth]{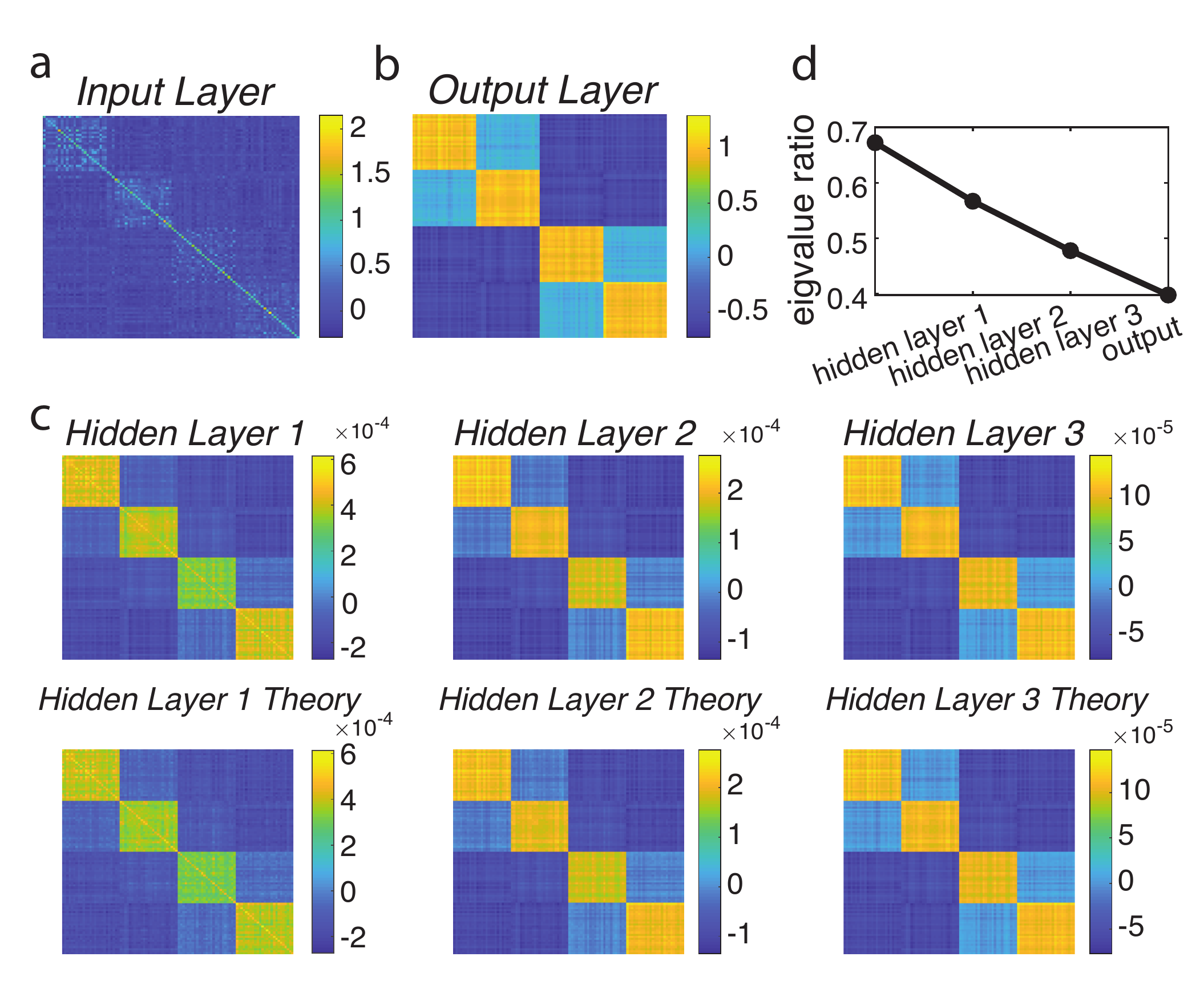}

\caption{\label{fig:deepkernels}Simulation and theory for the mean kernel
for binary classification task on MNIST. The network is trained on
4 different MNIST digits which are grouped into two higher order categories
(see Appendix $\text{\ref{c)-Binary-classification}}$). The output
of the network is 6-dimensional: 4 of the output units are ‘one-hot’
representations of the 4 digits, the other two outputs label the inputs
according to their high order category. (a) The input similarity matrix.
(b) The output similarity matrix. (c) Average kernel of the hidden
layer for $l=1,2,3$. Top: simulation. Bottom: theory. (d) The ratio
between the mean of the second and third largest eigenvalues (corresponding
to the magnitude of the 4 smaller blocks) and the largest eigenvalue
(corresponding to the magnitude of the two larger blocks) of the non-GP
correction terms in the layerwise mean kernel is monotonically decreasing
with $l$.}
\end{figure}

Another example shown in Fig.$\text{\ref{fig:deepkernels}}$ considers
a linear network with 3 hidden layers and 6 output units each performing
a binary classification task on MNIST input images of $4$ digits
(see details of this task in Appendix $\text{\ref{c)-Binary-classification}}$).
Here the input similarity matrix $\frac{1}{N}X^{\top}X$ shows a weak
but noticeable 4-block structure (Fig.$\text{\ref{fig:deepkernels}}$(a))
corresponding to 4 different digits. The output similarity matrix
$\frac{1}{m}YY^{\top}$ exhibits a pronounced hierarchical block structure
(Fig.$\text{\ref{fig:deepkernels}}$(b)). We ask how the block structure
is modified in the different hidden-layer kernels (Fig.$\text{\ref{fig:deepkernels}}$).

We observe three major effects of the changes in the average kernels
across layers. First, the magnitude of the task related contribution
to the mean kernel increases as $l$ increases as expected from the
theory (Appendix $\text{\ref{subsec:Mean-layer-kernels}}$ and SM
VB). Second, in this example we find that the finer-scale structure
becomes more pronounced in the mean layer kernel than in the output
similarity matrix, as can be seen from comparing Fig.$\text{\ref{fig:deepkernels}(b)}$
and Fig.$\text{\ref{fig:deepkernels}}$(c). Third, the contributions
from finer-scale structure becomes less pronounced for deeper layers,
as seen in Fig.$\text{\ref{fig:deepkernels}}$(c). The second and
third point can also be observed more straight-forwardly in Fig.$\text{\ref{fig:deepkernels}}$(d),
where we show the ratio between the mean of the second and third largest
eigenvalues (corresponding to the 4 smaller blocks) and the largest
eigenvalue (corresponding to the 2 larger blocks) of the non-GP correction
terms in the layerwise mean kernels. This ratio decreases with $l$,
suggesting that the finer structure becomes less pronounced for large
$l$, and this ratio for all hidden layers is larger than that for
the output layer
The origin for agreement of the mean kernels with their GP limit to
leading order in $N$, is that the second-order statistics of hidden-layer
weights are just their GP values to leading order in \emph{$N$}.
Their renormalization appears only in the $\mathcal{O}(1/N)$ corrections
to their covariance matrix (Appendix $\text{\ref{subsec:Mean-layer-kernels}}$).
This is because the learning-induced terms in the effective Hamiltonians,
such as Eq.\ref{eq:HL-l}, are of order $P$ (as there are $P$ training
constraints) which is of the order of $N$, while the $L_{2}$ Gaussian
term is of the order of the number of weights in each layer, which
is $N^{2}$. On the other hand, the leading term and the correction
terms scale differently with $\sigma$, such that in the low-noise
limit the strength of the correction relative to the GP term grows
as $\sigma^{-2l}/N$ (SM ${\rm VB}$).

\subsection{\label{subsec:Mean-Inverse-Kernels}Mean inverse kernels}

\begin{figure}
\includegraphics[width=1\columnwidth]{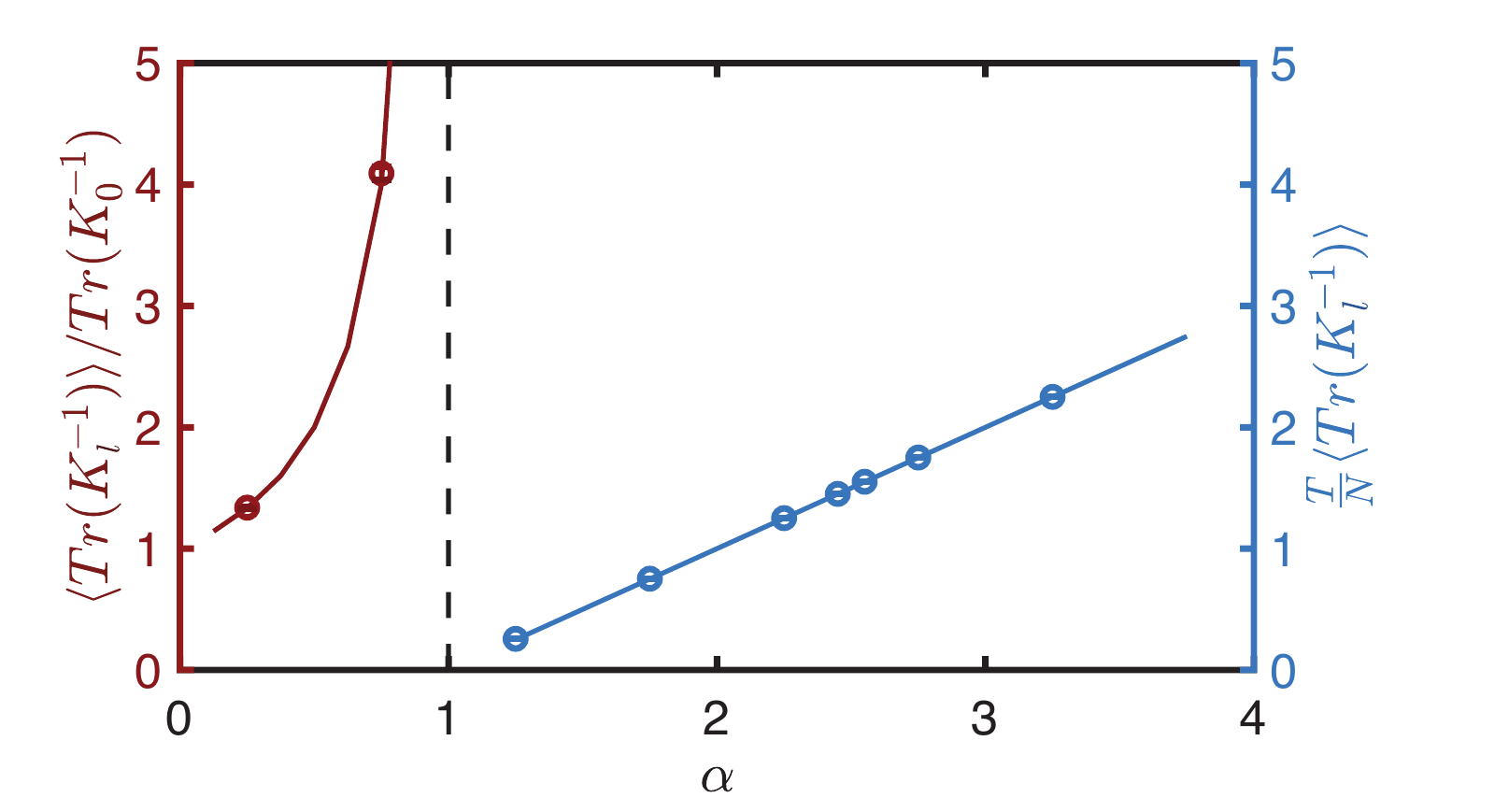}

\caption{\label{fig:invkernel}The trace of the average inverse kernel of a
single hidden-layer network. Lines: theory. Circles: simulations.
Simulations are done with the same model and parameters as Fig.$\text{\ref{fig:egvsP}},$
described in Appendix $\text{\ref{a)-Template-model}}$. Left: The
mean inverse kernel diverges as $\alpha\rightarrow1$ at zero temperature.
Right: The trace of the average inverse kernel for the single hidden-layer
network multiplied by $T$, showing the divergenc of the trace for
all $\alpha>1$ as $\frac{1}{T}$ as $T\rightarrow0$.}
\end{figure}

While the average kernel retains, to leading order, its GP value,
the \emph{average inverse} kernel does not. In fact, to leading order
in $N$, we obtain

\begin{equation}
\langle K_{l}^{-1}\rangle=\frac{1}{\sigma^{2l}(1-\alpha)^{l}}K_{0}^{-1}\label{eq:invKl}
\end{equation}
However, similar to the mean, the average inverse kernels encode the
target outputs only in the correction terms. Eq.\ref{eq:invKl} implies
that the mean inverse kernel matrix diverges as $\alpha\rightarrow1$
at zero temperature (Fig.$\text{\ref{fig:invkernel}}$) . In fact,
its trace for all $\alpha>1$ is proportional to $1/T$ for small
$T$ (see Fig.$\text{\ref{fig:invkernel}}$ and SM VI).
This divergence of the zero-temperature mean inverse kernels in narrow
networks is expected as discussed in Section $\text{\ref{sec:The-Back-propagating-KR}}$.
Note that due to averaging over weights, the mean kernel, Eq.\ref{eq:Kl-1 averaged},
has a rank of $P$, even when $N<P$. Furthermore, the divergence
of the mean inverse kernels when $\alpha\geq1$, does not lead to
divergence of the mean squared readout parameters $r_{l}$ and the
renormalization scalars $u_{l}$, as observed above (and explained
in Appendix $\text{\ref{subsec:appendixa}}$ and SM IA, IB). 

Concluding this section, we note that even though the second-order
weight statistics and the related mean kernel are to leading order
equal to their GP limit, other statistics of the weights and kernels,
and in particular the predictor statistics and the generalization
error deviate from the GP limit already to leading order, for all
$\alpha=\mathcal{O}(1)$, as shown here and in Section $\text{\ref{sec:Generalization}}$.

\section{\label{sec:Deep-ReLu-Networks}BPKR in Deep ReLU Networks}

\subsection{Approximate BPKR for ReLU networks}

Our theory applies to deep networks with linear units, which are limited
in their expressive power. To enhance the system's expressivity, one
might adopt an architecture comprising a fixed (non-learned) nonlinear
mapping of the input to a shallow layer that then projects to the
deep linear networks with learned synapses, as has been studied extensively
in recent years (e.g., inputs projecting to a nonlinear kernel representation
or to a layer of nonlinear neurons via random weights \citep{belkin2019reconciling,cutajar2017random,rahimi2008random}).
Since our theory does not rely on specific assumptions about the input
statistics, our BPKR applies readily to this architecture, with the
input vectors and the associated input kernel defined by the nonlinear
representation of the shallow layer.

Our theory is not expected to hold for architectures where \emph{the
learned weights} project to nonlinear units, as is the case in most
applications of DNNs. In such cases integration of even one layer
of synapses is hard. Here we ask to what extent our theory can be
adapted to nonlinear networks to yield a reasonable approximation
in some parameter regimes. For simplicity we assume a single linear
output unit.

We recall that in the GP limit, the properties of DNNs are accounted
for by the GP kernels appropriate for the chosen nonlinearity \citep{cho2009kernel}.
For example, infinitely wide deep networks with ReLU nonlinearity,
which will be studied here, yield GP kernels for the $l$-th layer
of the form

\begin{align}
\langle K_{l}^{GP}(x,y)\rangle & =\frac{\sigma^{2}}{2\pi}\sqrt{\langle K_{l-1}^{GP}(x,x)\rangle\langle K_{l-1}^{GP}(y,y)\rangle}J(\theta_{l-1})\label{eq:iterationGP}\\
J(\theta_{l-1})= & \sin(\theta_{l-1})+(\pi-\theta_{l-1})\cos(\theta_{l-1})\nonumber 
\end{align}

\begin{equation}
\theta_{l-1}=\arccos\left(\frac{\langle K_{l-1}^{GP}(x,y)\rangle}{\sqrt{\langle K_{l-1}^{GP}(x,x)\rangle\langle K_{l-1}^{GP}(y,y)\rangle}}\right)
\end{equation}
where $\theta_{l-1}$ represents the angle between the $l-1$ representations
of $x$ and $y$ , and the superscript $GP$ in $K_{l}^{GP}$ specifies
the \emph{GP kernel} of the $l$-th layer, differentiating from the
$K_{l}$ we previously defined in Eq.$\text{\ref{eq:layerkernels}}$,
which represents the dot product of activations of the network. These
equations can be solved by iteration from the initial condition $K_{0}^{GP}(x,y)=K_{0}(x,y)=\frac{\sigma^{2}}{N_{0}}x^{\top}y$,
for a pair of input vectors $x$ and $y$.

The average symbol is the result of (self-)averaging w.r.t. Gaussian
weights. In the GP limit, the predictor statistics of a network with
$L$ layers are given in terms of these kernels as 
\begin{equation}
\langle f(x)\rangle=\langle k_{L}^{GP\top}(x)\rangle\langle K_{L}^{GP}\rangle^{-1}Y\label{eq:meanfiteration-relu}
\end{equation}

\begin{multline}
\langle\left(\delta f(x)\right)^{2}\rangle\\
=\langle K_{L}^{GP}(x,x)\rangle-\langle k_{L}^{GP\top}(x)\rangle\langle K_{L}^{GP}\rangle^{-1}\langle k_{L}^{GP}(x)\rangle\label{eq:varfiteration-relu}
\end{multline}
where $\langle k_{L}^{GP\mu}(x)\rangle=\langle K_{L}^{GP}(x,x^{\mu})\rangle$.

To extend the BPKR to ReLU networks with finite $\alpha$, we make
the ansatz that the weight statistics are modified relative to their
GP value by a scalar kernel renormalization, $u_{0}$. Because in
the ReLU nonlinearity $K_{l}$ is a linear function of the amplitude
of $K_{l-1}$ we reason that the iterative equation has a similar
structure to the linear network case, culminating in 
\begin{equation}
1-\sigma^{-2}u_{0}=\alpha(1-u_{0}^{-L}r_{0})\label{eq:u0-1}
\end{equation}

\begin{equation}
r_{0}=\frac{\sigma^{2L}}{P}Y^{\top}\langle K_{L}^{GP}\rangle^{-1}Y\label{eq:b0-1}
\end{equation}
and consequently, the mean predictor is unchanged from Eq.\ref{eq:meanfiteration-relu},
while the variance is given by

\begin{multline}
\langle\left(\delta f(x)\right)^{2}\rangle\\
=u_{0}^{L}\sigma^{-2L}(\langle K_{L}^{GP}(x,x)\rangle-\langle k_{L}^{GP\top}(x)\rangle\langle K_{L}^{GP}\rangle^{-1}\langle k_{L}^{GP}(x)\rangle)
\end{multline}
Note that in the linear case, the Gaussian averaged $\langle K_{L}^{GP}\rangle=\sigma^{2L}K_{0}$,
which reduces the above equations to the exact BPKR equations (see
Eqs.$\ref{eq:meanfiteration}$,-$\text{\ref{eq:varfiteration}}$).
Also, for $\alpha=0$ Eq.\ref{eq:u0-1} yields $u_{0}=\sigma^{2}$
and the theory reduces to the GP limit for ReLU networks.

\subsection{Generalization in ReLU networks}

\begin{figure*}
\includegraphics[width=0.9\textwidth]{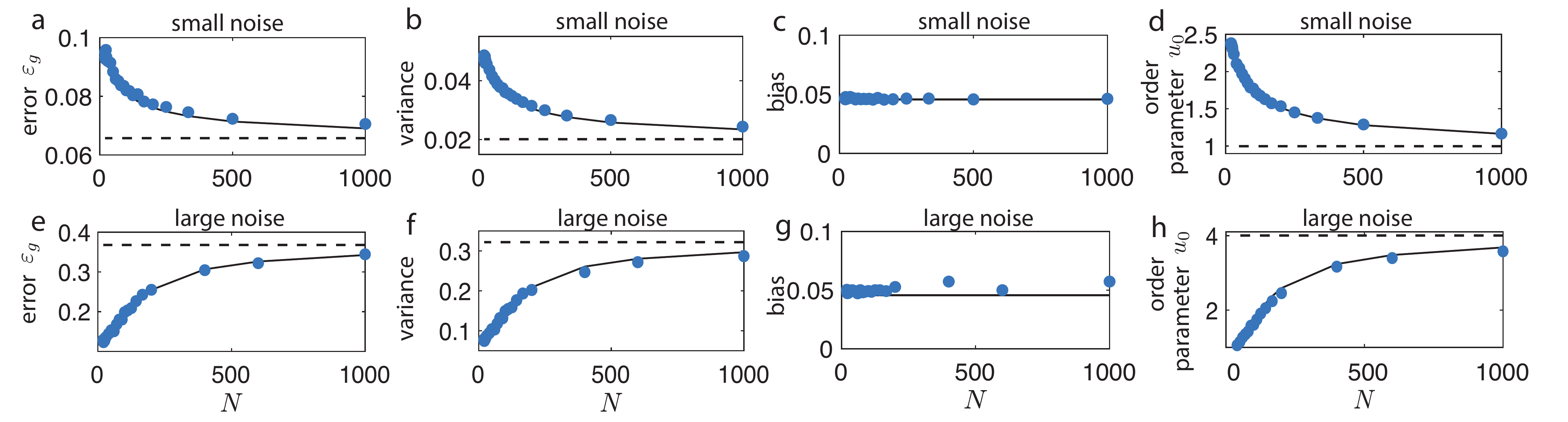}

\caption{\label{fig:relu_linearteacher}A single hidden-layer ($L=1$) ReLU
network trained on the ‘template’ model with labels generated by a
noisy linear teacher with details of parameters in Appendix $\text{\ref{a)-Template-model}}$.
Generalization error (a,e), variance (b,f) ; bias (c,g) of the predictor;
and the order parameter $u_{0}$ (d,h) as a function of $N$. Black
lines: theory. Blue dots: simulation. Black dashed lines: GP limit
($N=\infty)$. (a-d) Results in the small noise regime where the generalization
error decreases with $N$. (e-h) Results in the large noise regime
where the generalization error increases with $N$.}
\end{figure*}

\begin{figure*}
\includegraphics[width=0.9\textwidth]{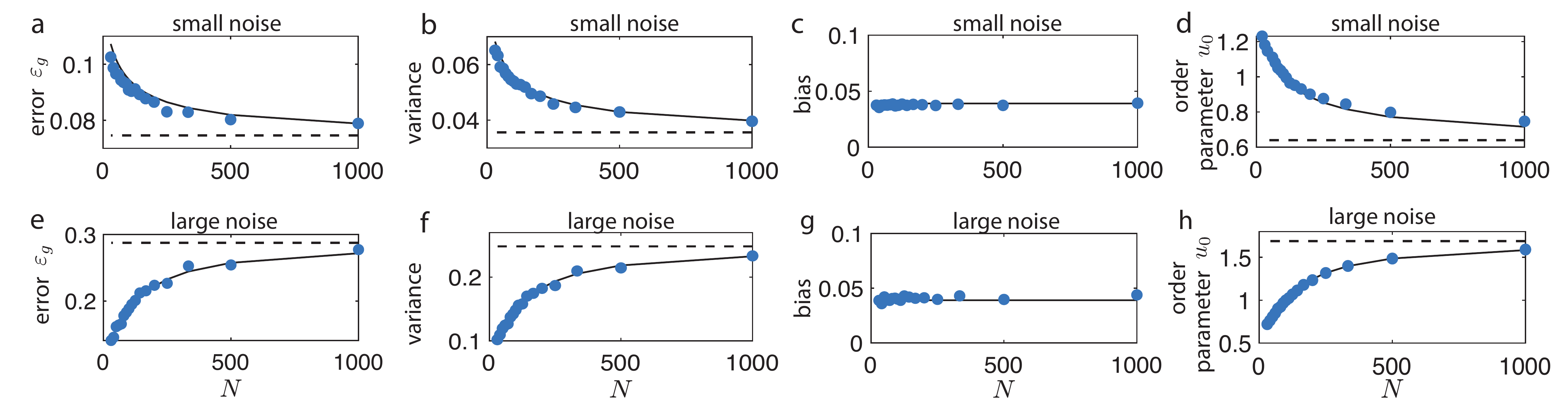}

\caption{\label{fig:relu_mnist}A single hidden-layer ($L=1$) ReLU network
trained on MNIST binary classification of 2 digits (0 and 1) with
details of parameters in Appendix $\text{\ref{d)-Binary-classification}}$.
Generalization error (a,e), variance (b,f) and squared bias (c,g)
of the predictor, and order parameter $u_{0}$ (d,h) as a function
of $N$. Black lines: theory. Blue dots: simulation. Black dashed
lines: GP limit ($N\rightarrow\infty$). (a)-(d) Results in the small-noise
regime where the generalization error decreases with $N$. (e-h) Results
in the large-noise regime where the generalization error increases
with $N$.}
\end{figure*}

Our approximate BPKR predicts that the generalization error increases
with $\alpha$ for low $\sigma$ and decreases for high $\sigma$.
We have checked these predictions for a ReLU network of a single hidden
layer network trained for the noisy linear teacher task described
in Appendix $\text{\ref{a)-Template-model}}$. Results are shown in
Fig.$\text{\ref{fig:relu_linearteacher}}$.
We have also checked these predictions for a ReLU network of a single
hidden layer trained for MNIST binary classification task (see details
in Appendix $\text{\ref{d)-Binary-classification}}$) as shown in
Fig.$\text{\ref{fig:relu_mnist}}$.

The simulation behaves qualitatively the same as predicted by the
theory, $\varepsilon_{g}$ in the ReLU network increases with $\alpha$
for small noise and decreases at high noise. Furthermore, surprisingly,
there is also a good quantitative agreement between the simulations
and the approximate BPKR for ReLU networks, even for small $N$ (i.e.,
$\alpha\sim10)$. The mean predictor contributing to the bias component
of the generalization error is constant and its value fits the prediction
given by Eq.\ref{eq:meanfiteration-relu}. The predictor variance
as well as $\varepsilon_{g}$ vary with $N$ in close agreement with
the approximate BPKR prediction. Furthermore, the order parameter
$u_{0}$ defined by $u_{0}^{L}=\frac{\sigma^{2L}Y^{\top}\langle K_{L}^{GP}\rangle^{-1}Y}{Y^{\top}\left\langle K_{L}^{-1}\right\rangle Y}$
varies with $N$ in close agreement with Eq.\ref{eq:u0-1}.

\begin{figure*}
\includegraphics[width=0.9\textwidth]{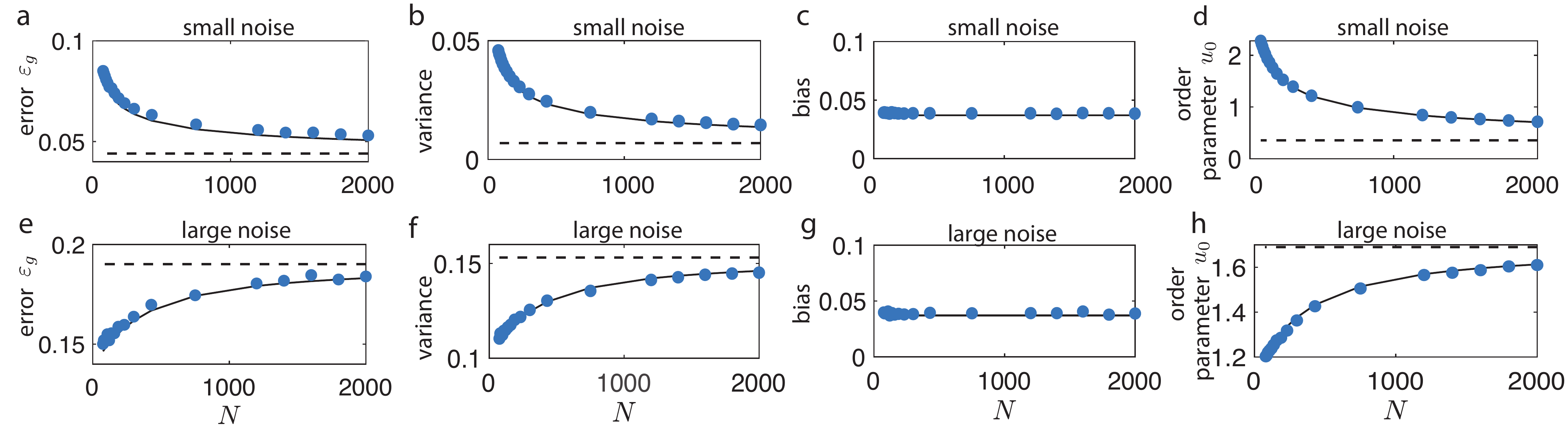}

\caption{\label{fig:alpha0>1}A single hidden-layer ($L=1$) ReLU network trained
on the ‘template’ model linear teacher example with details of parameters
in Appendix $\text{\ref{a)-Template-model}}$, in the $\alpha_{0}>1$
regime. Generalization error (a,e), variance (b,f) and squared bias
(c,g) of the predictor, and the order parameter $u_{0}$ (d,h) as
a function of $N$. Black lines: theory. Blue dots: simulation. Black
dashed lines: GP limit ($N=\infty)$. (a-d) Results in the small-noise
regime where the generalization error decreases with $N$. (e-h) Results
in the large-noise regime where the generalization error increases
with $N$.}
\end{figure*}

In the examples above, $\alpha_{0}<1$. In the linear network we found
a divergence of the bias and the generalization error at $\alpha_{0}=1$
and vanishing of the predictive variance for $\alpha_{0}>1$. These
features are not expected to hold for the nonlinear network due to
the stronger expressivity contributed by the trained nonlinear hidden
layer. We asked whether our ansatz serves as a good approximation
also in the regime of $\alpha_{0}>1$ where the nonlinearity plays
a crucial role in allowing for zero training error. The results are
shown in Fig.$\text{\ref{fig:alpha0>1}}$.
Surprisingly, even here, results for both the predictor statistics
and the order parameter are in good agreement with the theory.

\begin{figure*}
\includegraphics[width=1\textwidth]{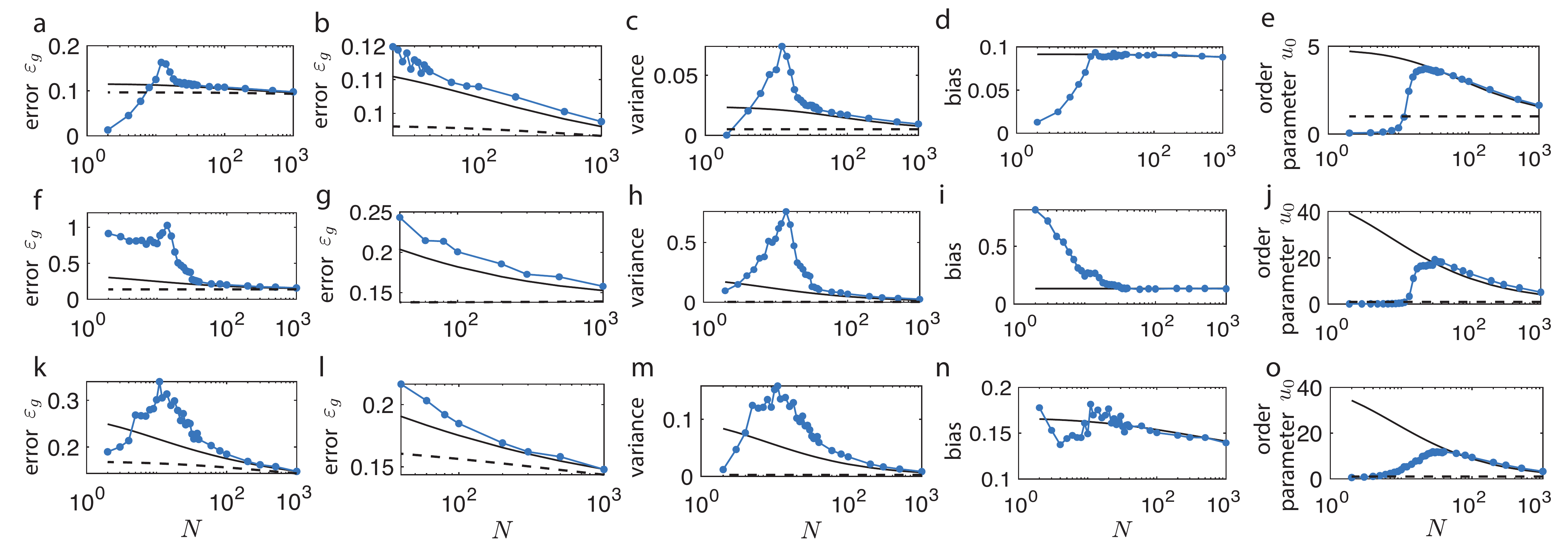}

\caption{\label{fig:doubledescent}A single hidden-layer ReLU network with
smaller $N_{0}$ to push the network closer to its capacity. Black
lines: theory. Blue lines: simulation. Black dashed lines: GP limit.
The generalization error (a,f,k), ( (b,g,l) zooming in on the large
$N$ part of (a,f,k) to show the convergence to the theory, and to
compare our ansatz with the GP approximation), variance (c,h,m), and
bias (d,i,n) of the predictor, and the order parameter $u_{0}$ (e,j,o).
(a-e) Trained on the linear teacher task with detailed parameters
in Appendix $\text{\ref{a)-Template-model}}$. The minimum generalization
error is observed at small $N$, on the LHS of the interpolation threshold
due to the linearity of the task. (f-j) Trained on the ‘template’
model with random labeling of each cluster (see detailed parameters
in Appendix $\text{\ref{a)-Template-model}}$). The task itself requires
nonlinearity, and the minimum generalization error is achieved in
the over-parameterized regime. (k-o) Trained on the randomly projected
MNIST data of two digits (see Appendix $\text{\ref{c)-Binary-classification}}$).
This task also requires nonlinearity, and the minimum generalization
error is achieved in the over-parameterized regime. (In this figure
we also use the finite T ansatz (SM VII) for the theory curves.) }
\end{figure*}

In all previous examples we did not observe double descent in $\varepsilon_{g}$
because we are in the regime where the network achieves zero training
error for all $N\geq$2 (because our $N_{0}$ is sufficiently large).
We therefore test our results with small $N_{0}$, pushing the network
closer to its interpolation threshold, which is roughly when $N\sim\alpha_{0}$,
i.e., $P\sim NN_{0}$ (i.e., the number of learned parameters equals
the number of training data, \citep{vershynin2020memory}). Indeed,
we see significant deviation from the approximate BPKR as $N$ decreases
and approaches $\alpha_{0}$, which suggests that the scalar renormalization
of the kernel becomes inadequate as the network approaches its expressivity
capacity. While the simulation shows a double descent behavior, our
theoretical ansatz does not (Fig.$\text{\ref{fig:doubledescent}}$).
The theoretical results agree with the simulations only on the RHS
of the interpolation threshold, i.e., larger $N,$ and they fit the
simulations significantly better than the GP approximation as shown
in Fig.$\text{\ref{fig:doubledescent}}$(b,g,l). Incidentally, it
is interesting to compare the generalization behavior in the three
tasks which differ in their complexity. In the linear teacher task
(Appendix $\text{\ref{a)-Template-model}}$), the \emph{minimum }generalization
error is on the LHS of the interpolation threshold. However, for the
random labeling task (Appendix $\text{\ref{a)-Template-model}}$)
and classification of MNIST data (Appendix $\text{\ref{d)-Binary-classification}}$),
due to the nonlinearity of the task, a large number of network parameters
are required in order to generalize well, and the minimum generalization
error is achieved on the RHS of the interpolation threshold, which
is similar to the linear network (Fig.$\text{\ref{fig:egvsP}}$).
Importantly, for $N$ below the interpolation threshold while the
training error is nonzero, the minimal training error solution is
not unique, and this degeneracy in the weights induces variability
in the input-output mapping of the network, as shown by the non-vanishing
of the predictor variance in the left side of the peak in Fig.$\text{\ref{fig:doubledescent}}$,
except at $N=1$. This is different from the linear case, where for
$\alpha_{0}>1$ the predictor variance vanishes (see Eq.$\text{\ref{eq:varfiteration}}$
and Fig.$\text{\ref{fig:egvsP}}$).

\begin{figure*}
\includegraphics[width=0.9\textwidth]{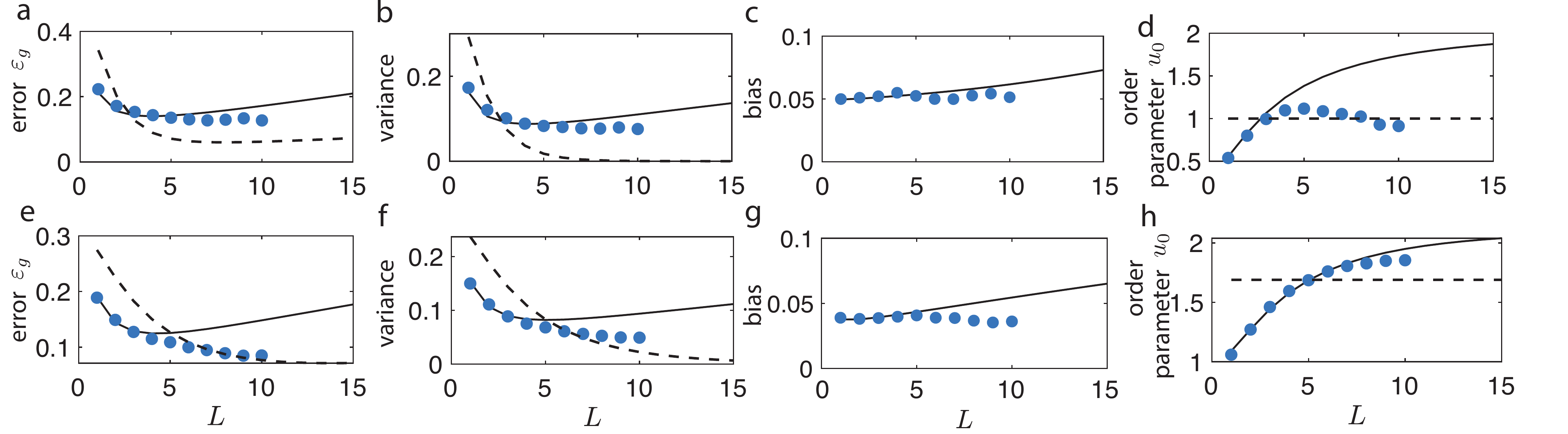}

\caption{\label{fig:deeprelu}The generalization error of deep ReLU networks
as a function of depth $L$. Blue dots: simulation. Black lines: theoretical
approximation. Black dashed lines: GP limit. Generalization error
(a,e), variance (b,f), and bias (c,g) of the predictor, and the order
parameter $u_{0}$ (d,h). (a-d) Results for the ‘template’ model with
noisy linear teacher labels, with parameters in Appendix $\text{\ref{a)-Template-model}}$.
(e-h) Results for a binary MNIST classification task with parameters
in Appendix $\text{\ref{d)-Binary-classification}}$. }
\end{figure*}
All the examples above were for a single hidden layer. We also test
our ansatz against the simulation results for the ReLU network with
multiple hidden layers. As we see in Fig.$\text{\ref{fig:deeprelu}}$,
our approximate BPKR agrees reasonably well with the simulation for
$L=1\sim5$, and is significantly better than the predictions of the
GP limit, but the agreement fails for large $L$. This suggests that
for finite $\alpha$, when $L$ becomes larger, renormalization of
the kernel just by a scalar becomes inadequate.

\section{Discussion}

\textbf{Summary: }Since the seminal work of Gardner \citep{gardner1988space,gardner1988optimal},
statistical mechanics has served as one of the major theoretical frameworks
for understanding the complexity of supervised learning. However,
so far it has focused mostly on shallow architectures and addressed
the classical bias-variance tradeoff where calculated learning curves
displayed improvement of generalization when the number of examples
was large compared to the system size \citep{domingos2000unified,geman1992neural}.
Statistical mechanics has also focused on phase transitions, local
minima, and spin-glass properties due to the underlying nonlinearity
of the learning cost function and the quenched randomness of the training
data \citep{seung1992statistical,ganguli2010statistical}. It is well-known
that Deep Learning challenges many of the above intuitions, calling
for a new theory of learning in deep architectures. In this work we
have developed a new statistical mechanics framework of learning in
a family of networks with deep architectures. To make the theory analytically
tractable, we have focused on networks with linear units. Despite
their limited expressive power, they do share important aspects of
nonlinear deep networks, as is highlighted in our work. Importantly,
unlike most previous statistical mechanical theories of learning,
which resorted to extremely simplifying assumptions about the input
statistics and the target labels, our theory is general – fully exposing
the relation between network properties and task details. 

DLNNs have been the focus of several studies. \citep{laurent2018deep,lu2017depth}
prove the absence of sub-optimal local minima under mild conditions,
a result which is consistent with our results. A very interesting
work \citep{saxe2019mathematical} studied the gradient descent dynamics
of learning in DLNNs, with results that depended critically on the
initial conditions (small random weights) and only became tractable
with simplifying assumptions about the data ($XX^{\top}=I$ and $P>N$).
Keeping $N$ and $P$ fixed for most of the simulation and analysis,
\citep{saxe2019mathematical} addressed the changes in representation
during training and across different layers. Under the restricted
assumptions, they found that (when there are multiple outputs) the
learning dynamics can be decomposed into multiple modes which evolve
independently – qualitatively similar to the multiple modes found
in our analysis (see further below). However, they did not address
the basic question of the system's performance such as the predictor
statistics and the generalization error and its critical dependence
on various network parameters. Here we study the nature of the Gibbs
distribution in the weight space induced by learning with the training
mean squared error as the Hamiltonian. We have focused mainly (but
not exclusively) on the properties of the feasible weight space consisting
of weight vectors that yield zero training error, which is the case
in many real-world applications of DNNs, with the well-known $L_{2}$
regularization (with an amplitude parameterized by inverse noise,
$\sigma^{-2}$). 

Due to the highly nonlinear nature of the training Hamiltonian, evaluating
the statistical mechanical properties of DLNNs seems intractable.
Here, we developed the BPKR method to integrate out the weight matrices
layer-by-layer, allowing us to derive equations for the system's properties
which are exact in the thermodynamic limit. Importantly, in contrast
to most kernel-based theories of deep networks, our thermodynamic
limit is defined by letting both the width, $N$, and training size
$P$ diverge while the load $\alpha=P/N$ remains of order $1$, extending
the well-known thermodynamic limit of statistical mechanics of learning
to deep architectures \citep{advani2013statistical,advani2017high}.
We have shown that the effect of the finite load is to change the
effective Hamiltonians through an $\alpha$-dependent kernel renormalization
at each successive step of weight integration. 

In addition to load, $\alpha$, depth $L$, and weight noise parameter
$\sigma^{2}$. Inputs and their labels in the training data affect
the properties of the system through the mean squared input-layer
readout parameter, $\sigma^{2}r_{0}$, Eq.\ref{eq:b0}. Our results
yield rich phase diagrams specifying the dependence of the generalization
error on the width and the depth of the network, Figs.$\ref{fig:egvsN}$,$\text{\ref{fig:egvsL},\ref{fig:summaryplot}}$.
Importantly, depending on $\sigma^{2}$ and $\sigma^{2}r_{0}$, the
generalization error may decrease upon increasing width (i.e., decreasing
$\alpha$) and increasing depth (i.e., increasing $L$). Since this
occurs within the over-parameterized regime where the training error
is zero throughout, our results prove that in an exactly solvable
deep network increasing network complexity may lead to a substantial
improvement in generalization. We were also able to identify the parameter
regimes where this improvement happens. 

Importantly, the BPKR also enables us to evaluate the posterior
properties of each layer's weights imposed by learning. We leverage
this to explore the effect of input and output data on the layerwise
similarity matrices induced through learning and show that due to
different renormalization strengths, amplification of modes in the
layer representations is not uniform, as demonstrated in the examples
in Figs.$\ref{fig:examplesynthetic},$$\text{\ref{fig:deepkernels}}$.
Recent studies have analyzed the similarity matrices of neuronal activities
for structured tasks and compared them with the representations at
the hidden layers of DNNs \citep{yamins2014performance,kriegeskorte2008representational,messinger2001neuronal}.
Therefore, our work may provide theoretical understanding of how neuronal
representations are constrained by the task structure. 

\textbf{BPKR and GD learning:} In the case of multiple outputs, we
show in Section $\text{\ref{subsec:Multiple-Outputs}}$ that the layerwise
renormalization order parameters are not scalars but matrices. The
renormalization order parameter after full averaging is diagonalized
by the unitary matrix which diagonalizes the input-layer readout covariance
matrix, Eq.\ref{eq:B0-2}. Different eigenvalues corresponding to
different modes obey an independent set of equations, analogous to
\citep{saxe2019mathematical}, which showed modes evolving independently
with time during GD learning. However, our renormalization modes are
defined by diagonalizing input-layer mean squared readout matrices,
and not by a fixed input-output covariance matrix as in \citep{saxe2019mathematical}.
As stated above, our results rely on the equilibrium assumption but
not on the special structure of the data nor on an initial condition. 

It is interesting to explore the similarity of the behavior of our
system with the properties of gradient descent, with implicit regularization
induced by early stopping starting from random initial conditions
\citep{dodier1996geometry,li2020gradient,advani2020high}. As we show
in SM VIII, the generalization properties of the early stopping dynamics
may exhibit qualitatively similar features to those predicted by our
theory for Gibbs learning, with the initial variance of the weights
in the early stopping dynamics playing the role of our noise parameter
$\sigma^{2}$. For example, the generalization error of weights learned
through the early stopping dynamics increases with the network width
for large initial weight variance, and decreases with the network
width for small initial weight variance, which qualitatively agrees
with the behavior of the generalization error in our BPKR theory in
different regimes of the noise parameter $\sigma$. Extending our
theory to the learning dynamics is an interesting ongoing study.

\textbf{Nonlinear DNNs:} We have extended our theory to ReLU networks
by applying a scalar kernel renormalization scheme on the GP nonlinear
kernels (Section $\text{\ref{sec:Deep-ReLu-Networks}}$, Eqs.$\text{\ref{eq:meanfiteration-relu}}$-$\ref{eq:b0-1}$).
Testing this approximation against numerical simulations of a few
learning tasks with ReLU networks with a moderate number of layers
revealed strikingly good qualitative and quantitative agreement regarding
the width and noise dependencies of the predictor statistics and the
generalization error, as well as the layerwise mean squared readout
order parameters – with much greater accuracy than the GP theory.

Importantly, this BPKR approximate theory for ReLU networks holds,
even in cases where for a linear network the system would be in a
highly under-parameterized regime (Fig.$\text{\ref{fig:alpha0>1}}$),
such that the neuronal nonlinearity plays a crucial role in the ability
of the system to yield zero training and low generalization error
(Fig.$\text{\ref{fig:deeprelu}})$. The failure of the approximation
for deeper networks ($L\geq5$) is expected. The GP theory for nonlinear
networks predicts that as $L\rightarrow\infty$ not only the magnitude
of the kernel matrices converges to a (finite or infinite) fixed point
but also its matrix structure converges to a fixed point, implying
the loss of information about the structure of the inputs in deep
networks with infinite width. Thus, when the width is finite, i.e.,
$\alpha=\text{\ensuremath{\mathcal{O}}}(1)$, we expect to see a renormalization
not only of the kernels' magnitudes (as in our scalar renormalization)
but also in their shape. In addition, in nonlinear networks with finite
width, the basic description of the system may depend on higher-order
statistics than the kernel matrices, as suggested by the recent work
of \citep{naveh2020predicting} and \citep{antognini2019finite}.

Even for shallow nonlinear networks, the approximate nonlinear BPKR
breaks down in the under-parameterized regime, on the left side of
the interpolation threshold at $N\approx\alpha_{0}$. Thus, if $\alpha_{0}$
is large, there is a substantial range of small $N$ for which the
system is in the under-parameterized regime, and this gives rise to
a peak in the generalization error (as a function of $N)$ near the
interpolation threshold – a genuine `double descent' phenomena as
studied in \citep{belkin2019reconciling,mei2019generalization}. Naturally,
our approximate theory predicts monotonic dependence in $N$, hence
it is valid only on the right side of the double descent peak, i.e.,
in the over-parameterized regime, Fig.$\text{\ref{fig:doubledescent}}$. 

\textbf{Relation to other methods:} Successive integration of random variables of joint distributions is used in belief propagation algorithms \citep{yedidia2000generalized,yedidia2003understanding,yedidia2005constructing,pearl1986fusion,pearl2014probabilistic,weiss2010belief,mezard2009information}. However, despite the Markovian property of the distribution of the deep network activations, the posterior distribution of the weights takes a complicated form, as described in Section \ref{subsec:Statistical-mechanics-of} (Eqs.\ref{eq:H_L0T},\ref{eq:H_L}), rendering the layer-wise weight integration intractible in general, and even in linear networks can be performed only in the thermodynamic limit, as shown here. 
Therefore, although Bayesian inference
algorithms such as message passing are commonly applied to study the
distribution of hidden-layer activations of Bayesian Neural Networks
\citep{winn2005variational,parr2019neuronal,kschischang2001factor},
they are not directly applicable for computing the posterior distribution
of the weights. Recent works on inference of the posterior weight
distribution have proposed to extend backprop learning algorithms
to update also the variances of the weights, by approximating the
weight distribution as an independent Gaussian distribution \citep{hernandez2015probabilistic,graves2011practical}.
As our work shows, the posterior distribution is far from being i.i.d.
Gaussian. Importantly, backprop learning algorithms do not necessarily
provide insight into the final solutions. In contrast, our work is
a theoretical study of the properties of the posterior distribution
of weights after learning. 

Our BPKR also has some analogy with the Renormalization Group (RG)
approach in physics. Similar to BPKR, RG evaluates properties of high-dimensional
systems by successive integration of subsets of the systems' DoFs
\citep{goldenfeld2018lectures}. However, the analogy is limited because
in contrast to RG, here there is no obvious notion of coarse graining
of DoFs. Our system combines properties of layered physical systems
\citep{chen1996renormalization,pierson1992renormalization,pierson1994critical,li2018neural}
with mean field aspect arising from the full layer-to-layer connectivity.
The latter is demonstrated by the fact that the behavior at the critical
point $\sigma^{2}(1-\alpha)=1$ is mean-field-like, see Eq. \ref{eq:ev0}. 

\textbf{Extensions of present work:} There are several paths for extending
our theory to deeper nonlinear networks. Exact mean-field equations
are possible for specific forms of nonlinearities. For a generic nonlinearity,
approximate methods might be possible. These methods would likely
involve renormalization not only of kernels but also of other terms
in the effective Hamiltonian, such as 4-th order kernels \citep{naveh2020predicting}.
These are topics of on going work. 

Our theory applies to fully connected networks without additional
constraints on the network structure, while in practice, other types
of neural networks such as Convolutional Neural Networks (CNNs) are
commonly used for image processing, speech recognition and various
tasks. Recent work discussed extension of the GP theory to CNNs \citep{van2017convolutional,garriga2018deep,novak2018bayesian}.
Incorporating such architectural restrictions into our theory induces
shape renormalization of the kernel (i.e., not simply renormalization
by a scalar) and is a topic of ongoing work. Other extensions of our
theory include loss functions other than MSE and regularization terms
other than $L_{2}$.

\textbf{Acknowledgements:}

We have benefitted from helpful discussions with Andrew Saxe, Gadi
Naveh and Zohar Ringel and useful comments on the manuscript from
Itamar Landau, Dar Gilboa, Haozhe Shan, and Jacob Zavatone-Veth. This research is partially
supported by the Swartz Program in Theoretical Neuroscience at Harvard,
the NIH grant from the NINDS (1U19NS104653) and the Gatsby Charitable
Foundation.

\appendix

\section*{Appendix}

\section{\label{subsec:appendixa}The Back-Propagating Kernel Renormalization
for DLNNs}

We begin with the partition function

\begin{multline}
Z=\int d\Theta\mathcal{}\\
\exp[-\frac{\beta}{2N}\sum_{\mu=1}^{P}(\sum_{i=1}^{N}a_{i}\phi_{i}(x^{\mu},W)-y^{\mu})^{2}-\frac{1}{2\sigma^{2}}\Theta^{T}\Theta]\label{eq:partitionfunction}
\end{multline}
and introduce $P$ auxiliary integration variables, $t^{\mu}(\mu=1,\cdots,P)$
to linearize the quadratic training error. 
\begin{multline}
Z=\int d\Theta\int\Pi_{\mu}^{P}dt_{\mu}\exp[-\frac{1}{2\sigma^{2}}\Theta^{\top}\Theta\\
-\sum_{\mu=1}^{P}it_{\mu}(\frac{1}{\sqrt{N}}\sum_{i=1}^{N}a_{i}\phi_{i}(x^{\mu},W)-y^{\mu})-\frac{T}{2}t^{\top}t]\label{eq:full Z}
\end{multline}
Integrating over $a$, we have $Z=\int dWZ_{L}(W)$ with 
\begin{multline}
Z_{L}(W)=\int dt\exp[-\frac{1}{2}t^{\top}(K_{L}+TI)t\\
+it^{\top}Y-\frac{1}{2\sigma^{2}}\mathrm{Tr}(W^{\top}W)]\label{eq:integratea}
\end{multline}
where the kernel matrix $K_{L}$ is defined in Eq.\ref{eq:layerkernels}
with $l=L$. Integrating over $t$ yields, 
\begin{multline}
Z_{L}=\exp[-\frac{1}{2}Y^{\top}(K_{L}+TI)^{-1}Y\\
-\frac{1}{2}\log\det(K_{L}+TI)-\frac{1}{2\sigma^{2}}\mathrm{Tr}(W^{\top}W)]
\end{multline}
To make further progress we will assume all the units are linear,
so that the hidden units are $x_{i,l}=\frac{1}{\sqrt{N}}w_{l}^{i\top}x_{l-1}$
(and the first layer units are $x_{i,1}=\frac{1}{\sqrt{N_{0}}}w_{1}^{i\top}x$).
We evaluate $Z=\int dWZ_{L}(W)$ by successive integrations of weight
matrices one at a time, starting from the top layer. Integrating the
top hidden-layer weights to compute $Z_{L-1}(W')=\int dW_{L}Z_{L}(W)=\exp[-H_{L-1}]$,
where the weights $W'$ consist of all weight matrices upstream of
$W_{L}$, $W'=\{W_{k}\}_{k<L}$, obtaining 
\begin{multline}
Z_{L-1}(W')=\int\Pi_{i=1}^{N_{L}}dw_{L}^{i}\int dt\exp[-\frac{1}{2}t^{\top}(K_{L}+TI)t\\
-\frac{1}{2\sigma^{2}}\mathrm{Tr}(W^{\top}W)+it^{\top}Y-\frac{1}{2\sigma^{2}}\mathrm{Tr}(W'^{\top}W')]\\
=\int dt\exp[it^{\top}Y+NG(t)-\frac{T}{2}t^{\top}t-\frac{1}{2\sigma^{2}}\mathrm{Tr}(W'^{\top}W')]
\end{multline}
\begin{equation}
G(t)=\log\left\langle \exp-\frac{1}{2N}t^{\top}K_{w}^{L}t\right\rangle _{w}\label{eq:w-average}
\end{equation}
where the average is w.r.t. to a \emph{single} $N$-dimensional weight
vector $w_{L}^{i}$ with i.i.d. $\mathcal{N}(0,\sigma)$ components,
and $K_{w}^{L,\mu\nu}=\sigma^{2}x_{i,L}^{\mu}x_{i,L}^{\nu}=\frac{\sigma^{2}}{N}x_{L-1}^{\mu\top}w_{L}^{i}w_{L}^{i\top}x_{L-1}^{\nu}$.
Performing the average in \ref{eq:w-average}, yields $G(t)=-\frac{1}{2}\log(1+h_{L-1})$
where

\begin{equation}
h_{L-1}=\frac{\sigma^{2}}{N}t^{\top}K_{L-1}t\label{eq:hL-1 AppA}
\end{equation}
To integrate over $t$, we enforce the identity Eq.\ref{eq:hL-1 AppA},
by Fourier representation of the delta function, introducing the auxiliary
variable, $u_{L-1}$, 
\begin{multline}
Z_{L-1}=\int du_{L-1}\int_{-1}dh_{L-1}\int dt\\
\exp[it^{\top}Y-\frac{N}{2}\log(1+h_{L-1})+\frac{N_{L}}{2\sigma^{2}}u_{L-1}h_{L-1}\\
-\frac{1}{2}t^{\top}(u_{L-1}K_{L-1}+TI)t-\frac{1}{2\sigma^{2}}\mathrm{Tr}(W'^{\top}W')]
\end{multline}
and integrating over $t$, 
\begin{multline}
Z_{L-1}=\int du_{L-1}\int_{-1}dh_{L-1}\exp[-\frac{N}{2}\log(1+h_{L-1})\\
+\frac{1}{2\sigma^{2}}Nu_{L-1}h_{L-1}-\frac{1}{2}Y^{T}(u_{L-1}K_{L-1}+TI)^{-1}Y\\
-\frac{1}{2}\log\det(u_{L-1}K_{L-1}+TI)-\frac{1}{2\sigma^{2}}\mathrm{Tr}(W'^{\top}W')]\label{eq:ZL_1_Appendix}
\end{multline}
In the limit of $N\rightarrow\infty,$$P\rightarrow\infty,\text{ and fixed }\alpha$,
we solve this integral with the saddle-point method. One of the saddle-point
equations yields $u_{L-1}=\frac{\sigma^{2}}{1+h_{L-1}}$, plugging
back in Eq.\ref{eq:ZL_1_Appendix}we obtain

\begin{equation}
Z_{L-1}(W')=\int du_{L-1}\exp-H_{L-1}(W',u_{L-1})
\end{equation}
with the effective Hamiltonian

\begin{multline}
H_{L-1}(W',u_{L-1})=\frac{1}{2}Y^{\top}(u_{L-1}K_{L-1}+TI)^{-1}Y\\
-\frac{N}{2}\log u_{L-1}+\frac{1}{2}\log\det(K_{L-1}u_{L-1}+TI)\\
+\frac{1}{2\sigma^{2}}\text{Tr}\ensuremath{W'^{\top}W'}+\frac{1}{2\sigma^{2}}Nu_{L-1}
\end{multline}
Thus, integrating over $W_{L}$ resulted in the presence of an auxiliary
scalar DOF, $u_{L-1}$. Finally, we eliminate $u_{L-1}$ through a
saddle-point equation,

\begin{multline}
N(1-\sigma^{-2}u_{L-1})\\
=-Y^{T}(u_{L-1}K_{L-1}+TI)^{-2}u_{L-1}K_{L-1}Y\\
+\mathrm{Tr}(u_{L-1}K_{L-1}+TI)^{-1}u_{L-1}K_{L-1}\label{eq:sp_1}
\end{multline}
At the $T\rightarrow0$ limit, we obtain Eq.$\text{\ref{eq:u_L-1}}$.

This procedure can be iterated layer-by-layer. We demonstrate it by
computing $H_{L-2}(W')$ defined via $Z_{L-2}(W'')=\int dW_{L-1}Z_{L-1}(W')=\exp[-H_{L-2}(W'')]$
where $W''$ denotes all weight matrices upstream of $W_{L-1}$, 
\begin{multline}
Z_{L-2}(W'')=\int du_{L-1}\int dW_{L-1}\exp[\frac{N}{2}\log u_{L-1}\\
-\frac{1}{2\sigma^{2}}Nu_{L-1}-\frac{1}{2}Y^{T}(u_{L-1}K_{L-1}+TI)^{-1}Y\\
-\frac{1}{2}\log\det(u_{L-1}K_{L-1}+TI)-\frac{1}{2\sigma^{2}}\mathrm{Tr}(W'^{\top}W')]\\
=\int du_{L-1}\int dt\exp[it^{\top}Y+NG(t)-\frac{T}{2}t^{\top}t\\
-\frac{1}{2\sigma^{2}}\mathrm{Tr}(W''^{\top}W'')+\frac{N}{2}\log u_{L-1}-\frac{1}{2\sigma^{2}}Nu_{L-1}]\label{eq:secondintegration}
\end{multline}

\begin{equation}
G(t)=\log\left\langle \exp-\frac{u_{L-1}}{2N}t^{\top}K_{w}^{L-1}t\right\rangle _{w}
\end{equation}
where the average is w.r.t. a single $N$-dimensional Gaussian vector
with i.i.d. $\mathcal{N}(0,\sigma$) components.

Performing this average, yields $G(t)=-\log(1+h_{L-2})$ with

\begin{equation}
h_{L-2}=\frac{\sigma^{2}u_{L-1}}{N_{L-1}}t^{\top}K_{L-2}t,\label{eq:eq:h_L-2}
\end{equation}
Similar to above, we introduce two additional scalar integration variables
$u_{L-2}$ and $h_{L-2}$,

\begin{multline}
Z_{L-2}=\int du_{L-2}\int_{-1}dh_{L-2}\int du_{L-1}\int dt\\
\exp[it^{\top}Y-\frac{N}{2}\log(1+h_{L-2})+\frac{N}{2\sigma^{2}}u_{L-2}h_{L-2}\\
-\frac{1}{2}t^{\top}(u_{L-1}u_{L-2}K_{L-1}+TI)t+\frac{N}{2}\log u_{L-1}\\
-\frac{1}{2\sigma^{2}}Nu_{L-1}-\frac{1}{2\sigma^{2}}\mathrm{Tr}(W''^{\top}W'')]
\end{multline}
Integrate over $t$ and plugging the saddle point of $h_{L-2}$ ($u_{L-2}=\frac{\sigma^{2}}{h_{L-2}+1}$),
we have the effective Hamiltonian 
\begin{multline}
H_{L-2}(W'',u_{L-1},u_{L-2})=\\
-\frac{N}{2}\log u_{L-1}+\frac{1}{2\sigma^{2}}\text{Tr}\ensuremath{W''^{\top}W''}+\frac{1}{2\sigma^{2}}Nu_{L-1}\\
+\frac{1}{2}Y^{\top}(u_{L-1}u_{L-2}K_{L-2}+TI)^{-1}Y-\frac{N}{2}\log u_{L-2}\\
+\frac{1}{2}\log\det(K_{L-2}u_{L-1}u_{L-2}+TI)+\frac{1}{2\sigma^{2}}Nu_{L-2}
\end{multline}
Finally, $u_{L-1}$ and $u_{L-2}$ are computed via saddle-point equations
\begin{align}
 & N(1-u_{L-1}\sigma^{-2})=\label{eq:op1}\\
 & -u_{L-1}u_{L-2}Y^{T}(u_{L-1}u_{L-2}K_{L-2}+TI)^{-2}K_{L-2}Y\nonumber \\
 & +u_{L-1}u_{L-2}Tr(u_{L-1}u_{L-2}K_{L-2}+TI)^{-1}K_{L-2}\nonumber \\
 & N(1-u_{L-2}\sigma^{-2})\label{eq:op2}\\
 & =-u_{L-1}u_{L-2}Y^{T}(u_{L-1}u_{L-2}K_{L-2}+TI)^{-2}K_{L-2}Y\nonumber \\
 & +u_{L-1}u_{L-2}Tr(u_{L-1}u_{L-2}K_{L-2}+TI)^{-1}K_{L-2}\nonumber 
\end{align}
The solution obeys $u_{L-1}=u_{L-2},$and we now have

\begin{equation}
Z=\int dW''\int du_{L-2}\exp-H_{L-2}(W'',u_{L-2})
\end{equation}
with 
\begin{multline}
H_{L-2}(W'',u_{L-2})=-N\log u_{L-2}+\frac{1}{2\sigma^{2}}\text{Tr}\ensuremath{W''^{\top}W''}\\
+\frac{1}{\sigma^{2}}Nu_{L-2}+\frac{1}{2}Y^{\top}(u_{L-2}^{2}K_{L-2}+TI)^{-1}Y\\
+\frac{1}{2}\log\det(K_{L-1}u_{L-2}^{2}+TI)
\end{multline}
Evaluating $u_{L-2}$ via the saddle-point equation yields

\begin{align}
 & N(1-u_{L-2}\sigma^{-2})\label{eq:sp L-2}\\
 & =-Y^{T}(u_{L-2}^{2}K_{L-2}+TI)^{-2}u_{L-2}^{2}K_{L-2}Y\nonumber \\
 & +\mathrm{Tr}[(u_{L-2}^{2}K_{L-2}+TI)^{-1}u_{L-2}^{2}K_{L-2}]\nonumber 
\end{align}
where the kernel $K_{L-2}$ is renormalized by $u_{L-2}^{2}$. Note
that in the integration of $W_{L-2}$ both $u_{L-1}$and $u_{L-2}$
are auxiliary integration variables (hence independent of weights)
and are determined at the \emph{last} step by the new saddle-point
equation Eq.\ref{eq:sp L-2} as functions of $W''$. In contrast,
the saddle-point value of $u_{L-1}$ in the \emph{first }renormalization
step, Eq.\ref{eq:sp_1} is a function of $W'$ . In fact the \emph{average}
of $u_{L-1}$ of the first renormalization step over $W_{L-1}$ obeys
$u_{L-1}=u_{L-2}$ of the second renormalization step, see paragraph
below on renormalization of order parameters.

Similarly, iterating this renormalization $l$ times, yields Eqs.$\text{\ref{eq:H_L-l-1}}$,\ref{eq:u_L-l-1}.

\textbf{Narrow network at zero temperature:} At finite temperature
the above derivation holds for all $\alpha$. However, in the zero-temperature
limit, we need to address the singularity of the hidden layers' kernel
matrices when $\alpha>1$. We begin with the partition function after
integrating the readout layer at zero temperature, 
\begin{multline}
Z_{L}(W)=\int dt\exp[-\frac{1}{2}t^{\top}K_{L}t+it^{\top}Y\\
-\frac{1}{2\sigma^{2}}\mathrm{Tr}(W^{\top}W)]
\end{multline}
With eigenvalue decomposition of $K_{L}$, $K_{L}=V\Sigma V^{\top}$,
where $V$ is a unitary $P\times P$ matrix, and $\Sigma$ is a $P\times P$
diagonal matrix with elements $(\Sigma_{1},\cdots,\Sigma_{N},0,\cdots,0)$,
and orthogonal transformation of variables $V^{\top}t\rightarrow t$,
we have 
\begin{multline}
Z_{L}(W)=\int dt\exp[-\frac{1}{2}t^{\top}\Sigma t+it^{\top}V^{\top}Y\\
-\frac{1}{2\sigma^{2}}\mathrm{Tr}(W^{\top}W)]
\end{multline}
We introduce notations $t_{||}=[t_{1,}\cdots,t_{N}]^{\top}\in\mathbb{R}^{N}$,
$t_{\perp}=[t_{N+1},\cdots,t_{P}]^{\top}\in\mathbb{R}^{N-P}$, $V_{||}=[V_{1},\cdots,V_{N}]\in\mathbb{R}^{P\times N}$,
$V_{\perp}=[V_{N+1},\cdots,V_{P}]\in\mathbb{R}^{P\times(P-N)}$, $\Sigma_{||}=\mathrm{diag}(\Sigma_{1},\cdots,\Sigma_{N})\in\mathbb{R}^{N\times N}$
. With these notations we can write 
\begin{multline}
Z_{L}(W)=\int dt_{\parallel}\int dt_{\perp}\exp[-\frac{1}{2}t_{||}^{\top}\Sigma_{||}t_{||}+it_{||}^{\top}V_{||}^{\top}Y\\
+it_{\perp}^{\top}V_{\perp}^{\top}Y-\frac{1}{2\sigma^{2}}\mathrm{Tr}(W^{\top}W)]
\end{multline}
Integrating over $t_{\perp}$ yields $\delta(V_{\perp}^{\top}Y)$.
The $\delta$-function enforces the projection of $Y$ onto the directions
perpendicular to $X_{L}$ to vanish. In the zero-temperature limit,
this constraint on the weights ensures zero training error, therefore
$Y$ must lie in the subspace spanned by $X_{L}$. We next integrate
$t_{||}$, and obtain $Z_{L}(W)=\delta(V_{\perp}Y)\exp[-H_{L}(W)]$,
with 
\begin{multline}
H_{L}(W)=\frac{1}{2}Y^{\top}K_{L}^{+}Y+\frac{1}{2}\log\det(C_{L})\\
+\frac{1}{2\sigma^{2}}\mathrm{Tr}(W^{\top}W)
\end{multline}
where $K_{L}^{+}=V_{||}\Sigma_{||}^{-1}V_{||}^{\top}$ is the pseudo-inverse
of $K_{L}$, and $C_{L}=\frac{\sigma^{2}}{N}X_{L}X_{L}^{\top}$ has
the same determinant as $\Sigma_{||}$ .

Similarly we have $Z_{L-l}(W)=\delta(V_{\perp}Y)\exp[-H_{L-l}(W)]$,
here $V_{\perp}$ are the eigenvectors of $K_{L-l}$ spanning its
null space, and 
\begin{multline}
H_{L-l}(W')=\frac{1}{2u_{L-l}^{l}}Y^{\top}K_{L-l}^{+}Y+\frac{1}{2}\log\det(u_{L-l}^{-l}C_{L-l})\\
+\frac{1}{2\sigma^{2}}\mathrm{Tr}(W'^{\top}W')\label{eq:hamiltoniannarrow}
\end{multline}
Differentiating Eq.$\text{\ref{eq:hamiltoniannarrow}}$ w.r.t. $u_{L-l}$
, we obtain Eqs.$\ref{eq:u_L-l-2-2},\ref{eq:b_l-1-2}$.

\textbf{Renormalization of the order parameters: }Here we show that
the order parameters $u_{l}$ undergo a trivial renormalization upon
averaging. For any function of $u_{l}$, we can write, 
\begin{multline}
\langle f(u_{l})\rangle_{l}=\frac{1}{Z_{l-1}}\int du_{l}f(u_{l})\int dW_{l}\int dt\exp[it^{\top}Y\\
-\frac{1}{2}t^{\top}(u_{l}^{L-l}K_{l}+TI)t+\frac{N(L-l)}{2}\log u_{l}\\
-\frac{N(L-l)}{2\sigma^{2}}u_{l}+\frac{1}{2\sigma^{2}}\mathrm{Tr}(W'^{\top}W')]\\
=\frac{1}{Z_{l-1}}\int du_{l}\int du_{l-1}\int dtf(u_{l})\exp[it^{\top}Y\\
-\frac{1}{2}t^{\top}(u_{l-1}u_{l}^{L-l}K_{l-1}+TI)t+\frac{N(L-l)}{2}\log u_{l}\\
-\frac{N(L-l)}{2\sigma^{2}}u_{l}+\frac{1}{2\sigma^{2}}\mathrm{Tr}(W''^{\top}W'')\\
+\frac{N}{2}\log u_{l-1}-\frac{N}{2\sigma^{2}}u_{l-1}]
\end{multline}
which is equal to the saddle-point value of $f(u_{l})$. Since $u_{l}=u_{l-1}$
at the saddle point, where $u_{l-1}$ obeys the saddle point Eq.\ref{eq:u_L-l-1}
appropriate for $L-l+1$ iterations, we have 
\begin{equation}
\langle f(u_{l})\rangle_{l}=f(u_{l-1})\label{eq:iterateU-1}
\end{equation}
which holds for all $0\leq\alpha<\infty$ and all $T$.

\textbf{Interpretation of the order parameters:} The order parameters
$u_{l}$ have a simple interpretation given by Eq.\ref{eq:u vs r-1}
for $1\leq l\leq L$ at zero temperature for $\alpha<1$(See SM IVA
for the derivation at finite temperature). We evaluate 
\begin{multline}
\text{\ensuremath{\frac{1}{P}}}\langle Y^{\top}K_{l}^{-1}Y\rangle_{l}\\
=\frac{1}{Z_{l-1}}\int dW_{l}\frac{1}{P}Y^{\top}K_{l}^{-1}Y\\
\times\int dt\exp[it^{\top}Y-\frac{1}{2}t^{\top}u_{l}^{L-l}K_{l}t+\frac{1}{2\sigma^{2}}\mathrm{Tr}(W'^{\top}W')]\\
=-\frac{1}{Z_{l-1}}\int dW_{l}\frac{1}{P}Y^{\top}\int dtitu_{l}^{L-l}\\
\exp[it^{\top}Y-\frac{1}{2}t^{\top}u_{l}^{L-l}K_{l}t+\frac{1}{2\sigma^{2}}\mathrm{Tr}(W'^{\top}W')]
\end{multline}
Performing integration over $W_{l}$ with the same approach we used
to compute the partition function $Z_{l-1}$ above, and introducing
the same order parameter $u_{l-1}$, we reduce the above expression
to

\begin{multline}
-\frac{1}{Z_{l-1}}\int du_{l-1}\frac{1}{P}Y^{\top}\int dtitu_{l-1}^{L-l}\\
\exp[it^{\top}Y+\frac{N(L-l+1)}{2}\log u_{l-1}\\
-\frac{1}{2}t^{\top}u_{l-1}^{L-l+1}K_{l-1}t-\frac{N(L-l+1)}{2\sigma^{2}}u_{l-1}]\\
=\frac{1}{P}u_{l-1}^{-1}Y^{\top}K_{l-1}^{-1}Y
\end{multline}
At zero temperature for $\alpha<1$, the expression leads to Eq.\ref{eq:u vs r-1}
for all $l$ ($1\leq l\leq L$). Note that this result is conditioned
on the upstream weights $\{W_{k}\}_{k<l}$.

For $\alpha>1$ at zero temperature, the OP obeys Eq.$\text{\ref{eq:u vs r-1}}$
for $1\leq l<L$ (partial averaging of the weights), but the relation
is replaced by Eq.$\text{\ref{eq:opinterpretation}}$ for averaging
over all hidden weights. The details of these results are delegated
to the SM ${\rm IA}$.

\textbf{Mean squared readout weights: }From Eq.$\text{\ref{eq:full Z}}$,
it follows that the $W$-dependent average of the readout weights
is 
\begin{equation}
\langle a\rangle=-\frac{\sigma^{2}}{\sqrt{N}}i\Phi^{\top}\langle t\rangle
\end{equation}
The statistics of $t$ can be obtained from Eq.$\text{\ref{eq:integratea}}$,
\begin{equation}
\langle t\rangle=i(K_{L}+TI)^{-1}Y
\end{equation}
Therefore we have $\langle a\rangle=\frac{\sigma^{2}}{\sqrt{N}}\Phi^{\top}(K_{L}+TI)^{-1}Y$,
and $\langle a\rangle^{\top}\langle a\rangle=\sigma^{2}Pr_{L}=\sigma^{2}Y^{\top}(K_{L}+TI)^{-2}K_{L}Y$.
In the zero-temperature limit, for $\alpha<1$, $\langle a\rangle^{\top}\langle a\rangle=\sigma^{2}Y^{\top}K_{L}^{-1}Y$,
for $\alpha>1$, $\langle a\rangle^{\top}\langle a\rangle=\sigma^{2}Y^{\top}K_{L}^{+}Y$.

Similarly, we can define $a_{l}$ as the readout weight vector trained
with inputs from the $l$-th layer of the trained network to produce
the target output $Y$, we can obtain the statistics of $a_{l}$ by
simply replacing the $K_{L}$ in the above equations with $K_{l}$.
At zero temperature, we have $\langle a_{l}\rangle^{\top}\langle a_{l}\rangle=\sigma^{2}Y^{\top}K_{l}^{-1}Y$
for $\alpha<1$, and $\langle a_{l}\rangle^{\top}\langle a_{l}\rangle=\sigma^{2}YK_{l}^{+}Y$
for $\alpha>1$. In Eqs.$\text{\ref{eq:b_l}}$,$\ref{eq:b_l-1-2}$,
the definition of $r_{l}$ is equivalent to $r_{l}=\frac{\sigma^{-2}}{P}\langle a_{l}\rangle^{\top}\langle a_{l}\rangle$,
therefore we name $r_{l}$ as the \textbf{mean squared layer readout.}

The second-order statistics of $a_{l}$, including its variance and
its norm, are discussed further in SM IB.

\textbf{Order parameter at the $L\rightarrow\infty$ limit: }Earlier
in this section we introduced the detailed derivation of the self-consistent
equation for the order parameter at finite temperature, given by Eq.$\text{\ref{eq:u_L-l-1}}$.
At the $L\rightarrow\infty$ limit, in the low-noise regime $\sigma^{2}(1-\alpha)<1$,
$u_{0}$ approaches 1. We asssume that $u_{0}$ goes to 1 as $u_{0}\approx1-\frac{v_{0}}{L}$,
as we discussed in Section $\text{\ref{sec:Generalization}}$ for
zero temperature. Plugging in Eq.$\text{\ref{eq:u_L-l-1}}$, we have
\begin{align}
 & 1-\sigma^{-2}\nonumber \\
 & =-\frac{1}{N}Y^{\top}(\exp(-v_{0})K_{0}+TI)^{-2}\exp(-v_{0})K_{0}Y\label{eq:v0finiteT}\\
 & +\frac{1}{N}\mathrm{Tr}((\exp(-v_{0})K_{0}+TI)^{-1}\exp(-v_{0})K_{0})
\end{align}
This self-consistent equation determines $\exp(v_{0})$, which is
the limit of $\lambda$ as $L\rightarrow\infty$ as we present in
Fig.$\text{\ref{fig:egvsT}}$(b,c) in Section $\text{\ref{subsec:Finite-temperature}}$.

\section{\label{subsec:Generalization}Generalization}

The mean squared generalization error depends only on the mean and
variance of the predictor, and they can be computed using the following
generating function 
\begin{align}
Z(t_{P+1}) & =\int D\Theta\nonumber \\
 & \exp[-\frac{\beta}{2}\sum_{\mu=1}^{P}(\frac{1}{\sqrt{N}}\sum_{i=1}^{N}a_{i}\phi_{i}(x^{\mu},W)-y^{\mu})^{2}\nonumber \\
 & +it_{P+1}\frac{1}{\sqrt{N}}\sum_{i=1}^{N}a_{i}\phi(W,x)-\frac{T}{2\sigma^{2}}\Theta^{\top}\Theta]
\end{align}
where $x$ is an arbitrary new point. The statistics of the predictor
are given by 
\begin{align}
\langle f(x)\rangle & =\partial_{it_{P+1}}\log Z|_{t_{P+1}=0}\\
\langle\delta^{2}f(x)\rangle & =\partial_{it_{P+1}}^{2}\log Z|_{t_{P+1}=0}
\end{align}
The integral can be performed similarly as in Appendix $\text{\ref{subsec:appendixa}}$
by introducing $P$ auxiliary integration variables, $t^{\mu}(\mu=1,\cdots,P)$,
integrating over $W$ and introducing order parameters $u_{l}$'s
layer-by-layer.

After integrating the weights of the entire network, we obtain 
\begin{multline}
Z(t_{p+1})=\int du_{0}\exp[\frac{NL}{2}\log u_{0}-\frac{NL}{2\sigma^{2}}u_{0}\\
+\frac{1}{2}(iY+t_{P+1}^{\top}u_{0}^{L}k_{0}(x))^{\top}(u_{0}^{L}K_{0}+TI)^{-1}(iY+t_{P+1}^{\top}u_{0}^{L}k_{0}(x))\\
-\frac{1}{2}\log\det(u_{0}^{L}K_{0}+TI)-\frac{1}{2}t_{P+1}^{\top}u_{0}^{L}K_{0}(x,x)t_{P+1}]
\end{multline}
where $T_{0}=u_{0}^{-L}T$, as defined in Section $\text{\ref{subsec:Finite-temperature}}$,
Differenitating $Z$ we obtain 
\begin{multline}
\langle f(x)\rangle=\partial_{it_{p+1}}\log Z|_{t_{p+1}=0}\\
=u_{0}^{L}k_{0}^{\top}(x)(u_{0}^{L}K_{0}+TI)^{-1}Y\label{eq:finiteTmean}
\end{multline}
Because the derivative is evaluated at $t_{P+1}=0$, the saddle point
$u_{0}$ satisfies the same equation as Eq.$\text{\ref{eq:u_L-l-1}}$
for $l=L$. Similarly, we calculate the second-order statistics 
\begin{multline}
\langle\delta^{2}f(x)\rangle=\partial_{it_{P+1}}^{2}\log Z|_{t_{p+1}=0}\\
=u_{0}^{L}K_{0}(x,x)-u_{0}^{L}k_{0}^{\top}(x)(u_{0}^{L}K_{0}+TI)^{-1}u_{0}^{L}k_{0}(x)\label{eq:finiteTvar}
\end{multline}
Taking the $T\rightarrow0$ limit we obtain Eq.$\text{\ref{eq:meanfiteration}}$
and Eq.$\text{\ref{eq:varfiteration}}$.

The dependence of the generalization error w.r.t. $\sigma$, $N$
and $L$ is determined by the behavior of $\sigma^{2}u_{0}$ w.r.t.
$\sigma$, $N$ and $L$, which is shown in SM $\ensuremath{\mathrm{IIA}},\text{{\rm IIB, IIC}}$.
The dependence on $P$ (which affects both $\alpha$ and $\alpha_{0}$)
hinges on the specific statistics of input and output. Here we analyze
the relatively simple case of input sampled from i.i.d. Gaussian distribution,
and target output generated by a linear teacher with additive noise,
and we focus on the behavior near $\alpha_{0}=1$.

For $P<N_{0}$, we first consider $r_{0}$, averaged over the linear
teacher noise. Near $\alpha_{0}=1$, since $\mathrm{Tr}K_{0}^{-1}$
diverges as $(1-\alpha_{0})^{-1}$ (see \citep{bai2010sample}), $r_{0}$
is dominated by the contribution from the noise term in the target
noisy teacher output $Y$, and yields $r_{0}\sim\sigma_{0}^{2}(1-\alpha_{0})^{-1}$,
where $\sigma_{0}$ denotes the amplitude of the teacher's noise.

Since $r_{0}$ is divergent as $\alpha_{0}\rightarrow1$, keeping
the dominant terms in Eq.$\text{\ref{eq:u0}}$, we obtain $u_{0}\sim(\sigma^{2}\alpha r_{0})^{1/L+1}\sim\alpha{}^{1/L+1}(1-\alpha_{0})^{-1/L+1}$,
thus $u_{0}^{L}$ diverges as $\alpha^{L/L+1}(1-\alpha_{0})^{-L/L+1}.$

The contribution of the squared mean predictor $\langle f(x)\rangle^{2}$
to $\varepsilon_{g}$ averaged over the test sample $x$ and the linear
teacher noise is given by the corresponding averages of $Y^{\top}K_{0}^{-1}k_{0}k_{0}^{\top}K_{0}^{-1}Y$.
Assuming $\langle xx^{\top}\rangle=\gamma I$, then the divergent
contribution, similar to $r_{0}$, comes from the noise in the linear
teacher, and is given by $\gamma\sigma_{0}^{2}\alpha_{0}(1-\alpha_{0})^{-1}$.

Since $\mathrm{Tr}(\sigma^{2}N_{0}^{-1}XK_{0}^{-1}X^{\top})$ scales
with the rank of $K_{0}$ and grows as $\alpha_{0}$, the $K_{0}(x,x)-k_{0}^{\top}K_{0}^{-1}k_{0}$
term in the variance of the predictor vanishes as $1-\alpha_{0}$
, and thus the variance vanishes as $\alpha^{L/L+1}(1-\alpha_{0})^{1/L+1}$,
as $\alpha_{0}\rightarrow1$. Therefore, the generalization error
is dominated by the divergent bias as $\alpha_{0}\rightarrow1$, and
diverges as $\alpha_{0}/(1-\alpha_{0}).$

For $P>N_{0}$, because now the network cannot achieve zero training
error, we replace the $Y$ in $r_{0}$ with $X^{\top}(XX^{\top})^{-1}XY$,
which is the output the network actually learns on the training data.
Near $\alpha_{0}=1$, since $\frac{1}{P}\mathrm{Tr}((XX^{\top})^{-1})$
diverges as $\alpha_{0}^{-1}(\alpha_{0}-1)^{-1}$(\citep{bai2010sample}),
$r_{0}$ is also dominated by the contribution from the noise term
in $Y$, and is given by $\sigma_{0}^{2}\alpha_{0}^{-1}(\alpha_{0}-1)^{-1}$.
Similarly, we obtain $u_{0}\sim(\sigma^{2}\frac{N_{0}}{N}r_{0})^{1/L+1}\sim[\alpha_{0}(\alpha_{0}-1)]^{-1/L+1}$
and $\langle\langle f(x)\rangle^{2}\rangle\sim\gamma\sigma_{0}^{2}(\alpha_{0}-1)^{-1}$.
The generalization error is dominated by the divergent bias as $\alpha_{0}\rightarrow1$,
and diverges as $(\alpha_{0}-1)^{-1}$.

The case of clustered inputs, as in our template model, is treated
analytically in SM $\text{{\rm IID}}$.

\section{\label{sec:Multiple-Outputs}Multiple Outputs}

\textbf{BPKR for multiple outputs: }Here we extend the calculations
in Appendix $\ref{subsec:appendixa}$ to multiple outputs ($m>1$)
in the zero-temperature limit for $\alpha<1$. For $m>1$ we introduce
the integration variables $t$ form an $P\times m$ matrix, hence,
\begin{alignat}{1}
Z_{L-1} & =\int\Pi_{i=1}^{N_{L}}dw_{L}^{i}\int dt\exp[-\frac{1}{2}\mathrm{Tr}(t^{\top}K_{L}t)\nonumber \\
 & +i\mathrm{Tr}(t^{\top}Y)-\frac{1}{2\sigma^{2}}\mathrm{Tr}(W{}^{\top}W)]\label{eq:Z_l_mul}\\
 & =\int dt\exp[i\mathrm{Tr}(t^{\top}Y)+NG(t)-\frac{1}{2\sigma^{2}}\mathrm{Tr}(W'^{\top}W')]\nonumber \\
G(t) & =\log\langle\exp-\frac{1}{2N}\mathrm{Tr}(t^{\top}K_{w}t)\rangle_{w}\label{eq:Gt_mul}
\end{alignat}
Integrating over $w$ yields $G(t)=-\frac{1}{2}\log\det(I+\mathcal{H}_{L-1})$
where the $m\text{x}m$ dim matrix is $\mathcal{H}_{L-1}=\frac{\sigma^{2}}{N}t^{\top}K_{L-1}t$,
a relation which is enforced by an auxiliary matrix variable $\mathcal{U}_{L-1}$
. With $\hat{t}=K_{L-1}^{1/2}t$, we have 
\begin{multline}
Z_{L-1}=\int d\mathcal{U}_{L-1}\int d\mathcal{H}_{L-1}\int dt\exp[i\mathrm{Tr}(t^{\top}Y)\\
-\frac{N}{2}\log\det(I+\mathcal{H}_{L-1})+\frac{N}{2\sigma^{2}}\mathrm{Tr}(\mathcal{U}_{L-1}\mathcal{H}_{L-1})\\
-\frac{1}{2}\mathrm{Tr}(\mathcal{U}_{L-1}t^{\top}K_{L-1}t)-\frac{1}{2\sigma^{2}}\mathrm{Tr}(W'^{\top}W')]\\
=\int d\mathcal{U}_{L-1}\int d\mathcal{H}_{L-1}\int d\hat{t}\exp[i\hat{t}^{\top}K_{L-1}^{-1/2}Y\\
-\frac{N}{2}\log\det(I+\mathcal{H}_{L-1})+\frac{N}{2\sigma^{2}}\mathrm{Tr}(\mathcal{U}_{L-1}\mathcal{H}_{L-1})\\
-\frac{m}{2}\log\det K_{L-1}-\frac{1}{2}\mathrm{Tr}(\hat{t}\mathcal{U}_{L-1}\hat{t}^{\top})-\frac{1}{2\sigma^{2}}\mathrm{Tr}(W'^{\top}W')]
\end{multline}
For $\alpha<1$, we can integrate over $\hat{t}$, yielding 
\begin{multline}
Z_{L-1}=\int d\mathcal{U}_{L-1}\int d\mathcal{H}_{L-1}\exp[-\frac{N}{2}\log\det(I+\mathcal{H}_{L-1})\\
+\frac{N}{2\sigma^{2}}\mathrm{Tr}(\mathcal{U}_{L-1}\mathcal{H}_{L-1})-\frac{1}{2}\mathrm{Tr}(\mathcal{U}_{L-1}^{-1}Y^{\top}K_{L-1}^{-1}Y)\\
-\frac{m}{2}\log\det K_{L-1}-\frac{P}{2}\log\det(\mathcal{U}_{L-1})-\frac{1}{2\sigma^{2}}\mathrm{Tr}(W'^{\top}W')]
\end{multline}
Again substituting the saddle point of $\mathcal{H}_{L-1}$, i.e.,
$I+\mathcal{H}_{L-1}=\sigma^{2}\mathcal{U}_{L-1}^{-1},$yields 
\begin{align}
Z_{L-1} & =\int d\mathcal{U}_{L-1}\exp[\frac{N}{2}\log\det\mathcal{U}_{L-1}-\frac{N}{2\sigma^{2}}\mathrm{Tr}(\mathcal{U}_{L-1})\nonumber \\
 & -\frac{1}{2}\mathrm{Tr}(\mathcal{U}_{L-1}^{-1}Y^{\top}K_{L-1}^{-1}Y)-\frac{m}{2}\log\det K_{L-1}\nonumber \\
 & -\frac{P}{2}\log\det(\mathcal{U}_{L-1})-\frac{1}{2\sigma^{2}}\mathrm{Tr}(W'^{\top}W')]
\end{align}
Differentiating w.r.t. $\mathcal{U}_{L-1}$ we obtain the self-consistent
equation for $\mathcal{U}_{L-1}$,

\begin{equation}
I-\sigma^{-2}\mathcal{U}_{L-1}=\alpha(I-\frac{1}{P}Y^{\top}K_{L-1}^{-1}Y\mathcal{U}_{L-1}^{-1})
\end{equation}
A similar conclusion can be extended to the following integration
steps, and we have 
\begin{equation}
I-\sigma^{-2}\mathcal{U}_{L-l}=\alpha(I-\frac{1}{P}Y^{\top}K_{L-l}^{-1}Y\mathcal{U}_{L-l}^{-l})\label{eq:mathcalU}
\end{equation}
From these equations, it follows that for all $l$ , $\mathcal{U}_{L-l}$
can be diagonalized with the eigenvectors of the mean squared readout
matrix. Writing the eigenvalue matrix of the readout matrix as $\mathrm{diag}(r_{1,L-l},\cdots,r_{k,L-l},\cdots,r_{m,L-l})=V_{L-l}^{\top}(\frac{1}{P}Y^{\top}K_{L-l}^{-1}Y)V_{L-l},$
the renormalization eigenvalue matrix can be written as $\mathrm{diag}(u_{1,L-l},\cdots,u_{k,L-l},\cdots,u_{m,L-l})=V_{L-l}^{\top}\mathcal{U}_{L-l}V_{L-l}$
and Eq.\ref{eq:mathcalU} can be reduced to independent equations
for the eigenvalues ${u_{k,L-l}}_{1\leq k\leq m}$, as given by Eq.$\text{\ref{eq:layerUl}}$.
However, Eq.$\text{\ref{eq:mathcalU}}$ holds only for $\alpha<1$,
due to the singularity of $K_{L-l}$ for $l<L$ at $\alpha>1$, the
equation for the eigenvalues of $\mathcal{U}_{L-l}$ is replaced by
Eq.$\text{\ref{eq:u_L-l-2-2}}$ for narrow networks. (See SM IIIB
for details).

For wide networks, we can calculate the statistics of $Y^{\top}K_{l}^{-1}Y$
with a similar approach as that for the single-output case, by relating
it to the average of $t$ (see SM IIIA for details), obtaining Eq.$\text{\ref{eq:u vs r-1-1}}$.
For narrow networks, we calculate the statistics of $Y^{\top}K_{l}^{+}Y$,
for the same reason as in the single-output calculations, we need
to relate the quantity to second-order moment of $t$, the procedure
is also similar as for the single-output case in Appendix $\text{\ref{subsec:appendixa}}$,
and we obtain Eqs.$\text{\ref{eq:u vs r-1-1}},\text{\ref{eq:narrowop}}$
(see details in SM IIIC).

Iterating the integration steps until all weights are integrated,
we obtain the equation for the eigenvalues of $\mathcal{U}_{0}$
\begin{equation}
I-\sigma^{-2}u_{k0}=\alpha(1-u_{k0}^{-L}r_{k0})\label{eq:mathcalU0}
\end{equation}
where $\mathcal{U}_{0}=V_{0}\mathrm{diag}(u_{10},\cdots,u_{k0},\cdots,u_{m0})V_{0}^{\top}$
, $V_{0}$ is defined through the input readout covariance matrix,
$\frac{1}{P}Y^{\top}K_{0}^{-1}Y=V_{0}\mathrm{diag}(r_{10},\cdots,r_{k0},\cdots,r_{m0})V_{0}^{\top}$,
proving Eq. \ref{eq:multipleU0}.

Eq.$\text{\ref{eq:mathcalU0}}$ holds for all $\alpha$ as long as
$\alpha_{0}<1$. We also note that a straightforward generalization
of Eq.\ref{eq:iterateU-1}, leads to 
\begin{equation}
\langle f(\mathcal{U}_{l})\rangle_{l}=f(\mathcal{U}_{l-1})\label{eq:iterateU-2}
\end{equation}
which also holds for all $0\leq\alpha<\infty.$

\textbf{Generalization for multiple outputs:} The generalization error
and related quantities can be calculated similarly as in Appendix
$\text{\ref{subsec:Generalization}}$, replacing the scalar order
parameter with a matrix.

\begin{multline}
Z_{L-l}(t_{P+1})=\int d\mathcal{U}_{L-l}\int dt\\
\exp[i\mathrm{Tr}(t^{\top}Y)+\frac{(N-P)l}{2}\log\det\mathcal{U}_{L-l}\\
-\frac{Nl}{2\sigma^{2}}Tr(\mathcal{U}_{L-l})-\frac{1}{2}\mathrm{Tr}(\mathcal{U}_{L-l}^{l}t^{\top}K_{L-l}t)\\
-\frac{1}{2\sigma^{2}}\mathrm{Tr}(W'^{\top}W')+\mathrm{Tr}(\mathcal{U}_{L-l}^{l}t_{P+1}^{\top}k_{L-l}^{\top}t)\\
-\frac{1}{2}\mathrm{Tr}(\mathcal{U}_{L-l}^{l}t_{P+1}^{\top}K_{L-l}(x,x)t_{P+1})]
\end{multline}
Integrate over $t$, and take $l=L$.
\begin{multline}
Z(t_{p+1})=\int d\mathcal{U}_{0}\exp[\frac{(N-P)L}{2}\log\det\mathcal{U}_{0}\\
-\frac{NL}{2\sigma^{2}}Tr(\mathcal{U}_{0})-\frac{1}{2}\mathrm{Tr}(\mathcal{U}_{0}^{L}t_{P+1}^{\top}K_{0}(x,x)t_{P+1})\\
+\frac{1}{2}\mathrm{Tr}[(iK_{0}^{-1/2}Y+K_{0}^{-1/2}k_{0}(x)t_{P+1}\mathcal{U}_{0}^{L})\mathcal{U}_{0}^{-L}\\
(iK_{0}^{-1/2}Y+K_{0}^{-1/2}k_{0}(x)t_{P+1}\mathcal{U}_{0}^{L})^{\top}]
\end{multline}
Taking derivative w.r.t. $t_{P+1}$
\begin{multline}
\langle f(x)\rangle=\partial_{it_{p+1}}\log Z|_{t_{p+1}=0}\\
=\frac{\partial\mathrm{Tr}(it_{P+1}^{\top}k_{0}^{\top}(x)K_{0}^{-1}Y)}{\partial it_{P+1}}=k_{0}^{\top}(x)K_{0}^{-1}Y
\end{multline}
\begin{multline}
\langle\delta f_{i}(x)\delta f_{j}(x)\rangle=\partial_{it_{P+1}^{i}}\partial_{it_{P+1}^{j}}\log Z|_{t_{p+1}=0}\\
=\frac{\frac{1}{2}\partial\mathrm{Tr}(U_{0}^{L}t_{P+1}^{\top}t_{P+1})}{\partial_{t_{P+1}^{i}}\partial_{t_{P+1}^{j}}}[K_{0}(x,x)-k_{0}^{\top}(x)K_{0}^{-1}k_{0}(x)]\\
=\mathcal{U}_{0i,j}^{L}[K_{0}(x,x)-k_{0}^{\top}(x)K_{0}^{-1}k_{0}(x)]
\end{multline}
we obtain the predictor statistics as described in Section $\text{\ref{subsec:Multiple-Outputs}}$.

\textbf{Multiple outputs at finite temperature:} In Section $\text{\ref{subsec:Multiple-Outputs}}$
we focused on results for multiple outputs at zero temperature. Here
we introduce the results for multiple outputs at finite $T$ (see
SM IVB for detailed derivations). The partition function after integrating
over $l$ layers is given by 
\begin{multline}
Z_{L-l}=\int d\mathcal{U}_{L-l}\exp[\frac{Nl}{2}\log\det\mathcal{U}_{L-l}\\
-\frac{Nl}{2\sigma^{2}}\mathrm{Tr}(\mathcal{U}_{L-l})-\frac{1}{2}\hat{Y}^{\top}(\mathcal{U}_{L-l}^{l}\otimes K_{L-l}+TI)^{-1}\hat{Y}\\
-\frac{1}{2}\log\det(\mathcal{U}_{L-l}^{l}\otimes K_{L-l}+TI)-\frac{1}{2\sigma^{2}}\mathrm{Tr}(W'^{\top}W')]
\end{multline}
where $\hat{Y}$ is a $mP$ dimensional vector that denotes the vectorized
$Y$. The corresponding saddlepoint equation for $\mathcal{U}_{L-l}$
for $1\leq l\leq L$ is given as \begin{widetext}

\begin{multline}
I-\sigma^{-2}\mathcal{U}_{L-l}=\frac{1}{N}\mathrm{Tr}_{P}[(\mathcal{U}_{L-l}^{l}\otimes K_{L-l}+TI)^{-1}(\mathcal{U}_{L-l}^{l}\otimes K_{L-l})]\\
+\frac{1}{N}\mathrm{Tr}_{P}[(\mathcal{U}_{L-l}^{l}\otimes K_{L-l}+TI)^{-1}\hat{Y}\hat{Y}^{\top}(\mathcal{U}_{L-l}^{l}\otimes K_{L-l}+TI)^{-1}(\mathcal{U}_{L-l}^{l}\otimes K_{L-l})]
\end{multline}
\end{widetext}where the matrices on the RHS before taking the trace
are of dimension $mP\times mP$, and $\mathrm{Tr}_{P}$ denotes summing
the $P$ diagonal blocks of size $m\times m$. Unlike the zero-temperature
case, we cannot obtain saddle-point equations for each of the eigenvalues
of $\mathcal{U}_{L-l}$. The kernel undergoes kernel renormalization
in the form of a Kronecker product with the renormalization matrix
$\mathcal{U}_{l}$. 

The predictor statistics of the multiple output case at finite temperature
is given by 
\begin{multline}
\langle f(x_{P+1})\rangle=(\mathcal{U}_{0}^{L}\otimes k_{0}^{\top}(x))(\mathcal{U}_{0}^{L}\otimes K_{0}+TI)^{-1}\hat{Y}
\end{multline}
\begin{multline}
\langle\delta f(x_{P+1})\delta f(x_{P+1})^{\top}\rangle=\mathcal{U}_{0}^{L}K_{0}(x,x)\\
-(\mathcal{U}_{0}^{L}\otimes k_{0}^{\top}(x))(\mathcal{U}_{0}^{L}\otimes K_{0}+TI)^{-1}(\mathcal{U}_{0}^{L}\otimes k_{0}(x))
\end{multline}
See another formulation of the results without the Kronecker product
in SM IVB. 

\section{\label{subsec:Mean-layer-kernels}Weight Covariance and Mean Layer
Kernels}

We derive the mean layer kernels for multiple output network, starting
from calculating $\langle w_{L}w_{L}^{\top}\rangle_{L}$, where $w_{L}$
is the weight vector corresponding to \emph{a single node} in the
$L$-th hidden layer, \emph{conditioned on the weights of the previous
$L-1$ layers.} Using Eqs.$\text{\ref{eq:Z_l_mul}}$,$\text{\ref{eq:Gt_mul}}$,
this quantity can be expressed as 
\begin{multline}
\langle w_{L}w_{L}^{\top}\rangle_{L}=\frac{1}{Z_{L-1}}\int dtA(t)\\
\exp[i\mathrm{Tr}(t^{\top}Y)+NG(t)-\frac{1}{2\sigma^{2}}\mathrm{Tr}(W'^{\top}W')]
\end{multline}
where 
\begin{multline}
A(t)=\frac{\langle w_{L}w_{L}^{\top}\exp-\frac{1}{2N}\mathrm{Tr}(t^{\top}K_{w}^{L}t)\rangle_{w_{L}}}{\langle\exp-\frac{1}{2N}\mathrm{Tr}(t^{\top}K_{w}^{L}t)\rangle_{w_{L}}}\\
=\frac{1}{z}\int dw_{L}w_{L}w_{L}^{\top}\\
\exp[-\frac{1}{2\sigma^{2}}w_{L}^{\top}(I+\frac{\sigma^{4}}{NN_{L-1}}X_{L-1}tt^{\top}X_{L-1}^{\top})w_{L}]\\
=\sigma^{2}[I+\frac{\sigma^{4}}{NN_{L-1}}X_{L-1}tt^{\top}X_{L-1}^{\top}]^{-1}\\
=\sigma^{2}[I-\frac{\sigma^{4}}{NN_{L-1}}X_{L-1}t[I+\frac{\sigma^{2}}{N}t^{\top}K_{L-1}t]^{-1}t^{\top}X_{L-1}^{\top}]
\end{multline}
Plugging $A(t)$ back in, we have 
\begin{multline}
\langle w_{L}w_{L}^{T}\rangle{}_{L}=\sigma^{2}I-\frac{\sigma^{6}}{NN_{L-1}Z_{L-1}}\int dt\exp[i\mathrm{Tr}(t^{\top}Y)\\
-\frac{N}{2}\log\det(I+\frac{\sigma^{2}}{N}t^{\top}K_{L-1}t)-\frac{1}{2\sigma^{2}}\mathrm{Tr}(W'^{\top}W')]\\
\times X_{L-1}t[I+\frac{\sigma^{2}}{N}t^{\top}K_{L-1}t]^{-1}t^{\top}X_{L-1}^{\top}\label{eq:average-w}
\end{multline}
The term $\mathrm{Tr}(W'^{\top}W')$ does not depend on $t$, therefore
we ignore it for simplicity below.

We compute first the integral over $t$, by introducing OPs $\mathcal{U}_{L-1}$
and $\mathcal{H}_{L-1}$ as in Appendix $\text{\ref{sec:Multiple-Outputs}}$,
and with change of variable $\hat{t}=K_{L-1}^{1/2}t$, and the saddle-point
relation $\mathcal{U}_{L-1}(I+\mathcal{H}_{L-1})=\sigma^{2}I$, we
write the integration over $t$ as 
\begin{multline}
\int d\hat{t}\exp[i\mathrm{Tr}(\hat{t}^{\top}K_{L-1}^{-1/2}Y)-\frac{1}{2}\mathrm{Tr}(\hat{t}\mathcal{U}_{L-1}\hat{t}^{\top})]\\
\times K_{L-1}^{-1/2}\hat{t}[I+\mathcal{H}_{L-1}]^{-1}t^{\top}K_{L-1}^{-1/2}\\
=K_{L-1}^{-1/2}[\mathrm{Tr}(\mathcal{U}_{L-1}^{-1}(I+\mathcal{H}_{L-1})^{-1})I\\
-\sigma^{-2}(K_{L-1}^{-1/2}Y\mathcal{\mathcal{U}}_{L-1}^{-1})\mathcal{U}_{L-1}(K_{L-1}^{-1/2}Y\mathcal{U}_{L-1}^{-1})^{\top}]K_{L-1}^{-1/2}\\
=\sigma^{-2}(mK_{L-1}^{-1}-K_{L-1}^{-1}Y\mathcal{U}_{L-1}^{-1}Y^{\top}K_{L-1}^{-1})
\end{multline}
Plugging back in Eq.$\text{\ref{eq:average-w}}$ yields 
\begin{multline}
\langle w_{L}w_{L}^{\top}\rangle{}_{L}=\sigma^{2}I-\frac{\sigma^{4}}{NN_{L-1}}[mX_{L-1}K_{L-1}^{-1}X_{L-1}^{\top}\\
-X_{L-1}K_{L-1}^{-1}Y\mathcal{U}_{L-1}^{-1}Y^{\top}K_{L-1}^{-1}X_{L-1}^{\top}]\label{eq:wwT}
\end{multline}
In particular, the weight variance, $\langle w_{L}^{\top}w_{L}\rangle_{L}$
equals 
\begin{multline}
\langle w_{L}^{\top}w_{L}\rangle_{L}=\mathrm{Tr}\langle w_{L}w_{L}^{\top}\rangle\\
=\sigma^{2}N-\sigma^{2}\alpha(m-u_{L-1}^{-1}r_{L-1})
\end{multline}
implying that while the GP term is of $\mathcal{O}(N)$ as expected,
the non-GP correction term is of $\mathcal{O}(1)$.

Using similar methods, we can derive for all layers,

\begin{multline}
\langle w_{L-l}w_{L-l}^{\top}\rangle_{L-l}\\
=\sigma^{2}I-\frac{\sigma^{4}}{NN_{L-l-1}}[mX_{L-l-1}K_{L-l-1}^{-1}X_{L-l-1}^{\top}\\
-X_{L-l-1}K_{L-l-1}^{-1}Y\mathcal{U}_{L-l-1}^{-(l+1)}Y^{\top}K_{L-l-1}^{-1}X_{L-l-1}^{\top}]\label{eq:wwT-1}
\end{multline}
Since $\langle K_{L-l}\rangle=\frac{\sigma^{2}}{N_{L-l-1}}X_{L-l-1}^{\top}\langle w_{L-l}w_{L-l}^{\top}\rangle X_{L-l-1}$,
\begin{equation}
\langle K_{L-l}\rangle=\sigma^{2}[(1-\frac{m}{N})K_{L-l-1}+\frac{1}{N}Y\mathcal{U}_{L-l-1}^{-(l+1)}Y^{\top}]\label{eq:iterationeq}
\end{equation}
Using Eqs.$\text{\ref{eq:iterationeq}}$,\ref{eq:iterateU-2}, we
derive Eqs.$\text{\ref{eq:Kl-1 averaged}}$,$\text{\ref{eq:Kl-1 averagedm}}$
through iterations. The above result holds for all $\alpha$. For
derivation of this result in narrow networks, see SM ${\rm VA}$ .

\section{\label{sec:Details-of-Numerical}Details of Numerical Studies}

\textbf{1. Examples:}

\textbf{a) \label{a)-Template-model}Template model}

Instead of having the ‘standard’ input statistics in many synthetic
models, namely sampled i.i.d. from a normal distribution, we assume
a ‘template’ model, in which inputs are clustered and the target rule
largely obeys this structure, in order to introduce a strong correlation
between input structure and target outputs, as explained below. These
types of examples are common in practice, for instance in MNIST and
CIFAR-10, where the network is trained on clustered input data and
the target labels exhibit significant correlations with the cluster
structure (see Figs. $\text{\ref{fig:deepkernels}}$,$\text{\ref{fig:relu_mnist}}$,$\text{\ref{fig:doubledescent}}$(k-o)
for results on MNIST example). Because of these input-output correlations,
the template model can yield good generalization performance even
well below the interpolation threshold. {[}The predicted dependence
of the system's properties on the various parameters are general{]}.
In principle, both training and test data would be sampled from the
same cluster statistics. To simplify the analysis, instead we assume
that training inputs consist of \textbf{the cluster centers}, or what
we call \textbf{the templates}. The test inputs are samples by adding
Gaussian noise to thet training inputs, such that the test data are
clustered around the training data, as illustrated in Fig.$\text{\ref{fig:egvsN}}$(a).
\begin{equation}
x_{test}^{\mu}=\sqrt{1-\gamma}x^{\mu}+\sqrt{\gamma}\eta\label{eq:testdata}
\end{equation}
We consider two types of labeling of the data which differ in task
complexity. One is a noisy linear teacher task, where the labels are
generated by a noisy linear teacher network, $y=\frac{1}{\sqrt{N_{0}}}w_{0}^{\top}x+\sigma_{0}\eta$.
Here $w_{0}\sim\mathcal{N}(0,\sigma_{w}^{2}I)$, $\eta\sim\mathcal{N}(0,I)$
are both Gaussian i.i.d., $\sigma_{w}$ represents the amplitude of
the linear teacher weights, and $\sigma_{0}$ represents the noise
level of the linear teacher. Here the optimal weights of the linear
DNNs are those that yield a linear input-output mapping identical
to that given by the teacher weights. Because of the linearity of
the rule, if $\sigma_{0}$ is small the system can yield small training
error even when $P>N_{0}$. If $\gamma$ is also small, the system
will again yield a good generalization error approaching its minimum
on the RHS of the interpolation threshold ($\alpha_{0}>1$), when
the network is in the under-parameterized regime, converging to the
optimal error for $\alpha_{0}\gg1$ , see Fig.$\text{\ref{fig:egvsP}}$(a)-(d).
The second task is random labeling of the data clusters that are centered
around the templates. We assign random binary labels to the data $x^{\mu}$,
$Y^{\mu}\in\{-1,1\}$. For the test data, we assign label $Y^{\mu}$
to it if it is generated by adding noise to the training data $x^{\mu}$.
Here, for $P>N_{0}$ the task is inherently nonlinear (even for small
$\gamma$ ) and the minimum generalization error is achieved on the
LHS of the interpolation threshold ($\alpha_{0}<1$), because small
$P$ implies not only small size of training data but also an easier
task.

\textbf{Parameters for the noisy linear teacher: }

Fig.$\text{\ref{fig:egvsN}}$: The parameters are $N_{0}=1000$, $P=300$,
$\gamma=0.05$, $\sigma_{0}=0.1$, $\sigma_{w}=1$. For the top panels
(c-f), we are in the small noise regime where the generalization error
decreases with $N$, here we choose $\sigma=0.5$. For the bottom
panels (g-j), we are in the larger noise regime where the generalization
error increases with $N$, with $\sigma=1.3$.

Fig.$\text{\ref{fig:egvsL}}$: The parameters are $N_{0}=200$, $P=100$,
$\gamma=0.05$, $\sigma_{0}=0.1$. For the top panels (a-d), we are
in the sub-regime where the generalization error decreases with $L$,
here the parameters are $\sigma_{w}=0.3$, $\sigma=1.1$, $\alpha=0.5.$
For the middle panels (e-h), we are in the sub-regime where the generalization
error increases with $L$ but goes to a finite limit as $L\rightarrow\infty$,
here the parameters are $\sigma_{w}=0.9$, $\sigma=1.35$, $\alpha=0.5.$
For the bottom panels (i-l), we are in the high-noise regime where
the generalization error increases with $L$ and diverges as $L\rightarrow\infty$,
here the parameters are $\sigma_{w}=0.9$, $\sigma=1.35$, $\alpha=0.4$.

Fig.$\text{\ref{fig:egvsP}}$ (a-d): The parameters are $N_{0}=500$,
$N=200$, $\gamma=0.1$, $\sigma_{0}=0.3$, $\sigma_{w}=1$, $\sigma=1$.

Fig.$\text{\ref{fig:relu_linearteacher}}$: The parameters are $N_{0}=400$,
$P=100$, $\gamma=0.05$, $\sigma_{0}=0.1$, $\sigma_{w}=1.$ For
the top panels (a-d), we are in the small-noise regime where the generalization
error decreases with $N$, here $\sigma=1$. For the bottom panels
(e-h), we are in the high-noise regime where the generalization error
increases with $N$, and here $\sigma=2$.

Fig\@.$\text{\ref{fig:alpha0>1}}$: The parameters are $N_{0}=100$,
$P=200$, $\gamma=0.05$, $\sigma_{0}=0.1$, $\sigma_{w}=1.$ For
the top panels (a-d), we are in the small-noise regime where the generalization
error decreases with $N$, here $\sigma=1$. For the bottom panels
(e-h), we are in the high-noise regime where the generalization error
increases with $N$, and here $\sigma=1.3$.

Fig.$\text{\ref{fig:doubledescent}}$ (a-e): The simulation parameters
are $N_{0}=20$, $P=300$, $\gamma=0.05$,$\sigma_{0}=0.3,$$\sigma_{w}=1$,
$\sigma=1$.

Fig.$\text{\ref{fig:deeprelu}}$ (a-d): The simulation parameters
are $N_{0}=200$, $P=100$, $\gamma=0.05$, $\alpha=0.7$, $\sigma_{0}=0.1$,
$\sigma_{w}=0.3$,$\sigma=1$.

\textbf{Parameters for the random cluster labeling: }

Fig.$\text{\ref{fig:egvsP}}$ (e-h): The parameters are $N_{0}=500$,
$N=200$, $\gamma=0.1$, $\sigma=1$.

Fig.$\text{\ref{fig:egvsT}}$: (a) The parameters are $N_{0}=400$,
$N=800$, $\gamma=0.1,$$\sigma=0.5$. (d) The parameters are $N_{0}=400$,
$P=300$, $N=600$, $\gamma=0.1$, $\sigma=1.5$.

Fig.$\text{\ref{fig:doubledescent}}$ (e-h): The simulation parameters
are $N_{0}=20$, $P=300$, $\gamma=0.1$, $\sigma=1$.

\textbf{b)\label{b)-Synthetic-example} Synthetic example with block
structure}

In Fig.$\text{\ref{fig:examplesynthetic}}$, we present the layerwise
mean kernels trained on a synthetic example with an output similarity
matrix that exhibits a block structure. The parameters used in the
simulation are $N=N_{0}=100,P=80,\sigma_{0}=0.1$. We choose $\sigma=0.1,$
so that the non-GP correction term ($\sim\sigma^{2}/N$) is of the
same order as the input term ($\sim\sigma^{4}$) for the single hidden-layer
network we consider. In Fig.$\text{\ref{fig:examplesynthetic}}$(c,d)
we show the simulation and theoretical results for the non-GP correction
given by $\text{\ensuremath{\frac{\sigma^{2}}{N}}}YVU_{1}V^{T}Y^{T}$.

\textbf{c) \label{c)-Binary-classification}Binary classification
of randomly projected MNIST data}

The example we show in Fig.$\text{\ref{fig:deepkernels}}$ is trained
on a binary classification task on a subset of randomly projected
MNIST data. For MNIST the input dimension is fixed to be 784, the
number of pixels in the images. In the example we show, we first appropriately
normalize and center the data, such that it has zero mean and standard
deviation 1, then randomly project the MNIST data to $N_{0}$ dimensions
with a Gaussian i.i.d. weight matrix $W_{0}\in\mathbb{R}^{N_{0}\times784}$,
$W_{0}\sim\mathcal{N}(0,I)$ and add a ReLU nonlinearity to the projected
data. 
\begin{equation}
x=\mathrm{ReLU}(\frac{1}{\sqrt{784}}W_{0}x_{mnist})
\end{equation}
We then further train the network with the input $x^{\mu}(\mu=1,\cdots,P)$
and their corresponding labels.

In Fig.$\text{\ref{fig:deepkernels}}$ the network is trained on a
subset of the MNIST data with 4 different digits (1,5,6,7). The output
of the network is 6-dimensional ($y\in\{1,-1\}^{6}$) designed to
have hierarchical block structure, 4 of the binary outputs are ‘one
hot’ vectors each encoding one digit. The 4 digits are divided in
to 2 categories ((1,7) and (5,6)), the other two binary outputs each
classifies one of the two categories. The parameters are $N=N_{0}=1000$,
$P=100$. We again choose small $\sigma$($\sigma=0.1$) so that the
non-GP correction term becomes evident in the layerwise mean kernels.

In Fig.$\text{\ref{fig:doubledescent}}$ we use the same example to
train a ReLU network with a single output to perform binary classification
on 2 digits 0 and 1. The parameters are $N_{0}=20,P=300,\sigma=1$.

\textbf{d) \label{d)-Binary-classification}Binary classification
on subsets of the MNIST data}

We show in Fig.$\text{\ref{fig:relu_mnist}}$ the result for a ReLU
network trained on a binary classification task on subset of MNIST
data directly. The MNIST data determines the input dimension $N_{0}=784.$
We properly normalize and center the data, such that it has zero mean
and standard deviation 1. We train on a subset of MNIST data with
digits 0 and 1, the network outputs $y\in{1,-1}$. The parameters
are $N_{0}=784,P=100.$ For the top panels (a-d), we are in the small
noise regime where the generalization error decreases with $N$, here
$\sigma=0.8$. For the bottom panels (e-h), we are in the large noise
regime where the generalization error increases with $N$, and $\sigma=1.3$.

In Fig.$\text{\ref{fig:deeprelu}}$ (e-h) we use the same example
with parameters $N_{0}=784,P=100,\alpha=0.5,\sigma=1.3.$

\textbf{2. Langevin dynamics }

We run simulations to sample from the Gibbs distribution corresponding
to the energy $E$ given by Eq.$\text{\ref{eq:Loss-1}}$, defined
in Section $\text{\ref{sec:The-Back-propagating-KR}}$, and compute
the statistics from the distribution to compare with our theory. We
use the well-known result, that the Langevin dynamics

\begin{equation}
\Delta\Theta=-\epsilon\partial_{\Theta}E+\sqrt{2\epsilon T}\eta
\end{equation}
generates a time dependent distribution on the state space that converges
at long times to the Gibbs distribution. We perform the Langevin dynamics,
at each iteration we compute the predictor on a set of new points
and $r_{l}$ for the current weight, and obtain samples of the predictor
and $r_{l}$ from the underlying Gibbs distribution. We can then calculate
statistics of the predictor including $\langle f(x)\rangle$, $\langle\delta f(x)^{2}\rangle$,
and $\langle r_{l}\rangle/r_{0}$, and compare them with our theoretical
results. Because we focused mostly on $T\rightarrow0$, in our simulations
we also choose small $T$ ($T=0.001$ for results presented in all
figures except for Figs.$\ref{fig:examplesynthetic}$,$\text{\ref{fig:deepkernels}}$,
where due to the small $\sigma$ values, we needed to use $T=0.0001$).

\textbf{3. \label{3.-Finite-T}Finite T effects}

In the simulations of Langevin dynamics, we chose small $T$ in order
to compare with the $T\rightarrow0$ theoretical results. In most
cases presented in this paper, choosing $T=0.001$ (or in Section
$\text{\ref{subsec:Mean-Layer-Kernels}}$ with $T=0.0001$) is sufficient
to approximate the $T\rightarrow0$ limit. However, in Fig.$\text{\ref{fig:egvsP}}$
where we consider the dependence of the generalization error on $P$,
as $\alpha_{0}\rightarrow1$, the kernel $K_{0}$ becomes singular,
and the closer $\alpha_{0}$ is to $\alpha_{0}=1$ the lower the temperature
needs to be to approximate the zero $T$ limit. For this reason, in
Fig.$\text{\ref{fig:egvsP}}$, the solid curves show the theory for
finite $T$ with $T=0.001$ as in the simulation, with the mean predictor
and the variance $\langle\delta f(x)^{2}\rangle$ as given by Eqs.$\ref{eq:meanfiteration-1}$,$\text{\ref{eq:varfiteration-1}}$,
and the order parameter $u_{0}$ given by Eq.$\text{\ref{eq:u_L-l-1}}$
with $l=L$. In SM IIE we show the same results as in Fig.$\text{\ref{fig:egvsP}}$,
but now with an extra curve plotting the theoretical result for zero
temperature. We see that $T=0.001$ is a good approximation to the
zero $T$ theory when $\alpha_{0}$ is not close to the interpolation
threshold $\alpha_{0}=1$ but deviates from it as $\alpha_{0}$ approaches
this threshold.


\begin{thebibliography}{74}%
\makeatletter
\providecommand \@ifxundefined [1]{%
 \@ifx{#1\undefined}
}%
\providecommand \@ifnum [1]{%
 \ifnum #1\expandafter \@firstoftwo
 \else \expandafter \@secondoftwo
 \fi
}%
\providecommand \@ifx [1]{%
 \ifx #1\expandafter \@firstoftwo
 \else \expandafter \@secondoftwo
 \fi
}%
\providecommand \natexlab [1]{#1}%
\providecommand \enquote  [1]{``#1''}%
\providecommand \bibnamefont  [1]{#1}%
\providecommand \bibfnamefont [1]{#1}%
\providecommand \citenamefont [1]{#1}%
\providecommand \href@noop [0]{\@secondoftwo}%
\providecommand \href [0]{\begingroup \@sanitize@url \@href}%
\providecommand \@href[1]{\@@startlink{#1}\@@href}%
\providecommand \@@href[1]{\endgroup#1\@@endlink}%
\providecommand \@sanitize@url [0]{\catcode `\\12\catcode `\$12\catcode
  `\&12\catcode `\#12\catcode `\^12\catcode `\_12\catcode `\%12\relax}%
\providecommand \@@startlink[1]{}%
\providecommand \@@endlink[0]{}%
\providecommand \url  [0]{\begingroup\@sanitize@url \@url }%
\providecommand \@url [1]{\endgroup\@href {#1}{\urlprefix }}%
\providecommand \urlprefix  [0]{URL }%
\providecommand \Eprint [0]{\href }%
\providecommand \doibase [0]{https://doi.org/}%
\providecommand \selectlanguage [0]{\@gobble}%
\providecommand \bibinfo  [0]{\@secondoftwo}%
\providecommand \bibfield  [0]{\@secondoftwo}%
\providecommand \translation [1]{[#1]}%
\providecommand \BibitemOpen [0]{}%
\providecommand \bibitemStop [0]{}%
\providecommand \bibitemNoStop [0]{.\EOS\space}%
\providecommand \EOS [0]{\spacefactor3000\relax}%
\providecommand \BibitemShut  [1]{\csname bibitem#1\endcsname}%
\let\auto@bib@innerbib\@empty
\bibitem [{\citenamefont {Foerster}\ \emph {et~al.}(2016)\citenamefont
  {Foerster}, \citenamefont {Assael}, \citenamefont {De~Freitas},\ and\
  \citenamefont {Whiteson}}]{foerster2016learning}%
  \BibitemOpen
  \bibfield  {author} {\bibinfo {author} {\bibfnamefont {J.}~\bibnamefont
  {Foerster}}, \bibinfo {author} {\bibfnamefont {I.~A.}\ \bibnamefont
  {Assael}}, \bibinfo {author} {\bibfnamefont {N.}~\bibnamefont {De~Freitas}},\
  and\ \bibinfo {author} {\bibfnamefont {S.}~\bibnamefont {Whiteson}},\
  }\bibfield  {title} {\bibinfo {title} {Learning to communicate with deep
  multi-agent reinforcement learning},\ }in\ \href@noop {} {\emph {\bibinfo
  {booktitle} {Advances in neural information processing systems}}}\ (\bibinfo
  {year} {2016})\ pp.\ \bibinfo {pages} {2137--2145}\BibitemShut {NoStop}%
\bibitem [{\citenamefont {Goldberg}(2017)}]{goldberg2017neural}%
  \BibitemOpen
  \bibfield  {author} {\bibinfo {author} {\bibfnamefont {Y.}~\bibnamefont
  {Goldberg}},\ }\bibfield  {title} {\bibinfo {title} {Neural network methods
  for natural language processing},\ }\href@noop {} {\bibfield  {journal}
  {\bibinfo  {journal} {Synthesis Lectures on Human Language Technologies}\
  }\textbf {\bibinfo {volume} {10}},\ \bibinfo {pages} {1} (\bibinfo {year}
  {2017})}\BibitemShut {NoStop}%
\bibitem [{\citenamefont {LeCun}\ \emph {et~al.}(1999)\citenamefont {LeCun},
  \citenamefont {Haffner}, \citenamefont {Bottou},\ and\ \citenamefont
  {Bengio}}]{lecun1999object}%
  \BibitemOpen
  \bibfield  {author} {\bibinfo {author} {\bibfnamefont {Y.}~\bibnamefont
  {LeCun}}, \bibinfo {author} {\bibfnamefont {P.}~\bibnamefont {Haffner}},
  \bibinfo {author} {\bibfnamefont {L.}~\bibnamefont {Bottou}},\ and\ \bibinfo
  {author} {\bibfnamefont {Y.}~\bibnamefont {Bengio}},\ }\bibfield  {title}
  {\bibinfo {title} {Object recognition with gradient-based learning},\ }in\
  \href@noop {} {\emph {\bibinfo {booktitle} {Shape, contour and grouping in
  computer vision}}}\ (\bibinfo  {publisher} {Springer},\ \bibinfo {year}
  {1999})\ pp.\ \bibinfo {pages} {319--345}\BibitemShut {NoStop}%
\bibitem [{\citenamefont {Deng}\ \emph {et~al.}(2013)\citenamefont {Deng},
  \citenamefont {Hinton},\ and\ \citenamefont {Kingsbury}}]{deng2013new}%
  \BibitemOpen
  \bibfield  {author} {\bibinfo {author} {\bibfnamefont {L.}~\bibnamefont
  {Deng}}, \bibinfo {author} {\bibfnamefont {G.}~\bibnamefont {Hinton}},\ and\
  \bibinfo {author} {\bibfnamefont {B.}~\bibnamefont {Kingsbury}},\ }\bibfield
  {title} {\bibinfo {title} {New types of deep neural network learning for
  speech recognition and related applications: An overview},\ }in\ \href@noop
  {} {\emph {\bibinfo {booktitle} {2013 IEEE international conference on
  acoustics, speech and signal processing}}}\ (\bibinfo {organization} {IEEE},\
  \bibinfo {year} {2013})\ pp.\ \bibinfo {pages} {8599--8603}\BibitemShut
  {NoStop}%
\bibitem [{\citenamefont {Banino}\ \emph {et~al.}(2018)\citenamefont {Banino},
  \citenamefont {Barry}, \citenamefont {Uria}, \citenamefont {Blundell},
  \citenamefont {Lillicrap}, \citenamefont {Mirowski}, \citenamefont {Pritzel},
  \citenamefont {Chadwick}, \citenamefont {Degris}, \citenamefont {Modayil}
  \emph {et~al.}}]{banino2018vector}%
  \BibitemOpen
  \bibfield  {author} {\bibinfo {author} {\bibfnamefont {A.}~\bibnamefont
  {Banino}}, \bibinfo {author} {\bibfnamefont {C.}~\bibnamefont {Barry}},
  \bibinfo {author} {\bibfnamefont {B.}~\bibnamefont {Uria}}, \bibinfo {author}
  {\bibfnamefont {C.}~\bibnamefont {Blundell}}, \bibinfo {author}
  {\bibfnamefont {T.}~\bibnamefont {Lillicrap}}, \bibinfo {author}
  {\bibfnamefont {P.}~\bibnamefont {Mirowski}}, \bibinfo {author}
  {\bibfnamefont {A.}~\bibnamefont {Pritzel}}, \bibinfo {author} {\bibfnamefont
  {M.~J.}\ \bibnamefont {Chadwick}}, \bibinfo {author} {\bibfnamefont
  {T.}~\bibnamefont {Degris}}, \bibinfo {author} {\bibfnamefont
  {J.}~\bibnamefont {Modayil}}, \emph {et~al.},\ }\bibfield  {title} {\bibinfo
  {title} {Vector-based navigation using grid-like representations in
  artificial agents},\ }\href@noop {} {\bibfield  {journal} {\bibinfo
  {journal} {Nature}\ }\textbf {\bibinfo {volume} {557}},\ \bibinfo {pages}
  {429} (\bibinfo {year} {2018})}\BibitemShut {NoStop}%
\bibitem [{\citenamefont {Guo}\ \emph {et~al.}(2016)\citenamefont {Guo},
  \citenamefont {Liu}, \citenamefont {Oerlemans}, \citenamefont {Lao},
  \citenamefont {Wu},\ and\ \citenamefont {Lew}}]{guo2016deep}%
  \BibitemOpen
  \bibfield  {author} {\bibinfo {author} {\bibfnamefont {Y.}~\bibnamefont
  {Guo}}, \bibinfo {author} {\bibfnamefont {Y.}~\bibnamefont {Liu}}, \bibinfo
  {author} {\bibfnamefont {A.}~\bibnamefont {Oerlemans}}, \bibinfo {author}
  {\bibfnamefont {S.}~\bibnamefont {Lao}}, \bibinfo {author} {\bibfnamefont
  {S.}~\bibnamefont {Wu}},\ and\ \bibinfo {author} {\bibfnamefont {M.~S.}\
  \bibnamefont {Lew}},\ }\bibfield  {title} {\bibinfo {title} {Deep learning
  for visual understanding: A review},\ }\href@noop {} {\bibfield  {journal}
  {\bibinfo  {journal} {Neurocomputing}\ }\textbf {\bibinfo {volume} {187}},\
  \bibinfo {pages} {27} (\bibinfo {year} {2016})}\BibitemShut {NoStop}%
\bibitem [{\citenamefont {Poggio}\ \emph {et~al.}(2020)\citenamefont {Poggio},
  \citenamefont {Banburski},\ and\ \citenamefont
  {Liao}}]{poggio2020theoretical}%
  \BibitemOpen
  \bibfield  {author} {\bibinfo {author} {\bibfnamefont {T.}~\bibnamefont
  {Poggio}}, \bibinfo {author} {\bibfnamefont {A.}~\bibnamefont {Banburski}},\
  and\ \bibinfo {author} {\bibfnamefont {Q.}~\bibnamefont {Liao}},\ }\bibfield
  {title} {\bibinfo {title} {Theoretical issues in deep networks},\ }\href@noop
  {} {\bibfield  {journal} {\bibinfo  {journal} {Proceedings of the National
  Academy of Sciences}\ } (\bibinfo {year} {2020})}\BibitemShut {NoStop}%
\bibitem [{\citenamefont {Zhang}\ \emph {et~al.}(2021)\citenamefont {Zhang},
  \citenamefont {Bengio}, \citenamefont {Hardt}, \citenamefont {Recht},\ and\
  \citenamefont {Vinyals}}]{zhang2021understanding}%
  \BibitemOpen
  \bibfield  {author} {\bibinfo {author} {\bibfnamefont {C.}~\bibnamefont
  {Zhang}}, \bibinfo {author} {\bibfnamefont {S.}~\bibnamefont {Bengio}},
  \bibinfo {author} {\bibfnamefont {M.}~\bibnamefont {Hardt}}, \bibinfo
  {author} {\bibfnamefont {B.}~\bibnamefont {Recht}},\ and\ \bibinfo {author}
  {\bibfnamefont {O.}~\bibnamefont {Vinyals}},\ }\bibfield  {title} {\bibinfo
  {title} {Understanding deep learning (still) requires rethinking
  generalization},\ }\href@noop {} {\bibfield  {journal} {\bibinfo  {journal}
  {Communications of the ACM}\ }\textbf {\bibinfo {volume} {64}},\ \bibinfo
  {pages} {107} (\bibinfo {year} {2021})}\BibitemShut {NoStop}%
\bibitem [{\citenamefont {Baity-Jesi}\ \emph {et~al.}(2018)\citenamefont
  {Baity-Jesi}, \citenamefont {Sagun}, \citenamefont {Geiger}, \citenamefont
  {Spigler}, \citenamefont {Arous}, \citenamefont {Cammarota}, \citenamefont
  {LeCun}, \citenamefont {Wyart},\ and\ \citenamefont
  {Biroli}}]{baity2018comparing}%
  \BibitemOpen
  \bibfield  {author} {\bibinfo {author} {\bibfnamefont {M.}~\bibnamefont
  {Baity-Jesi}}, \bibinfo {author} {\bibfnamefont {L.}~\bibnamefont {Sagun}},
  \bibinfo {author} {\bibfnamefont {M.}~\bibnamefont {Geiger}}, \bibinfo
  {author} {\bibfnamefont {S.}~\bibnamefont {Spigler}}, \bibinfo {author}
  {\bibfnamefont {G.~B.}\ \bibnamefont {Arous}}, \bibinfo {author}
  {\bibfnamefont {C.}~\bibnamefont {Cammarota}}, \bibinfo {author}
  {\bibfnamefont {Y.}~\bibnamefont {LeCun}}, \bibinfo {author} {\bibfnamefont
  {M.}~\bibnamefont {Wyart}},\ and\ \bibinfo {author} {\bibfnamefont
  {G.}~\bibnamefont {Biroli}},\ }\bibfield  {title} {\bibinfo {title}
  {Comparing dynamics: Deep neural networks versus glassy systems},\ }in\
  \href@noop {} {\emph {\bibinfo {booktitle} {International Conference on
  Machine Learning}}}\ (\bibinfo {year} {2018})\ pp.\ \bibinfo {pages}
  {314--323}\BibitemShut {NoStop}%
\bibitem [{\citenamefont {Ballard}\ \emph {et~al.}(2017)\citenamefont
  {Ballard}, \citenamefont {Das}, \citenamefont {Martiniani}, \citenamefont
  {Mehta}, \citenamefont {Sagun}, \citenamefont {Stevenson},\ and\
  \citenamefont {Wales}}]{ballard2017energy}%
  \BibitemOpen
  \bibfield  {author} {\bibinfo {author} {\bibfnamefont {A.~J.}\ \bibnamefont
  {Ballard}}, \bibinfo {author} {\bibfnamefont {R.}~\bibnamefont {Das}},
  \bibinfo {author} {\bibfnamefont {S.}~\bibnamefont {Martiniani}}, \bibinfo
  {author} {\bibfnamefont {D.}~\bibnamefont {Mehta}}, \bibinfo {author}
  {\bibfnamefont {L.}~\bibnamefont {Sagun}}, \bibinfo {author} {\bibfnamefont
  {J.~D.}\ \bibnamefont {Stevenson}},\ and\ \bibinfo {author} {\bibfnamefont
  {D.~J.}\ \bibnamefont {Wales}},\ }\bibfield  {title} {\bibinfo {title}
  {Energy landscapes for machine learning},\ }\href@noop {} {\bibfield
  {journal} {\bibinfo  {journal} {Physical Chemistry Chemical Physics}\
  }\textbf {\bibinfo {volume} {19}},\ \bibinfo {pages} {12585} (\bibinfo {year}
  {2017})}\BibitemShut {NoStop}%
\bibitem [{\citenamefont {Becker}\ \emph {et~al.}(2020)\citenamefont {Becker},
  \citenamefont {Zhang} \emph {et~al.}}]{becker2020geometry}%
  \BibitemOpen
  \bibfield  {author} {\bibinfo {author} {\bibfnamefont {S.}~\bibnamefont
  {Becker}}, \bibinfo {author} {\bibfnamefont {Y.}~\bibnamefont {Zhang}}, \emph
  {et~al.},\ }\bibfield  {title} {\bibinfo {title} {Geometry of energy
  landscapes and the optimizability of deep neural networks},\ }\href@noop {}
  {\bibfield  {journal} {\bibinfo  {journal} {Physical Review Letters}\
  }\textbf {\bibinfo {volume} {124}},\ \bibinfo {pages} {108301} (\bibinfo
  {year} {2020})}\BibitemShut {NoStop}%
\bibitem [{\citenamefont {Rifai}\ \emph {et~al.}(2011)\citenamefont {Rifai},
  \citenamefont {Vincent}, \citenamefont {Muller}, \citenamefont {Glorot},\
  and\ \citenamefont {Bengio}}]{rifai2011contractive}%
  \BibitemOpen
  \bibfield  {author} {\bibinfo {author} {\bibfnamefont {S.}~\bibnamefont
  {Rifai}}, \bibinfo {author} {\bibfnamefont {P.}~\bibnamefont {Vincent}},
  \bibinfo {author} {\bibfnamefont {X.}~\bibnamefont {Muller}}, \bibinfo
  {author} {\bibfnamefont {X.}~\bibnamefont {Glorot}},\ and\ \bibinfo {author}
  {\bibfnamefont {Y.}~\bibnamefont {Bengio}},\ }\bibfield  {title} {\bibinfo
  {title} {Contractive auto-encoders: Explicit invariance during feature
  extraction},\ }in\ \href@noop {} {\emph {\bibinfo {booktitle} {Icml}}}\
  (\bibinfo {year} {2011})\BibitemShut {NoStop}%
\bibitem [{\citenamefont {Carleo}\ \emph {et~al.}(2019)\citenamefont {Carleo},
  \citenamefont {Cirac}, \citenamefont {Cranmer}, \citenamefont {Daudet},
  \citenamefont {Schuld}, \citenamefont {Tishby}, \citenamefont
  {Vogt-Maranto},\ and\ \citenamefont {Zdeborov{\'a}}}]{carleo2019machine}%
  \BibitemOpen
  \bibfield  {author} {\bibinfo {author} {\bibfnamefont {G.}~\bibnamefont
  {Carleo}}, \bibinfo {author} {\bibfnamefont {I.}~\bibnamefont {Cirac}},
  \bibinfo {author} {\bibfnamefont {K.}~\bibnamefont {Cranmer}}, \bibinfo
  {author} {\bibfnamefont {L.}~\bibnamefont {Daudet}}, \bibinfo {author}
  {\bibfnamefont {M.}~\bibnamefont {Schuld}}, \bibinfo {author} {\bibfnamefont
  {N.}~\bibnamefont {Tishby}}, \bibinfo {author} {\bibfnamefont
  {L.}~\bibnamefont {Vogt-Maranto}},\ and\ \bibinfo {author} {\bibfnamefont
  {L.}~\bibnamefont {Zdeborov{\'a}}},\ }\bibfield  {title} {\bibinfo {title}
  {Machine learning and the physical sciences},\ }\href@noop {} {\bibfield
  {journal} {\bibinfo  {journal} {Reviews of Modern Physics}\ }\textbf
  {\bibinfo {volume} {91}},\ \bibinfo {pages} {045002} (\bibinfo {year}
  {2019})}\BibitemShut {NoStop}%
\bibitem [{\citenamefont {Engel}\ and\ \citenamefont {Van~den
  Broeck}(2001)}]{engel2001statistical}%
  \BibitemOpen
  \bibfield  {author} {\bibinfo {author} {\bibfnamefont {A.}~\bibnamefont
  {Engel}}\ and\ \bibinfo {author} {\bibfnamefont {C.}~\bibnamefont {Van~den
  Broeck}},\ }\href@noop {} {\emph {\bibinfo {title} {Statistical mechanics of
  learning}}}\ (\bibinfo  {publisher} {Cambridge University Press},\ \bibinfo
  {year} {2001})\BibitemShut {NoStop}%
\bibitem [{\citenamefont {Mezard}\ and\ \citenamefont
  {Montanari}(2009)}]{mezard2009information}%
  \BibitemOpen
  \bibfield  {author} {\bibinfo {author} {\bibfnamefont {M.}~\bibnamefont
  {Mezard}}\ and\ \bibinfo {author} {\bibfnamefont {A.}~\bibnamefont
  {Montanari}},\ }\href@noop {} {\emph {\bibinfo {title} {Information, physics,
  and computation}}}\ (\bibinfo  {publisher} {Oxford University Press},\
  \bibinfo {year} {2009})\BibitemShut {NoStop}%
\bibitem [{\citenamefont {Advani}\ and\ \citenamefont
  {Saxe}(2017)}]{advani2017high}%
  \BibitemOpen
  \bibfield  {author} {\bibinfo {author} {\bibfnamefont {M.~S.}\ \bibnamefont
  {Advani}}\ and\ \bibinfo {author} {\bibfnamefont {A.~M.}\ \bibnamefont
  {Saxe}},\ }\bibfield  {title} {\bibinfo {title} {High-dimensional dynamics of
  generalization error in neural networks},\ }\href@noop {} {\bibfield
  {journal} {\bibinfo  {journal} {arXiv preprint arXiv:1710.03667}\ } (\bibinfo
  {year} {2017})}\BibitemShut {NoStop}%
\bibitem [{\citenamefont {Yuan}\ \emph {et~al.}(2012)\citenamefont {Yuan},
  \citenamefont {Ho},\ and\ \citenamefont {Lin}}]{yuan2012recent}%
  \BibitemOpen
  \bibfield  {author} {\bibinfo {author} {\bibfnamefont {G.-X.}\ \bibnamefont
  {Yuan}}, \bibinfo {author} {\bibfnamefont {C.-H.}\ \bibnamefont {Ho}},\ and\
  \bibinfo {author} {\bibfnamefont {C.-J.}\ \bibnamefont {Lin}},\ }\bibfield
  {title} {\bibinfo {title} {Recent advances of large-scale linear
  classification},\ }\href@noop {} {\bibfield  {journal} {\bibinfo  {journal}
  {Proceedings of the IEEE}\ }\textbf {\bibinfo {volume} {100}},\ \bibinfo
  {pages} {2584} (\bibinfo {year} {2012})}\BibitemShut {NoStop}%
\bibitem [{\citenamefont {Saxe}\ \emph {et~al.}(2019)\citenamefont {Saxe},
  \citenamefont {McClelland},\ and\ \citenamefont
  {Ganguli}}]{saxe2019mathematical}%
  \BibitemOpen
  \bibfield  {author} {\bibinfo {author} {\bibfnamefont {A.~M.}\ \bibnamefont
  {Saxe}}, \bibinfo {author} {\bibfnamefont {J.~L.}\ \bibnamefont
  {McClelland}},\ and\ \bibinfo {author} {\bibfnamefont {S.}~\bibnamefont
  {Ganguli}},\ }\bibfield  {title} {\bibinfo {title} {A mathematical theory of
  semantic development in deep neural networks},\ }\href@noop {} {\bibfield
  {journal} {\bibinfo  {journal} {Proceedings of the National Academy of
  Sciences}\ }\textbf {\bibinfo {volume} {116}},\ \bibinfo {pages} {11537}
  (\bibinfo {year} {2019})}\BibitemShut {NoStop}%
\bibitem [{\citenamefont {Saxe}\ \emph {et~al.}(2014)\citenamefont {Saxe},
  \citenamefont {Mcclelland},\ and\ \citenamefont {Ganguli}}]{saxe2014exact}%
  \BibitemOpen
  \bibfield  {author} {\bibinfo {author} {\bibfnamefont {A.~M.}\ \bibnamefont
  {Saxe}}, \bibinfo {author} {\bibfnamefont {J.~L.}\ \bibnamefont
  {Mcclelland}},\ and\ \bibinfo {author} {\bibfnamefont {S.}~\bibnamefont
  {Ganguli}},\ }\bibfield  {title} {\bibinfo {title} {Exact solutions to the
  nonlinear dynamics of learning in deep linear neural network},\ }in\
  \href@noop {} {\emph {\bibinfo {booktitle} {In International Conference on
  Learning Representations}}}\ (\bibinfo {organization} {Citeseer},\ \bibinfo
  {year} {2014})\BibitemShut {NoStop}%
\bibitem [{\citenamefont {Tishby}\ \emph {et~al.}(1989)\citenamefont {Tishby},
  \citenamefont {Levin},\ and\ \citenamefont {Solla}}]{tishby1989consistent}%
  \BibitemOpen
  \bibfield  {author} {\bibinfo {author} {\bibfnamefont {N.}~\bibnamefont
  {Tishby}}, \bibinfo {author} {\bibfnamefont {E.}~\bibnamefont {Levin}},\ and\
  \bibinfo {author} {\bibfnamefont {S.~A.}\ \bibnamefont {Solla}},\ }\bibfield
  {title} {\bibinfo {title} {Consistent inference of probabilities in layered
  networks: Predictions and generalization},\ }in\ \href@noop {} {\emph
  {\bibinfo {booktitle} {International Joint Conference on Neural Networks}}},\
  Vol.~\bibinfo {volume} {2}\ (\bibinfo {year} {1989})\ pp.\ \bibinfo {pages}
  {403--409}\BibitemShut {NoStop}%
\bibitem [{\citenamefont {MacKay}(1992)}]{mackay1992practical}%
  \BibitemOpen
  \bibfield  {author} {\bibinfo {author} {\bibfnamefont {D.~J.}\ \bibnamefont
  {MacKay}},\ }\bibfield  {title} {\bibinfo {title} {A practical bayesian
  framework for backpropagation networks},\ }\href@noop {} {\bibfield
  {journal} {\bibinfo  {journal} {Neural computation}\ }\textbf {\bibinfo
  {volume} {4}},\ \bibinfo {pages} {448} (\bibinfo {year} {1992})}\BibitemShut
  {NoStop}%
\bibitem [{\citenamefont {Neal}(2012)}]{neal2012bayesian}%
  \BibitemOpen
  \bibfield  {author} {\bibinfo {author} {\bibfnamefont {R.~M.}\ \bibnamefont
  {Neal}},\ }\href@noop {} {\emph {\bibinfo {title} {Bayesian learning for
  neural networks}}},\ Vol.\ \bibinfo {volume} {118}\ (\bibinfo  {publisher}
  {Springer Science \& Business Media},\ \bibinfo {year} {2012})\BibitemShut
  {NoStop}%
\bibitem [{\citenamefont {Bahri}\ \emph {et~al.}(2020)\citenamefont {Bahri},
  \citenamefont {Kadmon}, \citenamefont {Pennington}, \citenamefont
  {Schoenholz}, \citenamefont {Sohl-Dickstein},\ and\ \citenamefont
  {Ganguli}}]{bahri2020statistical}%
  \BibitemOpen
  \bibfield  {author} {\bibinfo {author} {\bibfnamefont {Y.}~\bibnamefont
  {Bahri}}, \bibinfo {author} {\bibfnamefont {J.}~\bibnamefont {Kadmon}},
  \bibinfo {author} {\bibfnamefont {J.}~\bibnamefont {Pennington}}, \bibinfo
  {author} {\bibfnamefont {S.~S.}\ \bibnamefont {Schoenholz}}, \bibinfo
  {author} {\bibfnamefont {J.}~\bibnamefont {Sohl-Dickstein}},\ and\ \bibinfo
  {author} {\bibfnamefont {S.}~\bibnamefont {Ganguli}},\ }\bibfield  {title}
  {\bibinfo {title} {Statistical mechanics of deep learning},\ }\href@noop {}
  {\bibfield  {journal} {\bibinfo  {journal} {Annual Review of Condensed Matter
  Physics}\ } (\bibinfo {year} {2020})}\BibitemShut {NoStop}%
\bibitem [{\citenamefont {Amit}\ \emph {et~al.}(1987)\citenamefont {Amit},
  \citenamefont {Gutfreund},\ and\ \citenamefont
  {Sompolinsky}}]{amit1987statistical}%
  \BibitemOpen
  \bibfield  {author} {\bibinfo {author} {\bibfnamefont {D.~J.}\ \bibnamefont
  {Amit}}, \bibinfo {author} {\bibfnamefont {H.}~\bibnamefont {Gutfreund}},\
  and\ \bibinfo {author} {\bibfnamefont {H.}~\bibnamefont {Sompolinsky}},\
  }\bibfield  {title} {\bibinfo {title} {Statistical mechanics of neural
  networks near saturation},\ }\href@noop {} {\bibfield  {journal} {\bibinfo
  {journal} {Annals of physics}\ }\textbf {\bibinfo {volume} {173}},\ \bibinfo
  {pages} {30} (\bibinfo {year} {1987})}\BibitemShut {NoStop}%
\bibitem [{\citenamefont {Advani}\ \emph {et~al.}(2013)\citenamefont {Advani},
  \citenamefont {Lahiri},\ and\ \citenamefont
  {Ganguli}}]{advani2013statistical}%
  \BibitemOpen
  \bibfield  {author} {\bibinfo {author} {\bibfnamefont {M.}~\bibnamefont
  {Advani}}, \bibinfo {author} {\bibfnamefont {S.}~\bibnamefont {Lahiri}},\
  and\ \bibinfo {author} {\bibfnamefont {S.}~\bibnamefont {Ganguli}},\
  }\bibfield  {title} {\bibinfo {title} {Statistical mechanics of complex
  neural systems and high dimensional data},\ }\href@noop {} {\bibfield
  {journal} {\bibinfo  {journal} {Journal of Statistical Mechanics: Theory and
  Experiment}\ }\textbf {\bibinfo {volume} {2013}},\ \bibinfo {pages} {P03014}
  (\bibinfo {year} {2013})}\BibitemShut {NoStop}%
\bibitem [{\citenamefont {Seung}\ \emph {et~al.}(1992)\citenamefont {Seung},
  \citenamefont {Sompolinsky},\ and\ \citenamefont
  {Tishby}}]{seung1992statistical}%
  \BibitemOpen
  \bibfield  {author} {\bibinfo {author} {\bibfnamefont {H.~S.}\ \bibnamefont
  {Seung}}, \bibinfo {author} {\bibfnamefont {H.}~\bibnamefont {Sompolinsky}},\
  and\ \bibinfo {author} {\bibfnamefont {N.}~\bibnamefont {Tishby}},\
  }\bibfield  {title} {\bibinfo {title} {Statistical mechanics of learning from
  examples},\ }\href@noop {} {\bibfield  {journal} {\bibinfo  {journal}
  {Physical review A}\ }\textbf {\bibinfo {volume} {45}},\ \bibinfo {pages}
  {6056} (\bibinfo {year} {1992})}\BibitemShut {NoStop}%
\bibitem [{\citenamefont {Watkin}\ \emph {et~al.}(1993)\citenamefont {Watkin},
  \citenamefont {Rau},\ and\ \citenamefont {Biehl}}]{watkin1993statistical}%
  \BibitemOpen
  \bibfield  {author} {\bibinfo {author} {\bibfnamefont {T.~L.}\ \bibnamefont
  {Watkin}}, \bibinfo {author} {\bibfnamefont {A.}~\bibnamefont {Rau}},\ and\
  \bibinfo {author} {\bibfnamefont {M.}~\bibnamefont {Biehl}},\ }\bibfield
  {title} {\bibinfo {title} {The statistical mechanics of learning a rule},\
  }\href@noop {} {\bibfield  {journal} {\bibinfo  {journal} {Reviews of Modern
  Physics}\ }\textbf {\bibinfo {volume} {65}},\ \bibinfo {pages} {499}
  (\bibinfo {year} {1993})}\BibitemShut {NoStop}%
\bibitem [{\citenamefont {Lee}\ \emph {et~al.}(2017)\citenamefont {Lee},
  \citenamefont {Bahri}, \citenamefont {Novak}, \citenamefont {Schoenholz},
  \citenamefont {Pennington},\ and\ \citenamefont
  {Sohl-Dickstein}}]{lee2017deep}%
  \BibitemOpen
  \bibfield  {author} {\bibinfo {author} {\bibfnamefont {J.}~\bibnamefont
  {Lee}}, \bibinfo {author} {\bibfnamefont {Y.}~\bibnamefont {Bahri}}, \bibinfo
  {author} {\bibfnamefont {R.}~\bibnamefont {Novak}}, \bibinfo {author}
  {\bibfnamefont {S.~S.}\ \bibnamefont {Schoenholz}}, \bibinfo {author}
  {\bibfnamefont {J.}~\bibnamefont {Pennington}},\ and\ \bibinfo {author}
  {\bibfnamefont {J.}~\bibnamefont {Sohl-Dickstein}},\ }\bibfield  {title}
  {\bibinfo {title} {Deep neural networks as gaussian processes},\ }\href@noop
  {} {\bibfield  {journal} {\bibinfo  {journal} {arXiv preprint
  arXiv:1711.00165}\ } (\bibinfo {year} {2017})}\BibitemShut {NoStop}%
\bibitem [{\citenamefont {Cho}\ and\ \citenamefont
  {Saul}(2009)}]{cho2009kernel}%
  \BibitemOpen
  \bibfield  {author} {\bibinfo {author} {\bibfnamefont {Y.}~\bibnamefont
  {Cho}}\ and\ \bibinfo {author} {\bibfnamefont {L.~K.}\ \bibnamefont {Saul}},\
  }\bibfield  {title} {\bibinfo {title} {Kernel methods for deep learning},\
  }in\ \href@noop {} {\emph {\bibinfo {booktitle} {Advances in neural
  information processing systems}}}\ (\bibinfo {year} {2009})\ pp.\ \bibinfo
  {pages} {342--350}\BibitemShut {NoStop}%
\bibitem [{\citenamefont {Chung}\ \emph {et~al.}(2018)\citenamefont {Chung},
  \citenamefont {Lee},\ and\ \citenamefont
  {Sompolinsky}}]{chung2018classification}%
  \BibitemOpen
  \bibfield  {author} {\bibinfo {author} {\bibfnamefont {S.}~\bibnamefont
  {Chung}}, \bibinfo {author} {\bibfnamefont {D.~D.}\ \bibnamefont {Lee}},\
  and\ \bibinfo {author} {\bibfnamefont {H.}~\bibnamefont {Sompolinsky}},\
  }\bibfield  {title} {\bibinfo {title} {Classification and geometry of general
  perceptual manifolds},\ }\href@noop {} {\bibfield  {journal} {\bibinfo
  {journal} {Physical Review X}\ }\textbf {\bibinfo {volume} {8}},\ \bibinfo
  {pages} {031003} (\bibinfo {year} {2018})}\BibitemShut {NoStop}%
\bibitem [{\citenamefont {Ganguli}\ and\ \citenamefont
  {Sompolinsky}(2010)}]{ganguli2010statistical}%
  \BibitemOpen
  \bibfield  {author} {\bibinfo {author} {\bibfnamefont {S.}~\bibnamefont
  {Ganguli}}\ and\ \bibinfo {author} {\bibfnamefont {H.}~\bibnamefont
  {Sompolinsky}},\ }\bibfield  {title} {\bibinfo {title} {Statistical mechanics
  of compressed sensing},\ }\href@noop {} {\bibfield  {journal} {\bibinfo
  {journal} {Physical review letters}\ }\textbf {\bibinfo {volume} {104}},\
  \bibinfo {pages} {188701} (\bibinfo {year} {2010})}\BibitemShut {NoStop}%
\bibitem [{\citenamefont {Ganguli}\ and\ \citenamefont
  {Sompolinsky}(2012)}]{ganguli2012compressed}%
  \BibitemOpen
  \bibfield  {author} {\bibinfo {author} {\bibfnamefont {S.}~\bibnamefont
  {Ganguli}}\ and\ \bibinfo {author} {\bibfnamefont {H.}~\bibnamefont
  {Sompolinsky}},\ }\bibfield  {title} {\bibinfo {title} {Compressed sensing,
  sparsity, and dimensionality in neuronal information processing and data
  analysis},\ }\href@noop {} {\bibfield  {journal} {\bibinfo  {journal} {Annual
  review of neuroscience}\ }\textbf {\bibinfo {volume} {35}},\ \bibinfo {pages}
  {485} (\bibinfo {year} {2012})}\BibitemShut {NoStop}%
\bibitem [{\citenamefont {Babadi}\ and\ \citenamefont
  {Sompolinsky}(2014)}]{babadi2014sparseness}%
  \BibitemOpen
  \bibfield  {author} {\bibinfo {author} {\bibfnamefont {B.}~\bibnamefont
  {Babadi}}\ and\ \bibinfo {author} {\bibfnamefont {H.}~\bibnamefont
  {Sompolinsky}},\ }\bibfield  {title} {\bibinfo {title} {Sparseness and
  expansion in sensory representations},\ }\href@noop {} {\bibfield  {journal}
  {\bibinfo  {journal} {Neuron}\ }\textbf {\bibinfo {volume} {83}},\ \bibinfo
  {pages} {1213} (\bibinfo {year} {2014})}\BibitemShut {NoStop}%
\bibitem [{\citenamefont {Mei}\ and\ \citenamefont
  {Montanari}(2019)}]{mei2019generalization}%
  \BibitemOpen
  \bibfield  {author} {\bibinfo {author} {\bibfnamefont {S.}~\bibnamefont
  {Mei}}\ and\ \bibinfo {author} {\bibfnamefont {A.}~\bibnamefont
  {Montanari}},\ }\bibfield  {title} {\bibinfo {title} {The generalization
  error of random features regression: Precise asymptotics and double descent
  curve},\ }\href@noop {} {\bibfield  {journal} {\bibinfo  {journal} {arXiv
  preprint arXiv:1908.05355}\ } (\bibinfo {year} {2019})}\BibitemShut {NoStop}%
\bibitem [{\citenamefont {Belkin}\ \emph {et~al.}(2019)\citenamefont {Belkin},
  \citenamefont {Hsu}, \citenamefont {Ma},\ and\ \citenamefont
  {Mandal}}]{belkin2019reconciling}%
  \BibitemOpen
  \bibfield  {author} {\bibinfo {author} {\bibfnamefont {M.}~\bibnamefont
  {Belkin}}, \bibinfo {author} {\bibfnamefont {D.}~\bibnamefont {Hsu}},
  \bibinfo {author} {\bibfnamefont {S.}~\bibnamefont {Ma}},\ and\ \bibinfo
  {author} {\bibfnamefont {S.}~\bibnamefont {Mandal}},\ }\bibfield  {title}
  {\bibinfo {title} {Reconciling modern machine-learning practice and the
  classical bias--variance trade-off},\ }\href@noop {} {\bibfield  {journal}
  {\bibinfo  {journal} {Proceedings of the National Academy of Sciences}\
  }\textbf {\bibinfo {volume} {116}},\ \bibinfo {pages} {15849} (\bibinfo
  {year} {2019})}\BibitemShut {NoStop}%
\bibitem [{\citenamefont {Shawe-Taylor}\ \emph {et~al.}(2004)\citenamefont
  {Shawe-Taylor}, \citenamefont {Cristianini} \emph
  {et~al.}}]{shawe2004kernel}%
  \BibitemOpen
  \bibfield  {author} {\bibinfo {author} {\bibfnamefont {J.}~\bibnamefont
  {Shawe-Taylor}}, \bibinfo {author} {\bibfnamefont {N.}~\bibnamefont
  {Cristianini}}, \emph {et~al.},\ }\href@noop {} {\emph {\bibinfo {title}
  {Kernel methods for pattern analysis}}}\ (\bibinfo  {publisher} {Cambridge
  university press},\ \bibinfo {year} {2004})\BibitemShut {NoStop}%
\bibitem [{\citenamefont {Hofmann}\ \emph {et~al.}(2008)\citenamefont
  {Hofmann}, \citenamefont {Sch{\"o}lkopf},\ and\ \citenamefont
  {Smola}}]{hofmann2008kernel}%
  \BibitemOpen
  \bibfield  {author} {\bibinfo {author} {\bibfnamefont {T.}~\bibnamefont
  {Hofmann}}, \bibinfo {author} {\bibfnamefont {B.}~\bibnamefont
  {Sch{\"o}lkopf}},\ and\ \bibinfo {author} {\bibfnamefont {A.~J.}\
  \bibnamefont {Smola}},\ }\bibfield  {title} {\bibinfo {title} {Kernel methods
  in machine learning},\ }\href@noop {} {\bibfield  {journal} {\bibinfo
  {journal} {The annals of statistics}\ ,\ \bibinfo {pages} {1171}} (\bibinfo
  {year} {2008})}\BibitemShut {NoStop}%
\bibitem [{\citenamefont {Cutajar}\ \emph {et~al.}(2017)\citenamefont
  {Cutajar}, \citenamefont {Bonilla}, \citenamefont {Michiardi},\ and\
  \citenamefont {Filippone}}]{cutajar2017random}%
  \BibitemOpen
  \bibfield  {author} {\bibinfo {author} {\bibfnamefont {K.}~\bibnamefont
  {Cutajar}}, \bibinfo {author} {\bibfnamefont {E.~V.}\ \bibnamefont
  {Bonilla}}, \bibinfo {author} {\bibfnamefont {P.}~\bibnamefont {Michiardi}},\
  and\ \bibinfo {author} {\bibfnamefont {M.}~\bibnamefont {Filippone}},\
  }\bibfield  {title} {\bibinfo {title} {Random feature expansions for deep
  gaussian processes},\ }in\ \href@noop {} {\emph {\bibinfo {booktitle}
  {International Conference on Machine Learning}}}\ (\bibinfo {organization}
  {PMLR},\ \bibinfo {year} {2017})\ pp.\ \bibinfo {pages}
  {884--893}\BibitemShut {NoStop}%
\bibitem [{\citenamefont {Rahimi}\ and\ \citenamefont
  {Recht}(2008)}]{rahimi2008random}%
  \BibitemOpen
  \bibfield  {author} {\bibinfo {author} {\bibfnamefont {A.}~\bibnamefont
  {Rahimi}}\ and\ \bibinfo {author} {\bibfnamefont {B.}~\bibnamefont {Recht}},\
  }\bibfield  {title} {\bibinfo {title} {Random features for large-scale kernel
  machines},\ }in\ \href@noop {} {\emph {\bibinfo {booktitle} {Advances in
  neural information processing systems}}}\ (\bibinfo {year} {2008})\ pp.\
  \bibinfo {pages} {1177--1184}\BibitemShut {NoStop}%
\bibitem [{\citenamefont {Vershynin}(2020)}]{vershynin2020memory}%
  \BibitemOpen
  \bibfield  {author} {\bibinfo {author} {\bibfnamefont {R.}~\bibnamefont
  {Vershynin}},\ }\bibfield  {title} {\bibinfo {title} {Memory capacity of
  neural networks with threshold and rectified linear unit activations},\
  }\href@noop {} {\bibfield  {journal} {\bibinfo  {journal} {SIAM Journal on
  Mathematics of Data Science}\ }\textbf {\bibinfo {volume} {2}},\ \bibinfo
  {pages} {1004} (\bibinfo {year} {2020})}\BibitemShut {NoStop}%
\bibitem [{\citenamefont {Gardner}(1988)}]{gardner1988space}%
  \BibitemOpen
  \bibfield  {author} {\bibinfo {author} {\bibfnamefont {E.}~\bibnamefont
  {Gardner}},\ }\bibfield  {title} {\bibinfo {title} {The space of interactions
  in neural network models},\ }\href@noop {} {\bibfield  {journal} {\bibinfo
  {journal} {Journal of physics A: Mathematical and general}\ }\textbf
  {\bibinfo {volume} {21}},\ \bibinfo {pages} {257} (\bibinfo {year}
  {1988})}\BibitemShut {NoStop}%
\bibitem [{\citenamefont {Gardner}\ and\ \citenamefont
  {Derrida}(1988)}]{gardner1988optimal}%
  \BibitemOpen
  \bibfield  {author} {\bibinfo {author} {\bibfnamefont {E.}~\bibnamefont
  {Gardner}}\ and\ \bibinfo {author} {\bibfnamefont {B.}~\bibnamefont
  {Derrida}},\ }\bibfield  {title} {\bibinfo {title} {Optimal storage
  properties of neural network models},\ }\href@noop {} {\bibfield  {journal}
  {\bibinfo  {journal} {Journal of Physics A: Mathematical and general}\
  }\textbf {\bibinfo {volume} {21}},\ \bibinfo {pages} {271} (\bibinfo {year}
  {1988})}\BibitemShut {NoStop}%
\bibitem [{\citenamefont {Domingos}(2000)}]{domingos2000unified}%
  \BibitemOpen
  \bibfield  {author} {\bibinfo {author} {\bibfnamefont {P.}~\bibnamefont
  {Domingos}},\ }\bibfield  {title} {\bibinfo {title} {A unified bias-variance
  decomposition for zero-one and squared loss},\ }\href@noop {} {\bibfield
  {journal} {\bibinfo  {journal} {AAAI/IAAI}\ }\textbf {\bibinfo {volume}
  {2000}},\ \bibinfo {pages} {564} (\bibinfo {year} {2000})}\BibitemShut
  {NoStop}%
\bibitem [{\citenamefont {Geman}\ \emph {et~al.}(1992)\citenamefont {Geman},
  \citenamefont {Bienenstock},\ and\ \citenamefont
  {Doursat}}]{geman1992neural}%
  \BibitemOpen
  \bibfield  {author} {\bibinfo {author} {\bibfnamefont {S.}~\bibnamefont
  {Geman}}, \bibinfo {author} {\bibfnamefont {E.}~\bibnamefont {Bienenstock}},\
  and\ \bibinfo {author} {\bibfnamefont {R.}~\bibnamefont {Doursat}},\
  }\bibfield  {title} {\bibinfo {title} {Neural networks and the bias/variance
  dilemma},\ }\href@noop {} {\bibfield  {journal} {\bibinfo  {journal} {Neural
  computation}\ }\textbf {\bibinfo {volume} {4}},\ \bibinfo {pages} {1}
  (\bibinfo {year} {1992})}\BibitemShut {NoStop}%
\bibitem [{\citenamefont {Laurent}\ and\ \citenamefont
  {Brecht}(2018)}]{laurent2018deep}%
  \BibitemOpen
  \bibfield  {author} {\bibinfo {author} {\bibfnamefont {T.}~\bibnamefont
  {Laurent}}\ and\ \bibinfo {author} {\bibfnamefont {J.}~\bibnamefont
  {Brecht}},\ }\bibfield  {title} {\bibinfo {title} {Deep linear networks with
  arbitrary loss: All local minima are global},\ }in\ \href@noop {} {\emph
  {\bibinfo {booktitle} {International conference on machine learning}}}\
  (\bibinfo {organization} {PMLR},\ \bibinfo {year} {2018})\ pp.\ \bibinfo
  {pages} {2902--2907}\BibitemShut {NoStop}%
\bibitem [{\citenamefont {Lu}\ and\ \citenamefont
  {Kawaguchi}(2017)}]{lu2017depth}%
  \BibitemOpen
  \bibfield  {author} {\bibinfo {author} {\bibfnamefont {H.}~\bibnamefont
  {Lu}}\ and\ \bibinfo {author} {\bibfnamefont {K.}~\bibnamefont {Kawaguchi}},\
  }\bibfield  {title} {\bibinfo {title} {Depth creates no bad local minima},\
  }\href@noop {} {\bibfield  {journal} {\bibinfo  {journal} {arXiv preprint
  arXiv:1702.08580}\ } (\bibinfo {year} {2017})}\BibitemShut {NoStop}%
\bibitem [{\citenamefont {Yamins}\ \emph {et~al.}(2014)\citenamefont {Yamins},
  \citenamefont {Hong}, \citenamefont {Cadieu}, \citenamefont {Solomon},
  \citenamefont {Seibert},\ and\ \citenamefont
  {DiCarlo}}]{yamins2014performance}%
  \BibitemOpen
  \bibfield  {author} {\bibinfo {author} {\bibfnamefont {D.~L.}\ \bibnamefont
  {Yamins}}, \bibinfo {author} {\bibfnamefont {H.}~\bibnamefont {Hong}},
  \bibinfo {author} {\bibfnamefont {C.~F.}\ \bibnamefont {Cadieu}}, \bibinfo
  {author} {\bibfnamefont {E.~A.}\ \bibnamefont {Solomon}}, \bibinfo {author}
  {\bibfnamefont {D.}~\bibnamefont {Seibert}},\ and\ \bibinfo {author}
  {\bibfnamefont {J.~J.}\ \bibnamefont {DiCarlo}},\ }\bibfield  {title}
  {\bibinfo {title} {Performance-optimized hierarchical models predict neural
  responses in higher visual cortex},\ }\href@noop {} {\bibfield  {journal}
  {\bibinfo  {journal} {Proceedings of the national academy of sciences}\
  }\textbf {\bibinfo {volume} {111}},\ \bibinfo {pages} {8619} (\bibinfo {year}
  {2014})}\BibitemShut {NoStop}%
\bibitem [{\citenamefont {Kriegeskorte}\ \emph {et~al.}(2008)\citenamefont
  {Kriegeskorte}, \citenamefont {Mur},\ and\ \citenamefont
  {Bandettini}}]{kriegeskorte2008representational}%
  \BibitemOpen
  \bibfield  {author} {\bibinfo {author} {\bibfnamefont {N.}~\bibnamefont
  {Kriegeskorte}}, \bibinfo {author} {\bibfnamefont {M.}~\bibnamefont {Mur}},\
  and\ \bibinfo {author} {\bibfnamefont {P.~A.}\ \bibnamefont {Bandettini}},\
  }\bibfield  {title} {\bibinfo {title} {Representational similarity
  analysis-connecting the branches of systems neuroscience},\ }\href@noop {}
  {\bibfield  {journal} {\bibinfo  {journal} {Frontiers in systems
  neuroscience}\ }\textbf {\bibinfo {volume} {2}},\ \bibinfo {pages} {4}
  (\bibinfo {year} {2008})}\BibitemShut {NoStop}%
\bibitem [{\citenamefont {Messinger}\ \emph {et~al.}(2001)\citenamefont
  {Messinger}, \citenamefont {Squire}, \citenamefont {Zola},\ and\
  \citenamefont {Albright}}]{messinger2001neuronal}%
  \BibitemOpen
  \bibfield  {author} {\bibinfo {author} {\bibfnamefont {A.}~\bibnamefont
  {Messinger}}, \bibinfo {author} {\bibfnamefont {L.~R.}\ \bibnamefont
  {Squire}}, \bibinfo {author} {\bibfnamefont {S.~M.}\ \bibnamefont {Zola}},\
  and\ \bibinfo {author} {\bibfnamefont {T.~D.}\ \bibnamefont {Albright}},\
  }\bibfield  {title} {\bibinfo {title} {Neuronal representations of stimulus
  associations develop in the temporal lobe during learning},\ }\href@noop {}
  {\bibfield  {journal} {\bibinfo  {journal} {Proceedings of the National
  Academy of Sciences}\ }\textbf {\bibinfo {volume} {98}},\ \bibinfo {pages}
  {12239} (\bibinfo {year} {2001})}\BibitemShut {NoStop}%
\bibitem [{\citenamefont {Dodier}(1996)}]{dodier1996geometry}%
  \BibitemOpen
  \bibfield  {author} {\bibinfo {author} {\bibfnamefont {R.}~\bibnamefont
  {Dodier}},\ }\bibfield  {title} {\bibinfo {title} {Geometry of early stopping
  in linear networks},\ }\href@noop {} {\bibfield  {journal} {\bibinfo
  {journal} {Advances in neural information processing systems}\ ,\ \bibinfo
  {pages} {365}} (\bibinfo {year} {1996})}\BibitemShut {NoStop}%
\bibitem [{\citenamefont {Li}\ \emph {et~al.}(2020)\citenamefont {Li},
  \citenamefont {Soltanolkotabi},\ and\ \citenamefont
  {Oymak}}]{li2020gradient}%
  \BibitemOpen
  \bibfield  {author} {\bibinfo {author} {\bibfnamefont {M.}~\bibnamefont
  {Li}}, \bibinfo {author} {\bibfnamefont {M.}~\bibnamefont {Soltanolkotabi}},\
  and\ \bibinfo {author} {\bibfnamefont {S.}~\bibnamefont {Oymak}},\ }\bibfield
   {title} {\bibinfo {title} {Gradient descent with early stopping is provably
  robust to label noise for overparameterized neural networks},\ }in\
  \href@noop {} {\emph {\bibinfo {booktitle} {International Conference on
  Artificial Intelligence and Statistics}}}\ (\bibinfo {organization} {PMLR},\
  \bibinfo {year} {2020})\ pp.\ \bibinfo {pages} {4313--4324}\BibitemShut
  {NoStop}%
\bibitem [{\citenamefont {Advani}\ \emph {et~al.}(2020)\citenamefont {Advani},
  \citenamefont {Saxe},\ and\ \citenamefont {Sompolinsky}}]{advani2020high}%
  \BibitemOpen
  \bibfield  {author} {\bibinfo {author} {\bibfnamefont {M.~S.}\ \bibnamefont
  {Advani}}, \bibinfo {author} {\bibfnamefont {A.~M.}\ \bibnamefont {Saxe}},\
  and\ \bibinfo {author} {\bibfnamefont {H.}~\bibnamefont {Sompolinsky}},\
  }\bibfield  {title} {\bibinfo {title} {High-dimensional dynamics of
  generalization error in neural networks},\ }\href@noop {} {\bibfield
  {journal} {\bibinfo  {journal} {Neural Networks}\ }\textbf {\bibinfo {volume}
  {132}},\ \bibinfo {pages} {428} (\bibinfo {year} {2020})}\BibitemShut
  {NoStop}%
\bibitem [{\citenamefont {Naveh}\ \emph {et~al.}(2020)\citenamefont {Naveh},
  \citenamefont {Ben-David}, \citenamefont {Sompolinsky},\ and\ \citenamefont
  {Ringel}}]{naveh2020predicting}%
  \BibitemOpen
  \bibfield  {author} {\bibinfo {author} {\bibfnamefont {G.}~\bibnamefont
  {Naveh}}, \bibinfo {author} {\bibfnamefont {O.}~\bibnamefont {Ben-David}},
  \bibinfo {author} {\bibfnamefont {H.}~\bibnamefont {Sompolinsky}},\ and\
  \bibinfo {author} {\bibfnamefont {Z.}~\bibnamefont {Ringel}},\ }\bibfield
  {title} {\bibinfo {title} {Predicting the outputs of finite networks trained
  with noisy gradients},\ }\href@noop {} {\bibfield  {journal} {\bibinfo
  {journal} {arXiv preprint arXiv:2004.01190}\ } (\bibinfo {year}
  {2020})}\BibitemShut {NoStop}%
\bibitem [{\citenamefont {Antognini}(2019)}]{antognini2019finite}%
  \BibitemOpen
  \bibfield  {author} {\bibinfo {author} {\bibfnamefont {J.~M.}\ \bibnamefont
  {Antognini}},\ }\bibfield  {title} {\bibinfo {title} {Finite size corrections
  for neural network gaussian processes},\ }\href@noop {} {\bibfield  {journal}
  {\bibinfo  {journal} {arXiv preprint arXiv:1908.10030}\ } (\bibinfo {year}
  {2019})}\BibitemShut {NoStop}%
\bibitem [{\citenamefont {Yedidia}\ \emph {et~al.}(2000)\citenamefont
  {Yedidia}, \citenamefont {Freeman}, \citenamefont {Weiss} \emph
  {et~al.}}]{yedidia2000generalized}%
  \BibitemOpen
  \bibfield  {author} {\bibinfo {author} {\bibfnamefont {J.~S.}\ \bibnamefont
  {Yedidia}}, \bibinfo {author} {\bibfnamefont {W.~T.}\ \bibnamefont
  {Freeman}}, \bibinfo {author} {\bibfnamefont {Y.}~\bibnamefont {Weiss}},
  \emph {et~al.},\ }\bibfield  {title} {\bibinfo {title} {Generalized belief
  propagation},\ }in\ \href@noop {} {\emph {\bibinfo {booktitle} {NIPS}}},\
  Vol.~\bibinfo {volume} {13}\ (\bibinfo {year} {2000})\ pp.\ \bibinfo {pages}
  {689--695}\BibitemShut {NoStop}%
\bibitem [{\citenamefont {Yedidia}\ \emph {et~al.}(2003)\citenamefont
  {Yedidia}, \citenamefont {Freeman}, \citenamefont {Weiss} \emph
  {et~al.}}]{yedidia2003understanding}%
  \BibitemOpen
  \bibfield  {author} {\bibinfo {author} {\bibfnamefont {J.~S.}\ \bibnamefont
  {Yedidia}}, \bibinfo {author} {\bibfnamefont {W.~T.}\ \bibnamefont
  {Freeman}}, \bibinfo {author} {\bibfnamefont {Y.}~\bibnamefont {Weiss}},
  \emph {et~al.},\ }\bibfield  {title} {\bibinfo {title} {Understanding belief
  propagation and its generalizations},\ }\href@noop {} {\bibfield  {journal}
  {\bibinfo  {journal} {Exploring artificial intelligence in the new
  millennium}\ }\textbf {\bibinfo {volume} {8}},\ \bibinfo {pages} {236}
  (\bibinfo {year} {2003})}\BibitemShut {NoStop}%
\bibitem [{\citenamefont {Yedidia}\ \emph {et~al.}(2005)\citenamefont
  {Yedidia}, \citenamefont {Freeman},\ and\ \citenamefont
  {Weiss}}]{yedidia2005constructing}%
  \BibitemOpen
  \bibfield  {author} {\bibinfo {author} {\bibfnamefont {J.~S.}\ \bibnamefont
  {Yedidia}}, \bibinfo {author} {\bibfnamefont {W.~T.}\ \bibnamefont
  {Freeman}},\ and\ \bibinfo {author} {\bibfnamefont {Y.}~\bibnamefont
  {Weiss}},\ }\bibfield  {title} {\bibinfo {title} {Constructing free-energy
  approximations and generalized belief propagation algorithms},\ }\href@noop
  {} {\bibfield  {journal} {\bibinfo  {journal} {IEEE Transactions on
  information theory}\ }\textbf {\bibinfo {volume} {51}},\ \bibinfo {pages}
  {2282} (\bibinfo {year} {2005})}\BibitemShut {NoStop}%
\bibitem [{\citenamefont {Pearl}(1986)}]{pearl1986fusion}%
  \BibitemOpen
  \bibfield  {author} {\bibinfo {author} {\bibfnamefont {J.}~\bibnamefont
  {Pearl}},\ }\bibfield  {title} {\bibinfo {title} {Fusion, propagation, and
  structuring in belief networks},\ }\href@noop {} {\bibfield  {journal}
  {\bibinfo  {journal} {Artificial intelligence}\ }\textbf {\bibinfo {volume}
  {29}},\ \bibinfo {pages} {241} (\bibinfo {year} {1986})}\BibitemShut
  {NoStop}%
\bibitem [{\citenamefont {Pearl}(2014)}]{pearl2014probabilistic}%
  \BibitemOpen
  \bibfield  {author} {\bibinfo {author} {\bibfnamefont {J.}~\bibnamefont
  {Pearl}},\ }\href@noop {} {\emph {\bibinfo {title} {Probabilistic reasoning
  in intelligent systems: networks of plausible inference}}}\ (\bibinfo
  {publisher} {Elsevier},\ \bibinfo {year} {2014})\BibitemShut {NoStop}%
\bibitem [{\citenamefont {Weiss}\ and\ \citenamefont
  {Pearl}(2010)}]{weiss2010belief}%
  \BibitemOpen
  \bibfield  {author} {\bibinfo {author} {\bibfnamefont {Y.}~\bibnamefont
  {Weiss}}\ and\ \bibinfo {author} {\bibfnamefont {J.}~\bibnamefont {Pearl}},\
  }\bibfield  {title} {\bibinfo {title} {Belief propagation: technical
  perspective},\ }\href@noop {} {\bibfield  {journal} {\bibinfo  {journal}
  {Communications of the ACM}\ }\textbf {\bibinfo {volume} {53}},\ \bibinfo
  {pages} {94} (\bibinfo {year} {2010})}\BibitemShut {NoStop}%
\bibitem [{\citenamefont {Winn}\ \emph {et~al.}(2005)\citenamefont {Winn},
  \citenamefont {Bishop},\ and\ \citenamefont
  {Jaakkola}}]{winn2005variational}%
  \BibitemOpen
  \bibfield  {author} {\bibinfo {author} {\bibfnamefont {J.}~\bibnamefont
  {Winn}}, \bibinfo {author} {\bibfnamefont {C.~M.}\ \bibnamefont {Bishop}},\
  and\ \bibinfo {author} {\bibfnamefont {T.}~\bibnamefont {Jaakkola}},\
  }\bibfield  {title} {\bibinfo {title} {Variational message passing.},\
  }\href@noop {} {\bibfield  {journal} {\bibinfo  {journal} {Journal of Machine
  Learning Research}\ }\textbf {\bibinfo {volume} {6}} (\bibinfo {year}
  {2005})}\BibitemShut {NoStop}%
\bibitem [{\citenamefont {Parr}\ \emph {et~al.}(2019)\citenamefont {Parr},
  \citenamefont {Markovic}, \citenamefont {Kiebel},\ and\ \citenamefont
  {Friston}}]{parr2019neuronal}%
  \BibitemOpen
  \bibfield  {author} {\bibinfo {author} {\bibfnamefont {T.}~\bibnamefont
  {Parr}}, \bibinfo {author} {\bibfnamefont {D.}~\bibnamefont {Markovic}},
  \bibinfo {author} {\bibfnamefont {S.~J.}\ \bibnamefont {Kiebel}},\ and\
  \bibinfo {author} {\bibfnamefont {K.~J.}\ \bibnamefont {Friston}},\
  }\bibfield  {title} {\bibinfo {title} {Neuronal message passing using
  mean-field, bethe, and marginal approximations},\ }\href@noop {} {\bibfield
  {journal} {\bibinfo  {journal} {Scientific reports}\ }\textbf {\bibinfo
  {volume} {9}},\ \bibinfo {pages} {1} (\bibinfo {year} {2019})}\BibitemShut
  {NoStop}%
\bibitem [{\citenamefont {Kschischang}\ \emph {et~al.}(2001)\citenamefont
  {Kschischang}, \citenamefont {Frey},\ and\ \citenamefont
  {Loeliger}}]{kschischang2001factor}%
  \BibitemOpen
  \bibfield  {author} {\bibinfo {author} {\bibfnamefont {F.~R.}\ \bibnamefont
  {Kschischang}}, \bibinfo {author} {\bibfnamefont {B.~J.}\ \bibnamefont
  {Frey}},\ and\ \bibinfo {author} {\bibfnamefont {H.-A.}\ \bibnamefont
  {Loeliger}},\ }\bibfield  {title} {\bibinfo {title} {Factor graphs and the
  sum-product algorithm},\ }\href@noop {} {\bibfield  {journal} {\bibinfo
  {journal} {IEEE Transactions on information theory}\ }\textbf {\bibinfo
  {volume} {47}},\ \bibinfo {pages} {498} (\bibinfo {year} {2001})}\BibitemShut
  {NoStop}%
\bibitem [{\citenamefont {Hern{\'a}ndez-Lobato}\ and\ \citenamefont
  {Adams}(2015)}]{hernandez2015probabilistic}%
  \BibitemOpen
  \bibfield  {author} {\bibinfo {author} {\bibfnamefont {J.~M.}\ \bibnamefont
  {Hern{\'a}ndez-Lobato}}\ and\ \bibinfo {author} {\bibfnamefont
  {R.}~\bibnamefont {Adams}},\ }\bibfield  {title} {\bibinfo {title}
  {Probabilistic backpropagation for scalable learning of bayesian neural
  networks},\ }in\ \href@noop {} {\emph {\bibinfo {booktitle} {International
  Conference on Machine Learning}}}\ (\bibinfo {organization} {PMLR},\ \bibinfo
  {year} {2015})\ pp.\ \bibinfo {pages} {1861--1869}\BibitemShut {NoStop}%
\bibitem [{\citenamefont {Graves}(2011)}]{graves2011practical}%
  \BibitemOpen
  \bibfield  {author} {\bibinfo {author} {\bibfnamefont {A.}~\bibnamefont
  {Graves}},\ }\bibfield  {title} {\bibinfo {title} {Practical variational
  inference for neural networks},\ }in\ \href@noop {} {\emph {\bibinfo
  {booktitle} {Advances in neural information processing systems}}}\ (\bibinfo
  {organization} {Citeseer},\ \bibinfo {year} {2011})\ pp.\ \bibinfo {pages}
  {2348--2356}\BibitemShut {NoStop}%
\bibitem [{\citenamefont {Goldenfeld}(2018)}]{goldenfeld2018lectures}%
  \BibitemOpen
  \bibfield  {author} {\bibinfo {author} {\bibfnamefont {N.}~\bibnamefont
  {Goldenfeld}},\ }\href@noop {} {\emph {\bibinfo {title} {Lectures on phase
  transitions and the renormalization group}}}\ (\bibinfo  {publisher} {CRC
  Press},\ \bibinfo {year} {2018})\BibitemShut {NoStop}%
\bibitem [{\citenamefont {Chen}\ \emph {et~al.}(1996)\citenamefont {Chen},
  \citenamefont {Goldenfeld},\ and\ \citenamefont
  {Oono}}]{chen1996renormalization}%
  \BibitemOpen
  \bibfield  {author} {\bibinfo {author} {\bibfnamefont {L.-Y.}\ \bibnamefont
  {Chen}}, \bibinfo {author} {\bibfnamefont {N.}~\bibnamefont {Goldenfeld}},\
  and\ \bibinfo {author} {\bibfnamefont {Y.}~\bibnamefont {Oono}},\ }\bibfield
  {title} {\bibinfo {title} {Renormalization group and singular perturbations:
  Multiple scales, boundary layers, and reductive perturbation theory},\
  }\href@noop {} {\bibfield  {journal} {\bibinfo  {journal} {Physical Review
  E}\ }\textbf {\bibinfo {volume} {54}},\ \bibinfo {pages} {376} (\bibinfo
  {year} {1996})}\BibitemShut {NoStop}%
\bibitem [{\citenamefont {Pierson}\ and\ \citenamefont
  {Valls}(1992)}]{pierson1992renormalization}%
  \BibitemOpen
  \bibfield  {author} {\bibinfo {author} {\bibfnamefont {S.~W.}\ \bibnamefont
  {Pierson}}\ and\ \bibinfo {author} {\bibfnamefont {O.~T.}\ \bibnamefont
  {Valls}},\ }\bibfield  {title} {\bibinfo {title} {Renormalization-group study
  of a layered-superconductor model},\ }\href@noop {} {\bibfield  {journal}
  {\bibinfo  {journal} {Physical Review B}\ }\textbf {\bibinfo {volume} {45}},\
  \bibinfo {pages} {13076} (\bibinfo {year} {1992})}\BibitemShut {NoStop}%
\bibitem [{\citenamefont {Pierson}(1994)}]{pierson1994critical}%
  \BibitemOpen
  \bibfield  {author} {\bibinfo {author} {\bibfnamefont {S.~W.}\ \bibnamefont
  {Pierson}},\ }\bibfield  {title} {\bibinfo {title} {Critical behavior of
  vortices in a layered system},\ }\href@noop {} {\bibfield  {journal}
  {\bibinfo  {journal} {Physical review letters}\ }\textbf {\bibinfo {volume}
  {73}},\ \bibinfo {pages} {2496} (\bibinfo {year} {1994})}\BibitemShut
  {NoStop}%
\bibitem [{\citenamefont {Li}\ and\ \citenamefont {Wang}(2018)}]{li2018neural}%
  \BibitemOpen
  \bibfield  {author} {\bibinfo {author} {\bibfnamefont {S.-H.}\ \bibnamefont
  {Li}}\ and\ \bibinfo {author} {\bibfnamefont {L.}~\bibnamefont {Wang}},\
  }\bibfield  {title} {\bibinfo {title} {Neural network renormalization
  group},\ }\href@noop {} {\bibfield  {journal} {\bibinfo  {journal} {Physical
  review letters}\ }\textbf {\bibinfo {volume} {121}},\ \bibinfo {pages}
  {260601} (\bibinfo {year} {2018})}\BibitemShut {NoStop}%
\bibitem [{\citenamefont {Van~der Wilk}\ \emph {et~al.}(2017)\citenamefont
  {Van~der Wilk}, \citenamefont {Rasmussen},\ and\ \citenamefont
  {Hensman}}]{van2017convolutional}%
  \BibitemOpen
  \bibfield  {author} {\bibinfo {author} {\bibfnamefont {M.}~\bibnamefont
  {Van~der Wilk}}, \bibinfo {author} {\bibfnamefont {C.~E.}\ \bibnamefont
  {Rasmussen}},\ and\ \bibinfo {author} {\bibfnamefont {J.}~\bibnamefont
  {Hensman}},\ }\bibfield  {title} {\bibinfo {title} {Convolutional gaussian
  processes},\ }in\ \href@noop {} {\emph {\bibinfo {booktitle} {Advances in
  Neural Information Processing Systems}}}\ (\bibinfo {year} {2017})\ pp.\
  \bibinfo {pages} {2849--2858}\BibitemShut {NoStop}%
\bibitem [{\citenamefont {Garriga-Alonso}\ \emph {et~al.}(2018)\citenamefont
  {Garriga-Alonso}, \citenamefont {Rasmussen},\ and\ \citenamefont
  {Aitchison}}]{garriga2018deep}%
  \BibitemOpen
  \bibfield  {author} {\bibinfo {author} {\bibfnamefont {A.}~\bibnamefont
  {Garriga-Alonso}}, \bibinfo {author} {\bibfnamefont {C.~E.}\ \bibnamefont
  {Rasmussen}},\ and\ \bibinfo {author} {\bibfnamefont {L.}~\bibnamefont
  {Aitchison}},\ }\bibfield  {title} {\bibinfo {title} {Deep convolutional
  networks as shallow gaussian processes},\ }in\ \href@noop {} {\emph {\bibinfo
  {booktitle} {International Conference on Learning Representations}}}\
  (\bibinfo {year} {2018})\BibitemShut {NoStop}%
\bibitem [{\citenamefont {Novak}\ \emph {et~al.}(2018)\citenamefont {Novak},
  \citenamefont {Xiao}, \citenamefont {Lee}, \citenamefont {Bahri},
  \citenamefont {Yang}, \citenamefont {Hron}, \citenamefont {Abolafia},
  \citenamefont {Pennington},\ and\ \citenamefont
  {Sohl-Dickstein}}]{novak2018bayesian}%
  \BibitemOpen
  \bibfield  {author} {\bibinfo {author} {\bibfnamefont {R.}~\bibnamefont
  {Novak}}, \bibinfo {author} {\bibfnamefont {L.}~\bibnamefont {Xiao}},
  \bibinfo {author} {\bibfnamefont {J.}~\bibnamefont {Lee}}, \bibinfo {author}
  {\bibfnamefont {Y.}~\bibnamefont {Bahri}}, \bibinfo {author} {\bibfnamefont
  {G.}~\bibnamefont {Yang}}, \bibinfo {author} {\bibfnamefont {J.}~\bibnamefont
  {Hron}}, \bibinfo {author} {\bibfnamefont {D.~A.}\ \bibnamefont {Abolafia}},
  \bibinfo {author} {\bibfnamefont {J.}~\bibnamefont {Pennington}},\ and\
  \bibinfo {author} {\bibfnamefont {J.}~\bibnamefont {Sohl-Dickstein}},\
  }\bibfield  {title} {\bibinfo {title} {Bayesian deep convolutional networks
  with many channels are gaussian processes},\ }\href@noop {} {\bibfield
  {journal} {\bibinfo  {journal} {arXiv preprint arXiv:1810.05148}\ } (\bibinfo
  {year} {2018})}\BibitemShut {NoStop}%
\bibitem [{\citenamefont {Bai}\ and\ \citenamefont
  {Silverstein}(2010)}]{bai2010sample}%
  \BibitemOpen
  \bibfield  {author} {\bibinfo {author} {\bibfnamefont {Z.}~\bibnamefont
  {Bai}}\ and\ \bibinfo {author} {\bibfnamefont {J.~W.}\ \bibnamefont
  {Silverstein}},\ }\bibfield  {title} {\bibinfo {title} {Sample covariance
  matrices and the mar{\v{c}}enko-pastur law},\ }in\ \href@noop {} {\emph
  {\bibinfo {booktitle} {Spectral Analysis of Large Dimensional Random
  Matrices}}}\ (\bibinfo  {publisher} {Springer},\ \bibinfo {year} {2010})\
  pp.\ \bibinfo {pages} {39--58}\BibitemShut {NoStop}%
\end{thebibliography}
%

\end{document}